\documentclass[11pt]{article}

\usepackage[final]{acl}

\usepackage{times}
\usepackage{latexsym}
\usepackage[T1]{fontenc}
\usepackage[utf8]{inputenc}

\usepackage{float}
\usepackage{stfloats} 

\usepackage{microtype}
\usepackage{inconsolata}

\title{Towards Effective Extraction and Evaluation of Factual Claims}

\author{
  Dasha Metropolitansky, Jonathan Larson \\
  Microsoft Research \\
  \texttt{\{dasham,jolarso\}@microsoft.com}
}

\usepackage{hyperref}
\usepackage{graphicx}
\usepackage{multirow}
\usepackage{booktabs}
\usepackage{geometry}
\usepackage{array}
\usepackage{lscape} 
\usepackage{import}
\usepackage{adjustbox} 
\usepackage{xcolor}
\usepackage{soul}
\usepackage{enumitem}
\usepackage{ragged2e}
\usepackage{amsmath, amssymb}
\usepackage[most]{tcolorbox}
\usepackage{natbib}
\usepackage{placeins}

\definecolor{lightgreen}{RGB}{204,255,204}
\definecolor{lightred}{RGB}{255,204,204}

\newcommand{\tightarrow}{\!\Rightarrow\!}
\newcommand{\tightnarrow}{\!\nRightarrow\!}
\newcommand{\tinyand}{\hspace{1pt}\land\hspace{1pt}}

\makeatletter
\newcommand*{\appautoref}[1]{%
  \hyperref[#1]{Appendix~\ref*{#1}}%
}
\makeatother

\begin{document}
\maketitle
\begin{abstract}
A common strategy for fact-checking long-form content generated by Large Language Models (LLMs) is extracting simple claims that can be verified independently. Since inaccurate or incomplete claims compromise fact-checking results, ensuring claim quality is critical. However, the lack of a standardized evaluation framework impedes assessment and comparison of claim extraction methods. To address this gap, we propose a framework for evaluating claim extraction in the context of fact-checking along with automated, scalable, and replicable methods for applying this framework, including novel approaches for measuring coverage and decontextualization. We also introduce Claimify, an LLM-based claim extraction method, and demonstrate that it outperforms existing methods under our evaluation framework. A key feature of Claimify is its ability to handle ambiguity and extract claims only when there is high confidence in the correct interpretation of the source text.
\end{abstract}

\section{Introduction}
It is well known that Large Language Models (LLMs) are prone to producing unsubstantiated or inaccurate content \citep{huang:2025}. As LLM-generated content grows in volume and influence, reliable fact-checking systems become increasingly important. 

For long-form, information rich outputs, a common fact-checking strategy is to extract simple ``claims'' from the text, then retrieve relevant evidence and assess the veracity of each claim independently \citep{min:2023, hu:2024:decompdilemmas}. The effectiveness of such ``decompose-then-verify'' systems is contingent on the quality of the extracted claims: misrepresenting the source text or omitting factual content can result in misleading or incomplete conclusions. Therefore, rigorous evaluation of claim extraction methods is critical. 

While prior works have identified desirable properties of claims, classified common errors, and shown that fact-checking performance is sensitive to the decomposition method, there is currently no standardized approach for evaluating claim extraction \citep{hu:2024:decompdilemmas, wanner:2024:decomp}.

This paper makes the following contributions:
\begin{enumerate}
    \item We propose a framework for evaluating claim extraction methods in the context of fact-checking. We also introduce automated, scalable, and replicable methods for applying this framework, which are validated through human review. Two key innovations are: (1) a granular assessment of claims’ coverage of the source text, and (2) an outcome-based approach for evaluating decontextualization (i.e., whether a claim contains all necessary contextual information).
    \item We introduce Claimify, an LLM-based claim extraction method. We demonstrate that it outperforms existing methods under our evaluation framework. To the best of our knowledge, Claimify is also the first claim extraction method that identifies sentences with multiple possible interpretations and determines when the correct interpretation cannot be inferred from the sentence’s context – unlike existing methods, which either ignore ambiguity or assume it is always resolvable.
\end{enumerate}

\section{Evaluating Claim Extraction}
\label{sec:framework}
\subsection{Key Concepts}
\label{subsec:framework_definitions}

The definition of a ``claim'' varies across prior works \citep{daxenberger:2017}. We adopt the perspective from \citet{ni:2024}, which focuses on statements that ``present verifiable facts,'' where a fact is ``a statement or assertion that can be objectively verified as true or false based on empirical evidence or reality.'' We use the term ``factual claim'' throughout this paper instead of the full phrase ``verifiable factual claim.''

We argue that in the context of fact-checking, claim extraction methods should be evaluated based on three factors:
\begin{enumerate}
    \item \textbf{Entailment} means that if the source text is true, the extracted claims must also be true. The broader principle that the source text should support the claims has been described in previous works as faithfulness \citep{wright:2022,hu:2024:refchecker,chen:2024}, coherence \citep{wanner:2024:decomp}, and correctness \citep{kamoi:2023}. 
    \item \textbf{Coverage} means that extracted claims should capture the verifiable information in the source text without explicitly including the unverifiable information. We discuss a novel approach to evaluating coverage in \autoref{subsec:coverage}.
    \item \textbf{Decontextualization} is typically defined as: (1) each claim should be understandable on its own, without requiring additional context, and (2) each claim should retain the meaning it held in its original context \citep{choi:2021, gunjal:2024}. We propose an alternative definition in \autoref{subsec:decontext}.
\end{enumerate}

Claim extraction methods have also been evaluated based on \textbf{atomicity} \citep{wanner:2024:decomp, chen:2024}. For example, the claim \textit{``California and New York implemented a plastic bag ban''} is not atomic because it can be divided into \textit{``California implemented a plastic bag ban''} and \textit{``New York implemented a plastic bag ban.''} However, the pursuit of atomicity lacks a clear endpoint: the above claims could be further divided into \textit{``At least one state has implemented a plastic bag ban,''} \textit{``California has implemented a ban,''} and \textit{``California exists.''} Moreover, prior works suggest that atomicity does not consistently improve fact-checking performance \citep{chen:2023:felm, hu:2024:decompdilemmas, tang:2024}. As a result, we do not consider atomicity in our evaluation framework.

\subsection{Coverage}
\label{subsec:coverage}

Coverage (as defined in \autoref{subsec:framework_definitions}) can be evaluated at different levels of granularity. Prior works have primarily focused on \textbf{sentence-level} evaluation, assessing whether a method correctly determines that a sentence, as a whole, contains a factual claim (\citealp{konstantinovskiy:2021, majer:2024}). 

Consider the sentence ``\textit{The iconic American flag has 50 stars and 13 stripes,}'' where Method A extracts the claims [\textit{``The American flag is iconic'', ``The American flag has numerous stars and stripes''}] and Method B extracts the claims [\textit{``The American flag contains 50 stars'', ``The American flag contains 13 stripes''}]. Both methods correctly identified that the sentence contains a factual claim, so they performed equally well in terms of sentence-level coverage.

In contrast, we introduce the concept of \textbf{element-level coverage}, evaluated by breaking a sentence into distinct pieces of information (``elements''), classifying each element as verifiable or unverifiable, then assessing whether each element is ``covered'' by (i.e., present in) the extracted claims, either explicitly or implicitly (i.e., the element is stated or suggested). 

We define a \textbf{true positive} as a verifiable element that is covered implicitly or explicitly by the claims; a \textbf{true negative} as an unverifiable element that is either not covered or only implicitly covered (since implicit coverage may not reflect deliberate inclusion); a \textbf{false positive} as an unverifiable element that is explicitly covered; and a \textbf{false negative} as a verifiable element that is not covered. In \appautoref{app:coverage_example}, we provide an example that illustrates why explicit coverage of unverifiable elements should count as a false positive, but implicit coverage should not.

Unlike sentence-level coverage, element-level coverage recognizes that Method B is superior to Method A. The sentence contains three elements: (1) the American flag is iconic, (2) the American flag has 50 stars, and (3) the American flag has 13 stripes. Only elements 2 and 3 are verifiable. Method A has one false positive (it explicitly covers element 1) and two false negatives (it does not mention the number of stars and stripes, so it fails to cover elements 2 and 3), while Method B has one true negative (it does not cover element 1) and two true positives (it covers elements 2 and 3).

Prior works that came closest to evaluating element-level coverage, such as \citet{song:2024} and \citet{li:2024}, (1) relied on human annotation, making them difficult to scale, (2) lacked specificity (e.g., they considered whether verifiable content was omitted without quantifying the omissions), (3) failed to penalize the inclusion of unverifiable content, and/or (4) did not distinguish between implicit and explicit coverage.

\subsection{Decontextualization}
\label{subsec:decontext}

Numerous studies rely on human annotations to assess whether a unit of text (e.g., a sentence or claim) is sufficiently decontextualized \citep{choi:2021, kane:2023, bayat:2025}. However, we argue that such judgments are often subjective, difficult to apply consistently, and fail to reflect the claim’s suitability for fact-checking. 

Consider the claim ``\textit{John Smith supports government regulations}'' extracted from the sentence ``\textit{In the latest episode of Jane Doe’s podcast on electric vehicles, Doe’s free-market views clashed with John Smith’s support for government regulations.}'' According to the definition in \autoref{subsec:framework_definitions}, the claim appears sufficiently decontextualized. 

However, if a fact-checking system attempted to verify this claim, it might find evidence of John Smith opposing government regulations in other contexts (e.g., AI or healthcare) and conclude that the claim is false – even though this evidence does not contradict the source sentence. The mismatch between the evidence’s implications for the claim and for the sentence indicates that the claim was insufficiently decontextualized: it should have clarified that John’s comments were made during a specific podcast episode about electric vehicles. Critically, the underspecification only became apparent after the fact-checking process, not beforehand. 

Consider a second example: ``\textit{The court helped secure Bush’s presidency through its split decision to halt the Florida recount.}'' The claim appears to be insufficiently decontextualized, since ``\textit{The court}'' is not defined. However, for fact-checking purposes, the underspecification is not problematic, since the only plausible reference is the Bush v. Gore decision by the United States Supreme Court.

We posit that instead of making subjective judgments about whether a claim is ``sufficiently'' decontextualized, we should measure how the claim affects the outcome of the fact-checking system. In a fact-checking system, claims are used to retrieve evidence from a collection of documents, which informs a true/false verdict. Therefore, missing context is problematic only if its inclusion would change the verdict from true to false, or vice versa.\footnote{Prior works propose retrieval-based evaluations of decontextualization \citep{choi:2021, deng:2024}. However, in fact-checking, such approaches are insufficient because retrieval is only an intermediate step towards the final verdict.} This shift could occur if including the context results in retrieving a different pool of evidence with the opposite relationship to the claim, or if the same evidence is retrieved but its relationship to the claim changes when viewed with the added context.

Accordingly, we propose a three-step process for evaluating the decontextualization of a claim $c$ in the context of fact-checking:
\begin{enumerate}
    \item \textbf{Identify Missing Context.} Based on $c$ and its context, either:
    \begin{itemize}
        \item Generate $c_{\max}$, a maximally decontextualized version of $c$, ensuring $c$ is entailed by $c_{\max}$; or
        \item Determine that $c$ is already maximally decontextualized (i.e., $c = c_{\max}$)
    \end{itemize}
    In the John Smith example, $c_{\max}$ might be: \textit{``In the latest episode of Jane Doe’s podcast on electric vehicles, John Smith supports government regulations.''} If $c$ is already maximally decontextualized, no further steps are needed.\footnote{Just as there are often multiple ways to decontextualize a sentence, there is rarely a single ``correct'' formulation of $c_{\max}$. However, we argue that creating a claim that contains as much context as possible is less subjective than trying to evaluate whether a claim is ``sufficiently'' decontextualized. The evaluation can also be repeated for different $c_{\max}$ values to ensure robustness.} 

    \item \textbf{Retrieve Evidence.} In the collection of documents used for fact-checking, find relevant information for $c$ and $c_{\max}$, producing evidence sets $E_\mathrm{c}$ and $E_{\max}$, respectively.

    \item \textbf{Determine Veracity.} Perform the following checks\footnote{If $E_\mathrm{c}\tightnarrow c$, there is no need to check whether $E_\mathrm{c}\tightarrow c_{\max}$. Since $c_{\max}$ entails $c$, and $c$ is narrower than $c_{\max}$, any evidence that fails to support $c$ cannot support $c_{\max}$.}:
    \begin{itemize}
        \item $E_\mathrm{c}\tightarrow c$ (i.e., check if $E_\mathrm{c}$ supports $c$)
        \item $E_{\max}\tightarrow c_{\max}$ 
        \item If $E_\mathrm{c}\tightarrow c$, check if $E_\mathrm{c}  \tightarrow c_{\max}$
    \end{itemize}
\end{enumerate}
This process yields one of seven possible results: 
\begin{enumerate}
    \item $c = c_{\max}$
    \item $(E_\mathrm{c}\tightarrow c)\tinyand(E_{\max}\tightarrow c_{\max})\tinyand(E_\mathrm{c}\tightarrow c_{\max})$
    \item $(E_\mathrm{c}\tightarrow c)\tinyand(E_{\max}\tightarrow c_{\max})\tinyand(E_\mathrm{c}\tightnarrow c_{\max})$
    \item $(E_\mathrm{c}\tightarrow c)\tinyand(E_{\max}\tightnarrow c_{\max})\tinyand(E_\mathrm{c}\tightarrow c_{\max})$
    \item $(E_\mathrm{c}\tightarrow c)\tinyand(E_{\max}\tightnarrow c_{\max})\tinyand(E_\mathrm{c}\tightnarrow c_{\max})$
    \item $(E_\mathrm{c}\tightnarrow c)\tinyand(E_{\max}\tightarrow c_{\max})$
    \item $(E_\mathrm{c}\tightnarrow c)\tinyand(E_{\max}\tightnarrow c_{\max})$
\end{enumerate}

Results 5 and 6 are undesirable because the verdicts for $c$ and $c_{\max}$ are misaligned. Result 3 – where $c$ and $c_{\max}$ are supported by their respective evidence sets, but the evidence for $c$ does not support $c_{\max}$ – is problematic in scenarios where the rationale matters, not just the verdict.\footnote{Consider $c =$ ``\textit{Miller has been described as an architect,}'' extracted from the sentence $s =$ ``\textit{Miller has been described as the architect of Trump’s controversial immigration policies.}'' Let $c_{\max} = s$. Imagine $E_\mathrm{c}$ contains information about a building architect named John Miller, while $E_{\max}$ describes Stephen Miller, President Trump’s policy advisor, as an architect of the administration’s immigration policies. Although both $c$ and $c_{\max}$ are supported by their respective evidence sets, it would be highly problematic if a fact-checking system cited $E_\mathrm{c}$ as its rationale for the sentence’s veracity! Note that $c$ and $s$ were adapted from an example by \citet{wanner:2024:dnd}.} 

In contrast, Results 2 and 7 are desirable because the verdicts for $c$ and $c_{\max}$ are aligned. Result 4 is also favorable – in fact, it suggests that $c$ is superior to $c_{\max}$ because only the former retrieved evidence supporting both $c$ and $c_{\max}$. We classify Result 1 as desirable because it indicates that no contextual information was omitted.

We describe our implementation of this approach in \autoref{subsec:decontext_results}, comparing claim extraction methods based on the percentage of desirable results. 

\section{Claimify}
\label{sec:claimify}

This section describes Claimify, our novel LLM-based claim extraction method. \autoref{fig:claimify} in \appautoref{app:claimify} illustrates its key stages, and \appautoref{app:claimify_prompts} contains all prompts.

\subsection{Sentence Splitting and Context Creation}
\label{subsec:claimify_sentences}
Claimify accepts a question-answer pair as input. It uses NLTK’s sentence tokenizer to split the answer into sentences \citep[version 3.9.1]{bird:2004}. Context is created for each sentence $s$ based on a configurable combination of $p$ preceding sentences, $f$ following sentences, and optional metadata (e.g., the header hierarchy in a Markdown-style answer).\footnote{We did not use any metadata for the experiments described in this paper.} The parameters $p$ and $f$ are defined separately for the stages outlined in \autoref{subsec:selection}-\autoref{subsec:decomposition}, allowing each stage to have a distinct context.

\subsection{Selection}
\label{subsec:selection}

Next, Claimify uses an LLM to determine whether each sentence contains any verifiable content, in light of its context and the question. When the LLM identifies that a sentence contains both verifiable and unverifiable components, it rewrites the sentence, retaining only the verifiable components. 

More specifically, the LLM selects one of the following options: (1) state that the sentence does not contain any verifiable content, (2) return a modified version of the sentence that retains only verifiable content, or (3) return the original sentence, indicating that it does not contain any unverifiable content. If the LLM selects the first option, the sentence is labeled ``No verifiable claims'' and excluded from subsequent stages (\autoref{subsec:disambiguation} and \autoref{subsec:decomposition}). \autoref{tab:selection_examples} in \appautoref{app:claimify} provides examples where the LLM selected the first or second option.

\subsection{Disambiguation}
\label{subsec:disambiguation}
The primary goals of this stage are to identify ambiguity in the sentences returned by the Selection stage, and to determine whether the ambiguity has a clear resolution based on the question and the context. These objectives and capabilities are unique to Claimify (see \autoref{sec:related_work} for a discussion of related works). 

Claimify uses an LLM to identify two types of ambiguity. The first is \textbf{referential ambiguity}, which occurs when it is unclear what a word or phrase refers to. For example, in the sentence ``\textit{They will update the policy next year,}'' the terms ``\textit{They,}'' ``\textit{the policy,}'' and ``\textit{next year}'' are ambiguous. The second is \textbf{structural ambiguity}, which occurs when grammatical structure allows for multiple interpretations. For instance, the sentence ``\textit{AI has advanced renewable energy and sustainable agriculture at Company A and Company B}'' can be interpreted as: (1) AI has advanced renewable energy and sustainable agriculture at both Company A and Company B, or (2) AI has advanced renewable energy at Company A, and it has advanced sustainable agriculture at Company B. 

A special case of structural ambiguity involves distinguishing between factual claims and unverifiable interpretations added by the author. For example, the sentence ``\textit{John emphasized the support he received from executives throughout his career, highlighting the importance of mentorship,}'' can be interpreted as: (1) John both emphasized the support he received and highlighted the importance of mentorship, or (2) John emphasized the support he received, while the author added the interpretation about the importance of mentorship.

The LLM is also asked to determine whether each instance of ambiguity can be resolved using the question and the context. The standard for resolution is whether a group of readers would likely agree on the correct interpretation. For example, recall the sentence ``\textit{AI has advanced renewable energy and sustainable agriculture at Company A and Company B.}'' If the context specified that Company A builds solar panels and Company B reduces farms’ water usage, readers would likely conclude that AI has advanced renewable energy at Company A and sustainable agriculture at Company B. Conversely, if the context only described both companies as ``\textit{environmental pioneers,}'' readers would have insufficient information to determine the correct interpretation. 

If any ambiguity is unresolvable, the sentence is labeled ``Cannot be disambiguated'' and excluded from the Decomposition stage (\autoref{subsec:decomposition}), even if it has unambiguous, verifiable components. \autoref{tab:disambiguation_examples} in \appautoref{app:claimify} provides examples of such sentences. If the LLM resolves all ambiguity, it returns a clarified version of the sentence. If there is no ambiguity, it returns the original sentence.\footnote{The LLM also checks for partial names, abbreviations, and acronyms, which are not considered linguistic ambiguities. If full forms are provided in the question or context, the LLM includes them in the returned sentence; otherwise, the LLM leaves them unchanged to avoid factual inaccuracies.} 

Across the models tested in our experiments (\autoref{sec:experiments}), the largest proportion of sentences labeled ``Cannot be disambiguated'' was 5.4\%.\footnote{The proportion of sentences labeled ``Cannot be disambiguated'' per model was as follows: \texttt{mistral-large-2411} = 5.4\%, \texttt{gpt-4o-2024-08-06} = 3.2\%, \texttt{DeepSeek-V3} = 2.4\%.}

\subsection{Decomposition}
\label{subsec:decomposition}
In the final stage, Claimify uses an LLM to decompose each disambiguated sentence into decontextualized factual claims. If it does not return any claims (only 0.8\% of cases in our experiments), the sentence is labeled ``No verifiable claims.''

Extracted claims may include text in brackets, which typically represents information implied by the question or context but not explicitly stated in the source sentence. For example, given the question ``\textit{Provide an overview of celebrities’ stances on the Middle East,}'' and the sentence ``\textit{John has called for peace,}'' Claimify may return the claim ``\textit{John }[\textit{a celebrity}] \textit{has called for peace} [\textit{in the Middle East}].'' This notation resembles the ``markup-and-mask'' approach by \citet{eisenstein:2024}, which adds bracketed text to clarify context in passages. A benefit of bracketing is that it flags inferred content, which is inherently less reliable than content explicitly stated in the source sentence.

\section{Experimental Setup}
\subsection{Data}
\subsubsection{BingCheck}
\label{subsubsec:bingcheck}
We evaluated Claimify’s performance on the BingCheck dataset \citep{li:2024}, which consists of 396 answers generated by Microsoft Copilot (formerly Bing Chat). BingCheck spans a wide range of topics and question types, and its answers are significantly longer than those in comparable datasets \citep{li:2024}. As a result, it reflects the diversity and complexity of real-world LLM usage in long-form question answering. Moreover, since BingCheck answers are generated based on web search results, it is reasonable to expect that relevant evidence exists for many claims – a key consideration for the evidence retrieval step of the decontextualization evaluation described in \autoref{subsec:decontext}.

\subsubsection{Human Annotation Study}
\label{subsubsec:annotation}

We conducted a human annotation study to classify sentences in BingCheck answers as containing or not containing factual claims.\footnote{The dataset will be released at \url{https://aka.ms/claimify-dataset}.} A total of 6,490 sentences were labeled by three annotators who are familiar with natural language processing, including one of the authors. To ensure reliability, annotators completed two practice rounds on a subset of sentences, resolving disagreements via discussion, then independently annotated the remaining data. Krippendorff’s alpha \citep{krippendorff:2013, castro:2017} increased from 0.44 in the first practice round to 0.72 in the final round, reaching 0.86 for high-confidence annotations. Sentence splitting methodology, annotation procedure, guidelines, and results are detailed in \appautoref{app:annotation}. The labels from the study were used in our analysis of coverage (\autoref{subsec:coverage_results}). 

\subsection{Baseline Methods}
\label{subsec:methods}
We compared Claimify to five LLM-based methods:
\begin{enumerate}
    \item \textbf{AFaCTA \citep{ni:2024}} uses an ensemble of prompts to classify sentences as containing or not containing objectively verifiable content.
    \item \textbf{Factcheck-GPT \citep{wang:2024}} classifies sentences as factual claims, opinions, non-claims (e.g., questions or imperative statements), or other.
    \item \textbf{VeriScore \citep{song:2024}} combines sentence classification, decomposition, and decontextualization in a single prompt. It returns either ``No verifiable claim'' or a list of claims.
    \item \textbf{DnD \citep{wanner:2024:dnd}} decomposes and decontextualizes sentences in a single prompt.
    \item \textbf{SAFE \citep{wei:2024}} adds instructions to FActScore's decomposition prompt \citep{min:2023} and performs decontextualization in a separate prompt.
\end{enumerate}

DnD and SAFE do not provide instructions for handling sentences without factual claims. Therefore, when the LLM declined to extract claims, it did not use a consistent output format. If no claims were parsed from the output, we assumed that the LLM determined there were no factual claims.

We selected these methods because they allow for direct comparisons with Claimify. They process sentences independently, unlike other methods that analyze the entire answer as a single unit (e.g., \citealp{chern:2023, bayat:2025}). The methods with explicit sentence classification components (AFaCTA, Factcheck-GPT, VeriScore) share Claimify’s focus on detecting verifiable content, rather than ranking sentences by their ``check-worthiness'' (see \autoref{sec:related_work}). The methods that perform claim extraction (VeriScore, DnD, SAFE) involve both decomposition and decontextualization, unlike other approaches that focus solely on decomposition (e.g., \citealp{kamoi:2023, chen:2023:felm}).

To further enable direct comparisons, we used the sentence splitting logic described in \appautoref{app:ann_sentences} for all methods. We also made minimal edits to all prompts (except VeriScore, where edits were unnecessary) to include the question and clarify that the sentence was extracted from a response to the question. Additional settings are described in \appautoref{app:hyperparameters}.

The claim extraction methods (VeriScore, DnD, SAFE, Claimify) generated a total of 73,681 claims. Where a method produced duplicate claims for a sentence, we removed the duplicates, resulting in 73,229 claims. All subsequent sections refer to this de-duplicated claim set. 

\section{Experiments}
\label{sec:experiments}
This section describes our implementation of the evaluation framework outlined in \autoref{sec:framework} and the corresponding results.~\appautoref{app:eval_prompts} contains all prompts. \appautoref{app:method_contexts} provides the sentence context definitions, and \appautoref{app:exp_samples} describes the samples for all experiments. All results in this section were produced using OpenAI’s \texttt{gpt-4o-2024-08-06} model with a temperature of 0; results for other models are reported in \appautoref{app:additional_models}. All reported p-values were derived from two-proportion Z-tests, with Holm-Bonferroni correction for multiple comparisons.   

\subsection{Entailment}
\label{subsec:entailment}

To determine whether claims are entailed by their source sentences, we first used a pre-trained Natural Language Inference (NLI) model from \citet{nie:2020}, as was done by \citet{wanner:2024:decomp}. We tried two configurations, both of which revealed significant limitations, detailed in \appautoref{app:nli}. 

In light of the NLI model's limitations, we developed a prompt that classifies a claim as entailed or not entailed based on the source sentence, context, and question. To validate the prompt, we randomly sampled 20 claims from each claim extraction method (80 claims total) and labeled them without referencing the LLM’s outputs. The LLM’s classifications conflicted with our labels in only five cases, whereas the NLI model’s classifications conflicted with us in 32 and 12 cases for the first and second configurations, respectively. \appautoref{app:entailment_review} provides an overview of cases where we disagreed. 

\autoref{tab:entailment} shows the percentage of entailed claims for each method using our prompt. Claimify and VeriScore achieved the highest percentage of entailed claims (99\%), with no statistically significant difference between them (p=0.145). All pairwise comparisons between the methods, except for Claimify vs. VeriScore, showed statistically significant differences (p<0.001). These results align with a similar analysis by \citet{wanner:2024:decomp} where the percentage of supported claims for various claim extraction methods, averaged across different models, ranged from 86\% to 98\%. 

\begin{table}[H]
\centering
\renewcommand{\arraystretch}{1.3}
\begin{tabular}{ccc}
    \hline
    \textbf{Method} & \textbf{Claims} & \textbf{\% Entailed} \\
    \hline
    Claimify  & 12,406  & 99.0 \\
    DnD       & 27,717  & 89.1 \\
    SAFE      & 22,786  & 96.6 \\
    VeriScore & 7,420   & 99.2 \\
    \hline
\end{tabular}
\caption{Percentage of claims entailed by the combined source sentence, context, and question, along with the total number of claims (as described in \appautoref{app:filtering_statements}), per method.}
\label{tab:entailment}
\end{table} 

\subsection{Coverage}
\label{subsec:coverage_results}

To evaluate sentence-level coverage (defined in \autoref{subsec:coverage}), we used the results of the human annotation study (\autoref{subsubsec:annotation}) as ground truth. We refer to the 63\% of sentences labeled as containing a factual claim as ``verifiable'' and the remaining sentences as ``unverifiable.''

\subsubsection{Sentence-Level Coverage}
\label{subsubsec:sentence_coverage}
For VeriScore, DnD, SAFE, and Claimify, we assigned a ``verifiable'' label if at least one claim was extracted. For Factcheck-GPT, we treated only the ``factual claim'' label as ``verifiable.'' For AFaCTA, we replicated its majority voting procedure, treating ``contains objective information'' as the ``verifiable'' label. 

\autoref{tab:coverage} shows the sentence-level coverage results for all methods under the ``Sent.'' columns. Claimify achieved the highest accuracy (91.8\%) and macro $F_1$ score (91.2\%), followed by AFaCTA (accuracy = 81.6\%) and VeriScore (macro $F_1$ = 78.9\%). In other words, Claimify was the most effective at identifying whether a sentence contains at least one factual claim.

\begin{table*}[ht]
    \centering
    \renewcommand{\arraystretch}{1.1}  
    \setlength{\tabcolsep}{4.5pt}        

    \adjustbox{max width=\textwidth}{  
    \begin{tabular}{c|cccc|cccccccc}  
        \toprule
        \multirow{2}{*}{\textbf{Method}} & 
        \multicolumn{2}{c}{\textbf{Accuracy}} & 
        \multicolumn{2}{c|}{\textbf{Macro F$_1$}} & 
        \multicolumn{2}{c}{\textbf{Precision\textsubscript{$V$}}} & 
        \multicolumn{2}{c}{\textbf{Recall\textsubscript{$V$}}} & 
        \multicolumn{2}{c}{\textbf{Precision\textsubscript{$UV$}}} & 
        \multicolumn{2}{c}{\textbf{Recall\textsubscript{$UV$}}} \\
        \cmidrule(lr){2-3} \cmidrule(lr){4-5}
        \cmidrule(lr){6-7} \cmidrule(lr){8-9}
        \cmidrule(lr){10-11} \cmidrule(lr){12-13}
        & Sent. & Elem. & Sent. & Elem. & Sent. & Elem. & Sent. & Elem. & Sent. & Elem. & Sent. & Elem. \\
        \midrule
        Claimify       & \textbf{91.8} & \textbf{87.9} & \textbf{91.2} & \textbf{83.7} & 93.2 & 96.7 & 93.9 & 87.6 & 89.5 & \textbf{65.6} & 88.3 & 88.8 \\
        DnD            & 63.7 & 76.9 & 41.4 & 56.2 & 63.5 & 81.2 & \textbf{99.6} & \textbf{92.2} & 79.7 & 39.9 & 2.7  & 19.5 \\
        SAFE           & 65.0 & 74.6 & 45.1 & 57.3 & 64.3 & 81.7 & 99.5 & 87.4 & 88.2 & 35.6 & 6.5  & 26.2 \\
        VeriScore      & 79.0 & 64.7 & 78.9 & 62.5 & \textbf{98.2} & \textbf{98.6} & 67.8 & 56.1 & 64.2 & 37.0 & \textbf{97.9} & \textbf{96.9} \\
        AFaCTA         & 81.6 & –     & 78.7 & –     & 79.9 & –     & 94.5 & –     & 86.5 & –     & 59.8 & –    \\
        Factcheck-GPT  & 81.5 & –     & 78.0 & –     & 79.0 & –     & 96.1 & –     & \textbf{89.6} & –     & 56.5 & –    \\
        \bottomrule
    \end{tabular}
    }  

    \caption{Sentence- and element-level coverage metrics (\%), with class-specific precision and recall for verifiable (V) and unverifiable (UV) sentences (Sent.) and elements (Elem.). Since AFaCTA and Factcheck-GPT only determine whether a sentence contains a factual claim without extracting claims, element-level measures are not applicable. Bolded values represent the highest score in each column.}
    \label{tab:coverage}
\end{table*}

\subsubsection{Element-Level Coverage}
\label{subsubsec:element_coverage}
To evaluate element-level coverage of a sentence $s$ by the claims $\mathcal{C}=\{c_i\}_{i=1}^{n}$ extracted from $s$, we developed two prompts: one identifies and classifies elements of $s$ as verifiable or unverifiable, and the other determines if each element is covered by $\mathcal{C}$ and labels the coverage as explicit or implicit.

To compare coverage across methods, we used a single set of elements per sentence. To ensure consistency with the sentence-level coverage results, we analyzed 81\% of sentences where the element-based verifiability labels matched the annotation study labels (i.e., we included sentences with at least one verifiable element that were deemed ``verifiable'' in the annotation study, as well as sentences with no verifiable elements that were deemed ``unverifiable'' in the annotation study). 

\autoref{tab:coverage} shows the element-level results under the ``Elem.'' columns. Claimify achieved the highest accuracy (87.9\%) and macro $F_1$ score (83.7\%), followed by DnD (accuracy = 76.9\%) and VeriScore (macro $F_1$ = 62.5\%). In other words, the claims extracted by Claimify achieved the best balance between including verifiable content and excluding unverifiable content from the source text.

To validate the results, we manually reviewed a random sample of 80 sentences, assessing element quality and coverage labels. We found that 95\% of sentences met all quality criteria, and we agreed with 97\% of coverage labels. The evaluation criteria and results are detailed in \appautoref{app:coverage_review}.

\subsection{Decontextualization}
\label{subsec:decontext_results}
We evaluated decontextualization as follows (see \autoref{subsec:decontext}, \appautoref{app:filtering_statements}, and \appautoref{app:decontext_details} for details):
\begin{enumerate}
    \item \textbf{Identify Missing Context.} We developed a prompt that either returns $c_{\max}$, a maximally decontextualized version of a claim $c$, or indicates that $c$ is already maximally decontextualized. We  reviewed 80 outputs and found that 76 (95\%) were valid (see \autoref{app:decontext_review}).

    \item \textbf{Retrieve Evidence.} To assess consistency of results across retrieval systems, we replicated two configurations from prior works:
    \begin{itemize}
        \item \textbf{Google Search \citep{wei:2024}:} An LLM generates an initial query based on a claim, retrieves results from the Google Search API, and iteratively refines the query. In total, five queries are generated, with the top three results per query forming the evidence set.
        \item \textbf{Bing \citep{li:2024}:} An LLM generates a single query based on a claim, with the top three results from the Bing Web Search API forming the evidence set.
    \end{itemize}

    \item \textbf{Determine Veracity.} To assess whether a claim is supported by the retrieved evidence, we used the verification prompt from \citet{wei:2024}, as it demonstrated strong agreement with human annotators. If the queries from Step 2 above did not return any search results, we classified the claim as not supported.

    \end{enumerate}

\begin{table*}[hb]
    \centering
    \renewcommand{\arraystretch}{1.15}
    \setlength{\tabcolsep}{4.5pt}

    \adjustbox{max width=\textwidth}{
    \begin{tabular}{
        >{\centering\arraybackslash}m{2.5cm}
        cccccccccccccccc
    }
        \toprule
        \multirow{2}{*}{\textbf{Method}} & 
        \multirow{2}{*}{\textbf{1*}} & 
        \multicolumn{2}{c}{\textbf{2*}} & 
        \multicolumn{2}{c}{\textbf{3}} & 
        \multicolumn{2}{c}{\textbf{4*}} & 
        \multicolumn{2}{c}{\textbf{5}} & 
        \multicolumn{2}{c}{\textbf{6}} & 
        \multicolumn{2}{c}{\textbf{7*}} & 
        \multicolumn{2}{c}{\textbf{Desirable*}} \\
        \cmidrule(lr){3-4} \cmidrule(lr){5-6} \cmidrule(lr){7-8}  
        \cmidrule(lr){9-10} \cmidrule(lr){11-12} \cmidrule(lr){13-14}  
        \cmidrule(lr){15-16}
        & & G & B & G & B & G & B & G & B & G & B & G & B & G & B \\
        \midrule
        Claimify    & \textbf{16.3} & 47.6 & 47.7 & \underline{7.9} & \underline{7.0} & 5.8 & 6.2 & \underline{6.5} & 7.5 & 5.0 & 5.0 & 10.8 & 10.4 & \textbf{80.6} & \textbf{80.5} \\
        DnD         & 12.9 & 48.2 & 48.3 & 8.6 & 7.9 & \textbf{6.2} & 6.1 & 7.1 & 7.9 & 5.9 & 5.6 & \textbf{11.2} & \textbf{11.4} & 78.4 & 78.6 \\
        SAFE        & 10.4 & \textbf{51.2} & \textbf{51.7} & 9.2 & 8.4 & 5.9 & 6.3 & 7.0 & \underline{7.4} & 5.5 & 5.6 & 10.7 & 10.3 & 78.2 & 78.7 \\
        VeriScore   & 13.2 & 50.0 & 51.3 & 9.8 & 8.1 & 5.3 & \textbf{6.5} & 7.4 & 8.2 & \underline{4.5} & \underline{4.4} & 9.8 & 8.4 & 78.3 & 79.3 \\
        \bottomrule
    \end{tabular}
    }

    \caption{Percentage distribution of decontextualization result types 1-7 (as defined in \autoref{subsec:decontext}), per method. ``G'' and ``B'' refer to the Google Search and Bing configurations for evidence retrieval, respectively. Percentages may not total 100 within each configuration due to rounding. Only result types 1, 2, 4, and 7 are considered desirable, denoted by *. The ``Desirable'' column sums the desirable results. For desirable results, bolded values indicate the highest score per column; for undesirable results, underlined values indicate the lowest scores. }
    \label{tab:decontext}
\end{table*}

\autoref{tab:decontext} shows the distribution of the seven result types (\autoref{subsec:decontext}) per claim extraction method for both retrieval configurations (Google Search and Bing). We ensured that identical claims extracted by different methods from the same sentence were assigned the same result type. Result 1 ($c = c_{\max}$, i.e., no missing contextual information) is reported only once, since the same $c_{\max}$ was used for both configurations. 

Claimify had the largest percentage of Result 1 cases, significantly higher than all other methods (p<0.001). Across both retrieval configurations, Claimify achieved the largest percentage of desirable results (i.e., types 1, 2, 4, and 7 from \autoref{subsec:decontext}). For Google Search, Claimify significantly outperformed all other methods (p<0.001). For Bing, Claimify also outperformed other methods (p<0.001), except VeriScore, where the difference was not statistically significant (p=0.159).

\section{Analysis}
\label{sec:analysis}

To assess which stages of Claimify contribute most to its performance gains, we tested three variants: (1) removing the Selection stage, (2) using the Selection stage only to detect the presence of a factual claim without rewriting sentences to exclude unverifiable information (\autoref{subsec:selection}), and (3) removing the Disambiguation stage. 

Results are shown in \autoref{tab:claimify_ablation}. The complete Claimify system outperformed all variants on entailment and element-level coverage (p<0.001). For decontextualization, no pairwise differences between Claimify and the variants were statistically significant (p>0.05). Removing the Selection stage caused the largest performance drop, indicating the benefit of checking verifiability prior to extracting claims. Notably, the Claimify variants matched or outperformed most baseline methods (see \autoref{sec:experiments}), suggesting that simplified versions of Claimify may still be effective. 

\begin{table*}[ht]
    \centering
    \renewcommand{\arraystretch}{1.1}
    \setlength{\tabcolsep}{6pt}
    \begin{tabular}{cccc}
        \toprule
        \textbf{Method} & \textbf{Entailment} & \textbf{Element-Level Coverage} & \textbf{Decontextualization} \\
        \midrule
        Claimify & \textbf{99.0} & \textbf{83.7} & 80.5 \\
        Claimify Without Selection & 98.0 & 54.4 & \textbf{81.1} \\
        Claimify With Selection as Detector & 97.7 & 74.7 & 80.2 \\
        Claimify Without Disambiguation & 98.3 & 75.9 & 80.9 \\
        \bottomrule
    \end{tabular}
    \caption{Performance of Claimify variants. ``Entailment'' is the percentage of entailed claims. ``Element-Level Coverage'' is the macro $F_1$ score as a percentage. ``Decontextualization'' is the percentage of desirable result types (as defined in \autoref{subsec:decontext}) with Bing as the retriever. Bolded values indicate the highest score per column.}
    \label{tab:claimify_ablation}
\end{table*}

\section{Related Work}
\label{sec:related_work}

\textbf{Selection.} While Claimify focuses on identifying verifiable factual claims, many prior works aim to identify ``check-worthy'' claims (\citealp{gencheva:2017, jaradat:2018, arslan:2020}). Check-worthiness criteria include public interest \citep{hassan:2017}, potential harm \citep{nakov:2022}, and relevance to a topic, such as the environment \citep{stammbach:2023}. We agree with \citet{konstantinovskiy:2021} and \citet{ni:2024} that check-worthiness is subjective.  

\noindent\textbf{Disambiguation.} A key feature of Claimify is its ability to identify when ambiguity is present and cannot be resolved. Existing decomposition and decontextualization methods either ignore ambiguity or assume it is always resolvable. An example of the latter is Molecular Facts \citep{gunjal:2024}, a decontextualization method focused on cases where the main entity in a sentence could refer to multiple people (which Claimify would classify as referential ambiguity). Molecular Facts not only forces the LLM to resolve such ambiguities, but it also relies on the model’s parametric knowledge – rather than the sentence and its context – which risks introducing factual inaccuracies. Beyond claim extraction, prior works on ambiguity in fact-checking have explored ambiguous questions \citep{min:2020, kim:2023, zhang:2024} and investigated why annotators disagree on veracity judgements \citep{glockner:2024}.

\noindent\textbf{Decomposition.} Claimify uses an LLM to extract claims as complete declarative sentences. Alternative decomposition approaches include extracting subject-predicate-object tuples \citep{banko:2007, goodrich:2019, hu:2024:refchecker}, predicate-argument structures \citep{white:2016, zhang:2017, goyal:2020}, questions \citep{fan:2020, chen:2022}, and subsets of tokens \citep{chen:2023:propsegment}.

\section{Conclusion}

In this paper, we propose an evaluation framework for claim extraction in the context of fact-checking, based on entailment, coverage, and decontextualization. We provide automated, scalable, and replicable methods for applying the framework. For coverage, we augment sentence-level assessment with a more granular element-level approach that accounts for sentences containing both verifiable and unverifiable content. For decontextualization, we propose a novel method that quantifies the impact of omitted context on factuality verdicts.

We also introduce Claimify, an LLM-based claim extraction method. Unlike existing methods, Claimify explicitly accounts for ambiguity: it identifies cases where the source text has multiple plausible interpretations and the correct interpretation cannot be inferred from the context. We benchmarked Claimify against existing methods and found that across all models tested:
\begin{itemize}
    \item At least 95\% of claims extracted by Claimify were entailed, outperforming all methods with \texttt{mistral-large-2411}, and tying with one method when using \texttt{gpt-4o-2024-08-06} and \texttt{DeepSeek-V3};
    \item For both sentence- and element-level coverage, Claimify achieved the highest accuracy and macro $F_1$ scores; and
    \item Claimify was least likely to omit contextual information critical to the factuality verdict.
\end{itemize}

\section{Limitations}

\textbf{Dataset Scope.} We evaluated performance on a single dataset, albeit one that includes diverse question types and spans a wide range of domains. Future work could extend the analysis to additional datasets and explore how Claimify generalizes beyond long-form LLM-generated texts to other content types, such as political speeches \citep{ni:2024} and social media \citep{alam:2021}.

\noindent\textbf{Annotator Pool.} The annotation study involved three annotators due to limited availability of high-quality annotators. While all samples were labeled by multiple annotators, a larger annotator pool would increase the reliability of the results.

\noindent\textbf{Hyperparameter Configuration.} We did not conduct an exhaustive search for the optimal hyperparameter configuration for Claimify (\appautoref{app:hyperparameters}). For example, varying the number of completions and the minimum success threshold could yield valuable insights. Additionally, we anticipate that increasing the number of preceding sentences used as context may improve performance, especially for answers that contain lengthy bullet-point lists. Consider the following list item: ``\textit{- Investing in renewable energy sources.}'' Is it a recommendation for what one ought to do (not verifiable) or an example of an action a specific entity has taken (verifiable)? The correct interpretation is likely clarified by the preamble for the list (e.g., \textit{``Here are some steps businesses should take to mitigate their environmental impact:''}), but it might not have been included in our narrow context window.

\noindent\textbf{Evaluating Disambiguation.} Claimify’s Disambiguation stage (\autoref{subsec:disambiguation}) addresses two types of ambiguity that we identified as particularly relevant to claim extraction. We encourage future work to explore additional types of ambiguity and to develop methods for evaluating detection accuracy.

\noindent\textbf{Temporal Ambiguity.} While Claimify can identify temporal phrases with multiple interpretations (e.g., ``\textit{next year}'') as a type of referential ambiguity, a subtler challenge is the absence of temporal information. For example, in the sentence ``\textit{The unemployment rate decreased in California,}'' the relevant time period is unspecified, even though it is likely critical for fact-checking. However, Claimify may not flag this sentence as ``Cannot be disambiguated'' since it does not contain any phrases with multiple interpretations. It would be ideal if all claims included a temporal qualifier, as assumed by \citeauthor{rashkin:2023}'s (\citeyear{rashkin:2023}) definition of standalone propositions. In reality, many texts do not contain any temporal markers. Labeling all such cases as ``Cannot be disambiguated'' would severely limit the utility of claim extraction. We encourage future work to explore strategies for handling missing temporal context. 

\section{Ethics Statement}

\textbf{Licenses and Terms of Use.} We used BingCheck, a publicly available dataset released for research purposes, although \citet{li:2024} do not specify a license. For method replication, we complied with all provided licenses and terms of use. Factcheck-GPT, SAFE, and VeriScore are released under Apache 2.0. AFaCTA has a publicly available code repository but does not specify a license. DnD was replicated based on the methodology described in the arXiv publication. We adhered to the terms of use for the Bing Web Search API and the Serper Google Search API (\appautoref{app:decontext_details}).

\noindent\textbf{Human Annotation.} We used human-annotated data to evaluate claim extraction. Annotators were informed about the task and provided consent. No personally identifiable or sensitive information was collected or used.

\noindent\textbf{Potential Risks.} Claim extraction and fact-checking involve subjective judgments, which may introduce biases. To mitigate this risk, we propose structured, explainable, and replicable evaluation methods. Additionally, claim extraction systems can introduce factual inaccuracies or misinterpret the original text. We address these risks through our entailment evaluation and Claimify’s Disambiguation stage. Finally, even though our proposed evaluation framework and Claimify are both fully automated, we recommend human oversight in high-stakes contexts where inaccuracies or misinterpretations could have significant consequences.

\noindent\textbf{Use of AI Assistants.} We used ChatGPT for minor language refinement and proofreading, but all substantive contributions were developed independently.

\bibliography{custom}

\begin{thebibliography}{50}
\providecommand{\natexlab}[1]{#1}

\bibitem[{Alam et~al.(2021)Alam, Shaar, Dalvi, Sajjad, Nikolov, Mubarak, Da~San~Martino, Abdelali, Durrani, Darwish, Al-Homaid, Zaghouani, Caselli, Danoe, Stolk, Bruntink, and Nakov}]{alam:2021}
Firoj Alam, Shaden Shaar, Fahim Dalvi, Hassan Sajjad, Alex Nikolov, Hamdy Mubarak, Giovanni Da~San~Martino, Ahmed Abdelali, Nadir Durrani, Kareem Darwish, Abdulaziz Al-Homaid, Wajdi Zaghouani, Tommaso Caselli, Gijs Danoe, Friso Stolk, Britt Bruntink, and Preslav Nakov. 2021.
\newblock \href {https://doi.org/10.18653/v1/2021.findings-emnlp.56} {Fighting the {COVID}-19 infodemic: Modeling the perspective of journalists, fact-checkers, social media platforms, policy makers, and the society}.
\newblock In \emph{Findings of the Association for Computational Linguistics: EMNLP 2021}, pages 611--649, Punta Cana, Dominican Republic. Association for Computational Linguistics.

\bibitem[{Arslan et~al.(2020)Arslan, Hassan, Li, and Tremayne}]{arslan:2020}
Fatma Arslan, Naeemul Hassan, Chengkai Li, and Mark Tremayne. 2020.
\newblock \href {https://doi.org/10.1609/icwsm.v14i1.7346} {A benchmark dataset of check-worthy factual claims}.
\newblock \emph{Proceedings of the International AAAI Conference on Web and Social Media}, 14(1):821--829.

\bibitem[{Banko et~al.(2007)Banko, Cafarella, Soderland, Broadhead, and Etzioni}]{banko:2007}
Michele Banko, Michael~J. Cafarella, Stephen Soderland, Matt Broadhead, and Oren Etzioni. 2007.
\newblock Open information extraction from the web.
\newblock In \emph{Proceedings of the 20th International Joint Conference on Artifical Intelligence}, IJCAI'07, page 2670–2676, San Francisco, CA, USA. Morgan Kaufmann Publishers Inc.

\bibitem[{Bayat et~al.(2025)Bayat, Zhang, Munir, and Wang}]{bayat:2025}
Farima~Fatahi Bayat, Lechen Zhang, Sheza Munir, and Lu~Wang. 2025.
\newblock \href {https://arxiv.org/abs/2410.22257} {Factbench: A dynamic benchmark for in-the-wild language model factuality evaluation}.
\newblock \emph{Preprint}, arXiv:2410.22257.

\bibitem[{Bird and Loper(2004)}]{bird:2004}
Steven Bird and Edward Loper. 2004.
\newblock \href {https://aclanthology.org/P04-3031/} {{NLTK}: The natural language toolkit}.
\newblock In \emph{Proceedings of the {ACL} Interactive Poster and Demonstration Sessions}, pages 214--217, Barcelona, Spain. Association for Computational Linguistics.

\bibitem[{Castro(2017)}]{castro:2017}
Santiago Castro. 2017.
\newblock Fast {K}rippendorff: Fast computation of {K}rippendorff's alpha agreement measure.
\newblock \url{https://github.com/pln-fing-udelar/fast-krippendorff}.

\bibitem[{Chen et~al.(2022)Chen, Sriram, Choi, and Durrett}]{chen:2022}
Jifan Chen, Aniruddh Sriram, Eunsol Choi, and Greg Durrett. 2022.
\newblock \href {https://doi.org/10.18653/v1/2022.emnlp-main.229} {Generating literal and implied subquestions to fact-check complex claims}.
\newblock In \emph{Proceedings of the 2022 Conference on Empirical Methods in Natural Language Processing}, pages 3495--3516, Abu Dhabi, United Arab Emirates. Association for Computational Linguistics.

\bibitem[{Chen et~al.(2023{\natexlab{a}})Chen, Zhao, Zhang, Chern, Gao, Liu, and He}]{chen:2023:felm}
Shiqi Chen, Yiran Zhao, Jinghan Zhang, I-Chun Chern, Siyang Gao, Pengfei Liu, and Junxian He. 2023{\natexlab{a}}.
\newblock Felm: benchmarking factuality evaluation of large language models.
\newblock In \emph{Proceedings of the 37th International Conference on Neural Information Processing Systems}, NIPS '23, Red Hook, NY, USA. Curran Associates Inc.

\bibitem[{Chen et~al.(2023{\natexlab{b}})Chen, Buthpitiya, Fabrikant, Roth, and Schuster}]{chen:2023:propsegment}
Sihao Chen, Senaka Buthpitiya, Alex Fabrikant, Dan Roth, and Tal Schuster. 2023{\natexlab{b}}.
\newblock \href {https://doi.org/10.18653/v1/2023.findings-acl.565} {{P}rop{S}egm{E}nt: A large-scale corpus for proposition-level segmentation and entailment recognition}.
\newblock In \emph{Findings of the Association for Computational Linguistics: ACL 2023}, pages 8874--8893, Toronto, Canada. Association for Computational Linguistics.

\bibitem[{Chen et~al.(2024)Chen, Wang, Chen, Yu, Ma, Zhao, Zhang, and Yu}]{chen:2024}
Tong Chen, Hongwei Wang, Sihao Chen, Wenhao Yu, Kaixin Ma, Xinran Zhao, Hongming Zhang, and Dong Yu. 2024.
\newblock \href {https://doi.org/10.18653/v1/2024.emnlp-main.845} {Dense {X} retrieval: What retrieval granularity should we use?}
\newblock In \emph{Proceedings of the 2024 Conference on Empirical Methods in Natural Language Processing}, pages 15159--15177, Miami, Florida, USA. Association for Computational Linguistics.

\bibitem[{Chern et~al.(2023)Chern, Chern, Chen, Yuan, Feng, Zhou, He, Neubig, and Liu}]{chern:2023}
I-Chun Chern, Steffi Chern, Shiqi Chen, Weizhe Yuan, Kehua Feng, Chunting Zhou, Junxian He, Graham Neubig, and Pengfei Liu. 2023.
\newblock \href {https://arxiv.org/abs/2307.13528} {Factool: Factuality detection in generative ai -- a tool augmented framework for multi-task and multi-domain scenarios}.
\newblock \emph{Preprint}, arXiv:2307.13528.

\bibitem[{Choi et~al.(2021)Choi, Palomaki, Lamm, Kwiatkowski, Das, and Collins}]{choi:2021}
Eunsol Choi, Jennimaria Palomaki, Matthew Lamm, Tom Kwiatkowski, Dipanjan Das, and Michael Collins. 2021.
\newblock \href {https://doi.org/10.1162/tacl_a_00377} {Decontextualization: Making sentences stand-alone}.
\newblock \emph{Transactions of the Association for Computational Linguistics}, 9:447--461.

\bibitem[{Daxenberger et~al.(2017)Daxenberger, Eger, Habernal, Stab, and Gurevych}]{daxenberger:2017}
Johannes Daxenberger, Steffen Eger, Ivan Habernal, Christian Stab, and Iryna Gurevych. 2017.
\newblock \href {https://doi.org/10.18653/v1/D17-1218} {What is the essence of a claim? cross-domain claim identification}.
\newblock In \emph{Proceedings of the 2017 Conference on Empirical Methods in Natural Language Processing}, pages 2055--2066, Copenhagen, Denmark. Association for Computational Linguistics.

\bibitem[{Deng et~al.(2024)Deng, Schlichtkrull, and Vlachos}]{deng:2024}
Zhenyun Deng, Michael Schlichtkrull, and Andreas Vlachos. 2024.
\newblock \href {https://doi.org/10.18653/v1/2024.acl-long.645} {Document-level claim extraction and decontextualisation for fact-checking}.
\newblock In \emph{Proceedings of the 62nd Annual Meeting of the Association for Computational Linguistics (Volume 1: Long Papers)}, pages 11943--11954, Bangkok, Thailand. Association for Computational Linguistics.

\bibitem[{Eisenstein et~al.(2024)Eisenstein, Andor, Bohnet, Collins, and Mimno}]{eisenstein:2024}
Jacob Eisenstein, Daniel Andor, Bernd Bohnet, Michael Collins, and David Mimno. 2024.
\newblock \href {https://arxiv.org/abs/2210.02498} {Honest students from untrusted teachers: Learning an interpretable question-answering pipeline from a pretrained language model}.
\newblock \emph{Preprint}, arXiv:2210.02498.

\bibitem[{Fan et~al.(2020)Fan, Piktus, Petroni, Wenzek, Saeidi, Vlachos, Bordes, and Riedel}]{fan:2020}
Angela Fan, Aleksandra Piktus, Fabio Petroni, Guillaume Wenzek, Marzieh Saeidi, Andreas Vlachos, Antoine Bordes, and Sebastian Riedel. 2020.
\newblock \href {https://doi.org/10.18653/v1/2020.emnlp-main.580} {Generating fact checking briefs}.
\newblock In \emph{Proceedings of the 2020 Conference on Empirical Methods in Natural Language Processing (EMNLP)}, pages 7147--7161, Online. Association for Computational Linguistics.

\bibitem[{Gencheva et~al.(2017)Gencheva, Nakov, M{\`a}rquez, Barr{\'o}n-Cede{\~n}o, and Koychev}]{gencheva:2017}
Pepa Gencheva, Preslav Nakov, Llu{\'i}s M{\`a}rquez, Alberto Barr{\'o}n-Cede{\~n}o, and Ivan Koychev. 2017.
\newblock \href {https://doi.org/10.26615/978-954-452-049-6_037} {A context-aware approach for detecting worth-checking claims in political debates}.
\newblock In \emph{Proceedings of the International Conference Recent Advances in Natural Language Processing, {RANLP} 2017}, pages 267--276, Varna, Bulgaria. INCOMA Ltd.

\bibitem[{Glockner et~al.(2024)Glockner, Stali{\={u}}nait{\.{e}}, Thorne, Vallejo, Vlachos, and Gurevych}]{glockner:2024}
Max Glockner, Ieva Stali{\={u}}nait{\.{e}}, James Thorne, Gisela Vallejo, Andreas Vlachos, and Iryna Gurevych. 2024.
\newblock \href {https://doi.org/10.1162/tacl_a_00629} {{A}mbi{FC}: Fact-checking ambiguous claims with evidence}.
\newblock \emph{Transactions of the Association for Computational Linguistics}, 12:1--18.

\bibitem[{Goodrich et~al.(2019)Goodrich, Rao, Liu, and Saleh}]{goodrich:2019}
Ben Goodrich, Vinay Rao, Peter~J. Liu, and Mohammad Saleh. 2019.
\newblock \href {https://doi.org/10.1145/3292500.3330955} {Assessing the factual accuracy of generated text}.
\newblock In \emph{Proceedings of the 25th ACM SIGKDD International Conference on Knowledge Discovery \& Data Mining}, KDD '19, page 166–175, New York, NY, USA. Association for Computing Machinery.

\bibitem[{Goyal and Durrett(2020)}]{goyal:2020}
Tanya Goyal and Greg Durrett. 2020.
\newblock \href {https://doi.org/10.18653/v1/2020.findings-emnlp.322} {Evaluating factuality in generation with dependency-level entailment}.
\newblock In \emph{Findings of the Association for Computational Linguistics: EMNLP 2020}, pages 3592--3603, Online. Association for Computational Linguistics.

\bibitem[{Gunjal and Durrett(2024)}]{gunjal:2024}
Anisha Gunjal and Greg Durrett. 2024.
\newblock \href {https://doi.org/10.18653/v1/2024.findings-emnlp.215} {Molecular facts: Desiderata for decontextualization in {LLM} fact verification}.
\newblock In \emph{Findings of the Association for Computational Linguistics: EMNLP 2024}, pages 3751--3768, Miami, Florida, USA. Association for Computational Linguistics.

\bibitem[{Hassan et~al.(2017)Hassan, Arslan, Li, and Tremayne}]{hassan:2017}
Naeemul Hassan, Fatma Arslan, Chengkai Li, and Mark Tremayne. 2017.
\newblock \href {https://doi.org/10.1145/3097983.3098131} {Toward automated fact-checking: Detecting check-worthy factual claims by claimbuster}.
\newblock In \emph{Proceedings of the 23rd ACM SIGKDD International Conference on Knowledge Discovery and Data Mining}, KDD '17, page 1803–1812, New York, NY, USA. Association for Computing Machinery.

\bibitem[{Hu et~al.(2024{\natexlab{a}})Hu, Long, and Wang}]{hu:2024:decompdilemmas}
Qisheng Hu, Quanyu Long, and Wenya Wang. 2024{\natexlab{a}}.
\newblock \href {https://arxiv.org/abs/2411.02400} {Decomposition dilemmas: Does claim decomposition boost or burden fact-checking performance?}
\newblock \emph{Preprint}, arXiv:2411.02400.

\bibitem[{Hu et~al.(2024{\natexlab{b}})Hu, Ru, Qiu, Guo, Zhang, Xu, Luo, Liu, Zhang, and Zhang}]{hu:2024:refchecker}
Xiangkun Hu, Dongyu Ru, Lin Qiu, Qipeng Guo, Tianhang Zhang, Yang Xu, Yun Luo, Pengfei Liu, Yue Zhang, and Zheng Zhang. 2024{\natexlab{b}}.
\newblock \href {https://doi.org/10.18653/v1/2024.emnlp-main.395} {Knowledge-centric hallucination detection}.
\newblock In \emph{Proceedings of the 2024 Conference on Empirical Methods in Natural Language Processing}, pages 6953--6975, Miami, Florida, USA. Association for Computational Linguistics.

\bibitem[{Huang et~al.(2025)Huang, Yu, Ma, Zhong, Feng, Wang, Chen, Peng, Feng, Qin, and Liu}]{huang:2025}
Lei Huang, Weijiang Yu, Weitao Ma, Weihong Zhong, Zhangyin Feng, Haotian Wang, Qianglong Chen, Weihua Peng, Xiaocheng Feng, Bing Qin, and Ting Liu. 2025.
\newblock \href {https://doi.org/10.1145/3703155} {A survey on hallucination in large language models: Principles, taxonomy, challenges, and open questions}.
\newblock \emph{ACM Trans. Inf. Syst.}, 43(2).

\bibitem[{Jaradat et~al.(2018)Jaradat, Gencheva, Barr{\'o}n-Cede{\~n}o, M{\`a}rquez, and Nakov}]{jaradat:2018}
Israa Jaradat, Pepa Gencheva, Alberto Barr{\'o}n-Cede{\~n}o, Llu{\'i}s M{\`a}rquez, and Preslav Nakov. 2018.
\newblock \href {https://doi.org/10.18653/v1/N18-5006} {{C}laim{R}ank: Detecting check-worthy claims in {A}rabic and {E}nglish}.
\newblock In \emph{Proceedings of the 2018 Conference of the North {A}merican Chapter of the Association for Computational Linguistics: Demonstrations}, pages 26--30, New Orleans, Louisiana. Association for Computational Linguistics.

\bibitem[{Kamoi et~al.(2023)Kamoi, Goyal, Diego~Rodriguez, and Durrett}]{kamoi:2023}
Ryo Kamoi, Tanya Goyal, Juan Diego~Rodriguez, and Greg Durrett. 2023.
\newblock \href {https://doi.org/10.18653/v1/2023.emnlp-main.470} {{W}i{CE}: Real-world entailment for claims in {W}ikipedia}.
\newblock In \emph{Proceedings of the 2023 Conference on Empirical Methods in Natural Language Processing}, pages 7561--7583, Singapore. Association for Computational Linguistics.

\bibitem[{Kane and Schubert(2023)}]{kane:2023}
Benjamin Kane and Lenhart Schubert. 2023.
\newblock \href {https://doi.org/10.18653/v1/2023.emnlp-main.678} {We are what we repeatedly do: Inducing and deploying habitual schemas in persona-based responses}.
\newblock In \emph{Proceedings of the 2023 Conference on Empirical Methods in Natural Language Processing}, pages 10998--11016, Singapore. Association for Computational Linguistics.

\bibitem[{Kim et~al.(2023)Kim, Kim, Jeon, Park, and Kang}]{kim:2023}
Gangwoo Kim, Sungdong Kim, Byeongguk Jeon, Joonsuk Park, and Jaewoo Kang. 2023.
\newblock \href {https://doi.org/10.18653/v1/2023.emnlp-main.63} {Tree of clarifications: Answering ambiguous questions with retrieval-augmented large language models}.
\newblock In \emph{Proceedings of the 2023 Conference on Empirical Methods in Natural Language Processing}, pages 996--1009, Singapore. Association for Computational Linguistics.

\bibitem[{Konstantinovskiy et~al.(2021)Konstantinovskiy, Price, Babakar, and Zubiaga}]{konstantinovskiy:2021}
Lev Konstantinovskiy, Oliver Price, Mevan Babakar, and Arkaitz Zubiaga. 2021.
\newblock \href {https://doi.org/10.1145/3412869} {Toward automated factchecking: Developing an annotation schema and benchmark for consistent automated claim detection}.
\newblock \emph{Digital Threats}, 2(2).

\bibitem[{Krippendorff(2013)}]{krippendorff:2013}
K.~Krippendorff. 2013.
\newblock \href {https://books.google.com/books?id=s_yqFXnGgjQC} {\emph{Content Analysis: An Introduction to Its Methodology}}.
\newblock SAGE Publications.

\bibitem[{Li et~al.(2024)Li, Peng, Galley, Gao, and Zhang}]{li:2024}
Miaoran Li, Baolin Peng, Michel Galley, Jianfeng Gao, and Zhu Zhang. 2024.
\newblock \href {https://doi.org/10.18653/v1/2024.findings-naacl.12} {Self-checker: Plug-and-play modules for fact-checking with large language models}.
\newblock In \emph{Findings of the Association for Computational Linguistics: NAACL 2024}, pages 163--181, Mexico City, Mexico. Association for Computational Linguistics.

\bibitem[{Majer and {\v{S}}najder(2024)}]{majer:2024}
Laura Majer and Jan {\v{S}}najder. 2024.
\newblock \href {https://doi.org/10.18653/v1/2024.fever-1.27} {Claim check-worthiness detection: How well do {LLM}s grasp annotation guidelines?}
\newblock In \emph{Proceedings of the Seventh Fact Extraction and VERification Workshop (FEVER)}, pages 245--263, Miami, Florida, USA. Association for Computational Linguistics.

\bibitem[{Min et~al.(2023)Min, Krishna, Lyu, Lewis, Yih, Koh, Iyyer, Zettlemoyer, and Hajishirzi}]{min:2023}
Sewon Min, Kalpesh Krishna, Xinxi Lyu, Mike Lewis, Wen-tau Yih, Pang Koh, Mohit Iyyer, Luke Zettlemoyer, and Hannaneh Hajishirzi. 2023.
\newblock \href {https://doi.org/10.18653/v1/2023.emnlp-main.741} {{FA}ct{S}core: Fine-grained atomic evaluation of factual precision in long form text generation}.
\newblock In \emph{Proceedings of the 2023 Conference on Empirical Methods in Natural Language Processing}, pages 12076--12100, Singapore. Association for Computational Linguistics.

\bibitem[{Min et~al.(2020)Min, Michael, Hajishirzi, and Zettlemoyer}]{min:2020}
Sewon Min, Julian Michael, Hannaneh Hajishirzi, and Luke Zettlemoyer. 2020.
\newblock \href {https://doi.org/10.18653/v1/2020.emnlp-main.466} {{A}mbig{QA}: Answering ambiguous open-domain questions}.
\newblock In \emph{Proceedings of the 2020 Conference on Empirical Methods in Natural Language Processing (EMNLP)}, pages 5783--5797, Online. Association for Computational Linguistics.

\bibitem[{Nakov et~al.(2022)Nakov, Barr\'{o}n-Cede\~{n}o, Da~San~Martino, Alam, Stru\ss{}, Mandl, M\'{\i}guez, Caselli, Kutlu, Zaghouani, Li, Shaar, Shahi, Mubarak, Nikolov, Babulkov, Kartal, and Beltr\'{a}n}]{nakov:2022}
Preslav Nakov, Alberto Barr\'{o}n-Cede\~{n}o, Giovanni Da~San~Martino, Firoj Alam, Julia~Maria Stru\ss{}, Thomas Mandl, Rub\'{e}n M\'{\i}guez, Tommaso Caselli, Mucahid Kutlu, Wajdi Zaghouani, Chengkai Li, Shaden Shaar, Gautam~Kishore Shahi, Hamdy Mubarak, Alex Nikolov, Nikolay Babulkov, Yavuz~Selim Kartal, and Javier Beltr\'{a}n. 2022.
\newblock \href {https://doi.org/10.1007/978-3-030-99739-7_52} {The clef-2022 checkthat! lab on fighting the covid-19 infodemic and fake news detection}.
\newblock In \emph{Advances in Information Retrieval: 44th European Conference on IR Research, ECIR 2022, Stavanger, Norway, April 10–14, 2022, Proceedings, Part II}, page 416–428, Berlin, Heidelberg. Springer-Verlag.

\bibitem[{Ni et~al.(2024)Ni, Shi, Stammbach, Sachan, Ash, and Leippold}]{ni:2024}
Jingwei Ni, Minjing Shi, Dominik Stammbach, Mrinmaya Sachan, Elliott Ash, and Markus Leippold. 2024.
\newblock \href {https://doi.org/10.18653/v1/2024.acl-long.104} {{AF}a{CTA}: Assisting the annotation of factual claim detection with reliable {LLM} annotators}.
\newblock In \emph{Proceedings of the 62nd Annual Meeting of the Association for Computational Linguistics (Volume 1: Long Papers)}, pages 1890--1912, Bangkok, Thailand. Association for Computational Linguistics.

\bibitem[{Nie et~al.(2020)Nie, Williams, Dinan, Bansal, Weston, and Kiela}]{nie:2020}
Yixin Nie, Adina Williams, Emily Dinan, Mohit Bansal, Jason Weston, and Douwe Kiela. 2020.
\newblock \href {https://doi.org/10.18653/v1/2020.acl-main.441} {Adversarial {NLI}: A new benchmark for natural language understanding}.
\newblock In \emph{Proceedings of the 58th Annual Meeting of the Association for Computational Linguistics}, pages 4885--4901, Online. Association for Computational Linguistics.

\bibitem[{Rashkin et~al.(2023)Rashkin, Nikolaev, Lamm, Aroyo, Collins, Das, Petrov, Tomar, Turc, and Reitter}]{rashkin:2023}
Hannah Rashkin, Vitaly Nikolaev, Matthew Lamm, Lora Aroyo, Michael Collins, Dipanjan Das, Slav Petrov, Gaurav~Singh Tomar, Iulia Turc, and David Reitter. 2023.
\newblock \href {https://doi.org/10.1162/coli_a_00486} {Measuring attribution in natural language generation models}.
\newblock \emph{Computational Linguistics}, 49(4):777--840.

\bibitem[{Song et~al.(2024)Song, Kim, and Iyyer}]{song:2024}
Yixiao Song, Yekyung Kim, and Mohit Iyyer. 2024.
\newblock \href {https://doi.org/10.18653/v1/2024.findings-emnlp.552} {{V}eri{S}core: Evaluating the factuality of verifiable claims in long-form text generation}.
\newblock In \emph{Findings of the Association for Computational Linguistics: EMNLP 2024}, pages 9447--9474, Miami, Florida, USA. Association for Computational Linguistics.

\bibitem[{Stammbach et~al.(2023)Stammbach, Webersinke, Bingler, Kraus, and Leippold}]{stammbach:2023}
Dominik Stammbach, Nicolas Webersinke, Julia Bingler, Mathias Kraus, and Markus Leippold. 2023.
\newblock \href {https://doi.org/10.18653/v1/2023.acl-short.91} {Environmental claim detection}.
\newblock In \emph{Proceedings of the 61st Annual Meeting of the Association for Computational Linguistics (Volume 2: Short Papers)}, pages 1051--1066, Toronto, Canada. Association for Computational Linguistics.

\bibitem[{Tang et~al.(2024)Tang, Laban, and Durrett}]{tang:2024}
Liyan Tang, Philippe Laban, and Greg Durrett. 2024.
\newblock \href {https://doi.org/10.18653/v1/2024.emnlp-main.499} {{M}ini{C}heck: Efficient fact-checking of {LLM}s on grounding documents}.
\newblock In \emph{Proceedings of the 2024 Conference on Empirical Methods in Natural Language Processing}, pages 8818--8847, Miami, Florida, USA. Association for Computational Linguistics.

\bibitem[{Wang et~al.(2024)Wang, Gangi~Reddy, Mujahid, Arora, Rubashevskii, Geng, Mohammed~Afzal, Pan, Borenstein, Pillai, Augenstein, Gurevych, and Nakov}]{wang:2024}
Yuxia Wang, Revanth Gangi~Reddy, Zain~Muhammad Mujahid, Arnav Arora, Aleksandr Rubashevskii, Jiahui Geng, Osama Mohammed~Afzal, Liangming Pan, Nadav Borenstein, Aditya Pillai, Isabelle Augenstein, Iryna Gurevych, and Preslav Nakov. 2024.
\newblock \href {https://doi.org/10.18653/v1/2024.findings-emnlp.830} {Factcheck-bench: Fine-grained evaluation benchmark for automatic fact-checkers}.
\newblock In \emph{Findings of the Association for Computational Linguistics: EMNLP 2024}, pages 14199--14230, Miami, Florida, USA. Association for Computational Linguistics.

\bibitem[{Wanner et~al.(2024{\natexlab{a}})Wanner, Durme, and Dredze}]{wanner:2024:dnd}
Miriam Wanner, Benjamin~Van Durme, and Mark Dredze. 2024{\natexlab{a}}.
\newblock \href {https://arxiv.org/abs/2412.13175} {Dndscore: Decontextualization and decomposition for factuality verification in long-form text generation}.
\newblock \emph{Preprint}, arXiv:2412.13175.

\bibitem[{Wanner et~al.(2024{\natexlab{b}})Wanner, Ebner, Jiang, Dredze, and Van~Durme}]{wanner:2024:decomp}
Miriam Wanner, Seth Ebner, Zhengping Jiang, Mark Dredze, and Benjamin Van~Durme. 2024{\natexlab{b}}.
\newblock \href {https://doi.org/10.18653/v1/2024.starsem-1.13} {A closer look at claim decomposition}.
\newblock In \emph{Proceedings of the 13th Joint Conference on Lexical and Computational Semantics (*SEM 2024)}, pages 153--175, Mexico City, Mexico. Association for Computational Linguistics.

\bibitem[{Wei et~al.(2024)Wei, Yang, Song, Lu, Hu, Huang, Tran, Peng, Liu, Huang, Du, and Le}]{wei:2024}
Jerry Wei, Chengrun Yang, Xinying Song, Yifeng Lu, Nathan Hu, Jie Huang, Dustin Tran, Daiyi Peng, Ruibo Liu, Da~Huang, Cosmo Du, and Quoc~V. Le. 2024.
\newblock \href {https://arxiv.org/abs/2403.18802} {Long-form factuality in large language models}.

\bibitem[{White et~al.(2016)White, Reisinger, Sakaguchi, Vieira, Zhang, Rudinger, Rawlins, and {Van Durme}}]{white:2016}
Aaron~Steven White, Drew Reisinger, Keisuke Sakaguchi, Tim Vieira, Sheng Zhang, Rachel Rudinger, Kyle Rawlins, and Benjamin {Van Durme}. 2016.
\newblock \href {https://aclweb.org/anthology/D16-1177} {{Universal Decompositional Semantics on Universal Dependencies}}.
\newblock In \emph{Proceedings of the 2016 Conference on Empirical Methods in Natural Language Processing}, pages 1713--1723, Austin, Texas. Association for Computational Linguistics.

\bibitem[{Wright et~al.(2022)Wright, Wadden, Lo, Kuehl, Cohan, Augenstein, and Wang}]{wright:2022}
Dustin Wright, David Wadden, Kyle Lo, Bailey Kuehl, Arman Cohan, Isabelle Augenstein, and Lucy~Lu Wang. 2022.
\newblock \href {https://doi.org/10.18653/v1/2022.acl-long.175} {Generating scientific claims for zero-shot scientific fact checking}.
\newblock In \emph{Proceedings of the 60th Annual Meeting of the Association for Computational Linguistics (Volume 1: Long Papers)}, pages 2448--2460, Dublin, Ireland. Association for Computational Linguistics.

\bibitem[{Zhang et~al.(2017)Zhang, Rudinger, and {Van Durme}}]{zhang:2017}
Sheng Zhang, Rachel Rudinger, and Ben {Van Durme}. 2017.
\newblock {An Evaluation of PredPatt and Open IE via Stage 1 Semantic Role Labeling}.
\newblock In \emph{Proceedings of the 12th International Conference on Computational Semantics (IWCS)}, Montpellier, France.

\bibitem[{Zhang et~al.(2024)Zhang, Qin, Deng, Huang, Lei, Liu, Jin, Liang, and Chua}]{zhang:2024}
Tong Zhang, Peixin Qin, Yang Deng, Chen Huang, Wenqiang Lei, Junhong Liu, Dingnan Jin, Hongru Liang, and Tat-Seng Chua. 2024.
\newblock \href {https://doi.org/10.18653/v1/2024.acl-long.578} {{CLAMBER}: A benchmark of identifying and clarifying ambiguous information needs in large language models}.
\newblock In \emph{Proceedings of the 62nd Annual Meeting of the Association for Computational Linguistics (Volume 1: Long Papers)}, pages 10746--10766, Bangkok, Thailand. Association for Computational Linguistics.

\end{thebibliography}
\appendix
\newpage
\clearpage
\section{Explicit vs. Implicit Coverage of Unverifiable Elements}
\label{app:coverage_example}

Consider the following sentence: ``\textit{After drug X was approved, patient survival rates tripled, highlighting the power of modern medicine.}'' It has two elements:
\begin{enumerate}
    \item After drug X was approved, patient survival rates tripled.
    \item The tripling of patient survival rates highlights the power of modern medicine.
\end{enumerate}

Suppose we are evaluating two claim extraction methods:
\begin{itemize}
    \item Method A returns [\textit{``Patient survival rates tripled after drug X was approved''}]
    \item Method B returns [\textit{``Patient survival rates tripled after drug X was approved'', ``Drug X’s tripling of patient survival rates highlights the power of modern medicine''}]
\end{itemize}

Element 1 is verifiable, so we want to cover it. Method A and B both explicitly cover element 1, so they both have a true positive. 

Element 2 is not verifiable, so we do not want to cover it. Method B explicitly covers element 2, so it has a false positive. One might argue that the claim extracted by Method A implies element 2. If we counted the claim as a false positive, then Methods A and B would have the same score for element-level coverage as both have 1 true positive and 1 false positive. However, the same score would be unfair: Method A is superior to Method B since only the latter explicitly included the unverifiable element. Therefore, we score element 2 as a true negative for Method A.

\section{Claimify Overview and Examples}
\label{app:claimify}

\begin{figure*}[p]
    \centering
    \includegraphics[width=\textwidth]{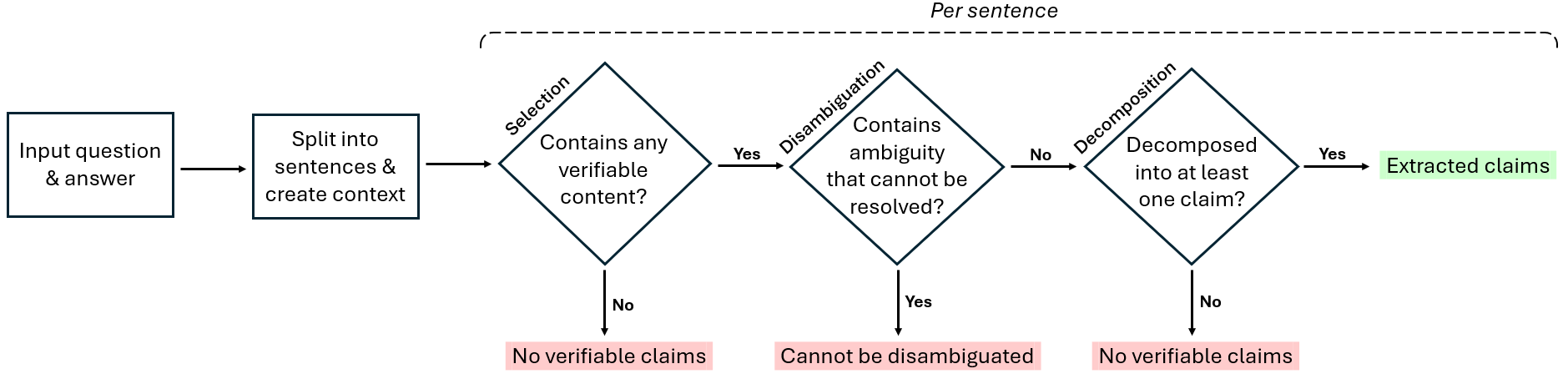}
    \caption{Claimify stages}
    \label{fig:claimify}
\end{figure*}

\autoref{fig:claimify} provides an overview of Claimify’s stages. \autoref{tab:selection_examples} contains examples of outputs from Claimify’s Selection stage. \autoref{tab:disambiguation_examples} contains examples of sentences labeled by Claimify as ``Cannot be disambiguated.'' 

\begin{table*}[ht]
    \centering
    \renewcommand{\arraystretch}{1.3}
    \setlength{\tabcolsep}{8pt}
    \adjustbox{max width=\textwidth}{
    \begin{tabular}{
        >{\centering\arraybackslash}p{5.8cm} 
        >{\centering\arraybackslash}p{4.5cm} 
        >{\centering\arraybackslash}p{6.2cm}
    }
        \toprule
        \textbf{Sentence} & \textbf{Output} & \textbf{Reasoning} \\
        \midrule

        \multicolumn{1}{>{\justifying\noindent\arraybackslash}p{5.8cm}}{This could revolutionize transplantation, wound healing, organ function and lifespan.} 
        & 
        \multicolumn{1}{>{\justifying\noindent\arraybackslash}p{4.5cm}}{Sentence does not contain any verifiable content} 
        & 
        \multicolumn{1}{>{\justifying\noindent\arraybackslash}p{6.2cm}}{The sentence uses the word ``could,'' indicating a potential or speculative outcome rather than a specific and verifiable proposition. It suggests possible future impacts of regenerative medicine but does not assert a current fact or event that can be verified.} \\
        \midrule

        \multicolumn{1}{>{\justifying\noindent\arraybackslash}p{5.8cm}}{When you reach your destination, try to use public transportation or walk/bike instead of renting a car or taking a taxi.} 
        & 
        \multicolumn{1}{>{\justifying\noindent\arraybackslash}p{4.5cm}}{Sentence does not contain any verifiable content} 
        & 
        \multicolumn{1}{>{\justifying\noindent\arraybackslash}p{6.2cm}}{The sentence is a recommendation, suggesting an action to take when reaching a destination. It does not contain a specific and verifiable proposition because it is advising a course of action rather than stating a fact or a verifiable event.} \\
        \midrule

        \multicolumn{1}{>{\justifying\noindent\arraybackslash}p{5.8cm}}{There have been many archaeological discoveries in the 21st century that have changed or challenged our understanding of human history.} 
        & 
        \multicolumn{1}{>{\justifying\noindent\arraybackslash}p{4.5cm}}{Sentence does not contain any verifiable content} 
        & 
        \multicolumn{1}{>{\justifying\noindent\arraybackslash}p{6.2cm}}{The sentence is broad and general, stating that many discoveries have changed or challenged our understanding, without specifying which discoveries or how they have done so. It serves as an introduction to the specific examples that follow, rather than providing a specific and verifiable proposition itself.} \\
        \midrule

        \multicolumn{1}{>{\justifying\noindent\arraybackslash}p{5.8cm}}{The explosion can spin the neutron star to mind blowing speeds — up to 600 rotations per second.} 
        & 
        \multicolumn{1}{>{\justifying\noindent\arraybackslash}p{4.5cm}}{The explosion can spin the neutron star to speeds up to 600 rotations per second.} 
        & 
        \multicolumn{1}{>{\justifying\noindent\arraybackslash}p{6.2cm}}{The phrase ``mind blowing speeds'' is subjective, but the specific claim of ``up to 600 rotations per second'' is verifiable.} \\
        \midrule

        \multicolumn{1}{>{\justifying\noindent\arraybackslash}p{5.8cm}}{However, some of these organisms are also threatened by human activities, such as logging, mining, grazing and climate change, and may not survive much longer without conservation efforts.} 
        & 
        \multicolumn{1}{>{\justifying\noindent\arraybackslash}p{4.5cm}}{Some of these organisms are threatened by human activities, such as logging, mining, grazing, and climate change.} 
        & 
        \multicolumn{1}{>{\justifying\noindent\arraybackslash}p{6.2cm}}{The sentence contains specific threats to organisms from human activities, which is a verifiable proposition. However, the part about the organisms potentially not surviving much longer without conservation efforts is speculative and not verifiable.} \\
        \midrule

        \multicolumn{1}{>{\justifying\noindent\arraybackslash}p{5.8cm}}{Islamic calligraphy developed various styles and scripts over time and across regions, reflecting the diversity and creativity of Islamic civilization.} 
        & 
        \multicolumn{1}{>{\justifying\noindent\arraybackslash}p{4.5cm}}{Islamic calligraphy developed various styles and scripts over time and across regions.} 
        & 
        \multicolumn{1}{>{\justifying\noindent\arraybackslash}p{6.2cm}}{The sentence makes a claim about the development of Islamic calligraphy, stating that it developed various styles and scripts over time and across regions. This is a specific claim that can be verified by historical and cultural research into the evolution of Islamic calligraphy. The latter part of the sentence, ``reflecting the diversity and creativity of Islamic civilization,'' is more interpretative and broad.} \\
        \bottomrule
    \end{tabular}
    }

    \caption{Examples of sentences where Claimify's Selection stage either (1) determined there is no verifiable content or (2) provided a revised version containing only verifiable content. Excerpts from the model's reasoning are also provided.}
    \label{tab:selection_examples}
\end{table*}
\begin{table*}[ht]
    \centering
    \renewcommand{\arraystretch}{1.3}  
    \setlength{\tabcolsep}{8pt}        

    \adjustbox{max width=\textwidth}{  
    \begin{tabular}{
        >{\centering\arraybackslash}p{7cm}   
        >{\centering\arraybackslash}p{9cm}   
    }
        \toprule
        \multicolumn{1}{c}{\textbf{Sentence}} & \multicolumn{1}{c}{\textbf{Reasoning}} \\  
        \midrule

        \multicolumn{1}{>{\justifying\noindent\arraybackslash}p{7cm}}{- The development of hydrogen and biofuels as alternative fuels for transportation and industry.} 
        & 
        \multicolumn{1}{>{\justifying\noindent\arraybackslash}p{9cm}}{\textbf{Structural Ambiguity}: The sentence could be interpreted as: (1) hydrogen and biofuels are being developed as alternative fuels for both transportation and industry, (2) hydrogen is being developed as an alternative fuel for transportation, and biofuels are being developed as an alternative fuel for industry.} \\
        \midrule

        \multicolumn{1}{>{\justifying\noindent\arraybackslash}p{7cm}}{- The announcement of the winning project by Ryo Taniguchi on 28 February 2018.} 
        & 
        \multicolumn{1}{>{\justifying\noindent\arraybackslash}p{9cm}}{\textbf{Structural Ambiguity}: The sentence could be interpreted as: (1) Ryo Taniguchi announced the winning project on 28 February 2018, (2) The winning project, created by Ryo Taniguchi, was announced on 28 February 2018.} \\
        \midrule

        \multicolumn{1}{>{\justifying\noindent\arraybackslash}p{7cm}}{According to CNN, solar power is one of the best potential solutions to the climate crisis, as it does not emit greenhouse gas or air pollution, and it could dominate the US electricity grid as early as 10 years from now.} 
        & 
        \multicolumn{1}{>{\justifying\noindent\arraybackslash}p{9cm}}{\textbf{Referential Ambiguity}: The phrase ``as early as 10 years from now'' is temporally ambiguous... There is no indication of the current year in the question or context.\newline \textbf{Structural Ambiguity}: The sentence could be interpreted as: (1) CNN claims that solar power is one of the best potential solutions to the climate crisis because it does not emit greenhouse gas or air pollution, and CNN also claims that solar power could dominate the US electricity grid as early as 10 years from now, (2) CNN claims that solar power is one of the best potential solutions to the climate crisis because it does not emit greenhouse gas or air pollution, while the claim that solar power could dominate the US electricity grid as early as 10 years from now is attributable to the writer, not to CNN.} \\
        \midrule

        \multicolumn{1}{>{\justifying\noindent\arraybackslash}p{7cm}}{- The development of **quantum mechanics** and **electron shell** theory by Niels Bohr, Erwin Schrödinger, Wolfgang Pauli, Linus Pauling, and others in the early 20th century.} 
        & 
        \multicolumn{1}{>{\justifying\noindent\arraybackslash}p{9cm}}{\textbf{Structural Ambiguity}: The sentence could be interpreted as: (1) Niels Bohr, Erwin Schrödinger, Wolfgang Pauli, Linus Pauling, and others developed both quantum mechanics and electron shell theory, (2) some of these individuals contributed to quantum mechanics while others contributed to electron shell theory.} \\
        \midrule

        \multicolumn{1}{>{\justifying\noindent\arraybackslash}p{7cm}}{- **Small modular nuclear reactors**: Nuclear power is a carbon-free source of electricity that can provide baseload power regardless of weather conditions.} 
        & 
        \multicolumn{1}{>{\justifying\noindent\arraybackslash}p{9cm}}{\textbf{Structural Ambiguity}: The sentence could be interpreted as: (1) Small modular nuclear reactors are a type of nuclear power that is a carbon-free source of electricity and can provide baseload power regardless of weather conditions, (2) Nuclear power in general is a carbon-free source of electricity and can provide baseload power regardless of weather conditions, with small modular nuclear reactors being an example of this.} \\
        \midrule

        \multicolumn{1}{>{\justifying\noindent\arraybackslash}p{7cm}}{- Using circular polybags that can be recycled into new polybags, such as those developed by **Cadel Deinking** and tested by **Adidas, Kering, and PVH**.} 
        & 
        \multicolumn{1}{>{\justifying\noindent\arraybackslash}p{9cm}}{\textbf{Structural Ambiguity}: The sentence could be interpreted as: (1) Cadel Deinking developed circular polybags, and Adidas, Kering, and PVH tested these specific polybags, (2) Cadel Deinking developed a type of circular polybag, and Adidas, Kering, and PVH tested circular polybags in general, not necessarily those developed by Cadel Deinking.} \\
        \bottomrule
    \end{tabular}
    }

    \caption{Examples of ambiguous sentences where Claimify found multiple possible interpretations and determined that the context and question did not clearly indicate a single correct interpretation. Excerpts from the model's reasoning are also provided.}
    \label{tab:disambiguation_examples}
\end{table*}

\section{Human Annotation Study}
\label{app:annotation}

\subsection{Sentence Splitting}
\label{app:ann_sentences}

To identify sentence boundaries in BingCheck answers, we first divided answers into paragraphs by splitting on newline characters. Then, for each paragraph, we applied Claimify’s sentence splitting methodology (described in \autoref{subsec:claimify_sentences}). Splitting by newlines was necessary because many answers contained bullet-point lists with items that lacked terminal punctuation, which would otherwise be treated as a single sentence by the NLTK tokenizer. This process produced 6,490 sentences. 

\subsection{Procedure}
The annotation team consisted of one of the authors and two members of the authors’ research group who are familiar with natural language processing but were not involved in the creation of Claimify or the writing of this paper.

Annotators reviewed question-answer pairs and labeled sentences as either containing or not containing a factual claim, distinguishing between high and low confidence labels. Detailed annotation guidelines are provided in \appautoref{app:ann_guidelines}, and an example of the annotation interface in Azure Machine Learning is provided in \appautoref{app:ann_interface}.

From the 396 BingCheck answers, 18 were randomly sampled as practice cases and divided into two rounds of nine samples each. Annotators independently labeled the first round, resolved disagreements through discussion, and repeated the process for the second round. The remaining 378 answers were split into three groups of 126, and each annotator was assigned two groups (252 answers) to ensure that every sample was independently annotated by two people.

\subsection{Results}
We measured inter-annotator agreement using Krippendorff’s alpha \citep{krippendorff:2013, castro:2017}. As expected, agreement improved across rounds, increasing from 0.44 in the first practice round to 0.54 in the second, and reaching 0.72 in the final round. Notably, for 82\% of sentences in the final round where both annotators reported high confidence, Krippendorff's alpha was 0.86.

For sentences where all annotators agreed on the label, the consensus was used as the ground truth. In the practice rounds, disagreements were resolved through discussions among the annotators. In the final round, disagreements within the two sample groups where the author was an annotator were settled by prioritizing the author’s label; in the third sample group, the author reviewed and resolved disagreements.

\clearpage

\onecolumn
\subsection{Guidelines}
\label{app:ann_guidelines}

Annotators were given the following guidelines:
\begin{tcolorbox}[
    breakable,                    
    colback=white,                
    colframe=black,              
    title=Annotation Guidelines,       
    title after break=Annotation Guidelines (Continued),
    fonttitle=\bfseries, 
    coltext=black,
]
\begin{lstlisting}[breaklines=true, breakindent=0pt, basicstyle=\small\ttfamily\raggedright, xleftmargin=-5pt, frame=none, xrightmargin=-5pt, aboveskip=-2pt, belowskip=-2pt]
## Overview
You will be given a set of question-answer pairs. The answers were generated by an LLM, based on some search results.

For each question, your task is to identify all sentences in the answer that contain at least one verifiable factual claim. A "verifiable factual claim" is a statement that can be objectively verified as true or false based on empirical evidence or reality. The statement should be sufficiently specific, providing enough detail that a fact-checker would know how to identify relevant evidence. 

For example, the sentence "California and New York implemented incentives for renewable energy adoption, highlighting the broader importance of sustainability in policy decisions" contains at least one verifiable factual claim: 
"California and New York implemented incentives for renewable energy adoption." 
(Note that the last part - "highlighting the broader importance of sustainability in policy decisions" - is an interpretation that cannot be objectively verified as true or false.)

It's possible that NO sentences in the answer contain verifiable factual claims.
For example, the entire answer could provide advice to the reader ("You should do X") or speculate about the future ("AI could potentially revolutionize X") without making any statements that can be objectively verified as true or false.

## Key Guidelines
- You are NOT being asked to determine whether the sentence is true or false, or to check whether evidence exists to confirm or refute the information in a sentence. We are only interested in whether the sentence has the potential to be objectively verified. 
- You should NOT consider whether the sentence is relevant to the question.
- Some sentences in the answer may have citations (e.g., [^2^]). Do NOT consider the presence or absence of a citation when deciding whether the sentence contains a verifiable factual claim.
- If the sentence is about the LLM's inability to answer the question (e.g., 
"The search results did not find any indication of X" or "I'm sorry, I'm unable to respond to this question"), it does NOT contain a verifiable factual claim.
- It is extremely important that you consider the context for a sentence, i.e., the preceding and following sentences. If a sentence is a high-level introduction for the following sentences, or a high-level conclusion for the preceding sentences, then it usually does NOT contain a verifiable factual 
claim.	
    - For example, if a sentence is "Climate change has had several significant economic effects, such as:" and it's followed by a list of specific examples of economic effects, then the sentence is merely an introduction and does NOT contain a verifiable factual claim.
    - For each paragraph in the answer: it is highly recommended that you read through the entire paragraph first without making any decisions, then consider each sentence individually.

## Examples
Here are some examples of sentences that do NOT contain any verifiable factual claims: 
- By prioritizing ethical considerations, companies can ensure that their innovations are not only groundbreaking but also socially responsible -> generic statement that cannot be objectively verified as true or false
- Technological progress should be inclusive -> opinion
- Leveraging AI is essential for maximizing productivity -> opinion
- Networking events can be crucial in shaping the paths of young entrepreneurs and providing them with valuable connections -> opinion
- AI could lead to advancements in healthcare -> speculation
- This implies that John Smith is a courageous person -> interpretation
- Try to show appreciation to your friends -> advice/recommendation
- Basketball is a fun, dynamic game, and an important part of many people's lives -> opinion and generic

As you can see from these examples, unverifiable claims can often be described as broad or generic statements, opinions, interpretations, speculations, and/or advice.

Here are some examples of sentences that do contain at least one verifiable factual claim:
- The partnership between Company X and Company Y illustrates the power of innovation -> a verifiable factual claim would be "there is a partnership between Company X and Company Y"; the rest (the partnership illustrates the power of innovation) is an unverifiable interpretation
- Jane Doe's approach of embracing adaptability and prioritizing customer feedback can be valuable advice for new executives -> a verifiable factual claim would be "Jane Doe's approach includes embracing adaptability and prioritizing customer feedback"; the rest (her approach can be valuable advice) is an opinion
- Smith's advocacy for renewable energy is crucial in addressing these challenges -> "Smith advocates for renewable energy"
- **John Smith**: instrumental in numerous renewable energy initiatives, playing a pivotal role in Project Green -> "John Smith is involved in renewable energy initiatives and played a role in Project Green"
- John, the CEO of Company X, is a notable example of strong leadership -> "John is the CEO of Company X"
- Therefore, leveraging industry events, as demonstrated by Jane's experience at the Tech Networking Club, can provide visibility and traction for new ventures -> "Jane had an experience at the Tech Networking Club"

You'll notice that in some of the above examples, only part of the sentence - not the entire sentence - contains a verifiable factual claim. It is NOT necessary for the entire sentence to convey a verifiable factual claim.

## How to Add Tags
In the annotation tool, you will have 3 options available to you: 
1. The "[HIGH CONF] Contains" tag - use this when you have high confidence that the sentence contains at least one verifiable factual claim
2. The "[LOW CONF] Lean towards contains" tag - use this when you have low confidence in the appropriate classification for the sentence, but you lean towards it containing at least one verifiable factual claim
3. The "[LOW CONF] Lean against contains" tag - use this when you have low confidence in the appropriate classification for the sentence, but you lean towards it NOT containing any verifiable factual claims

Important: 
- If you have high confidence that a sentence does NOT contain any verifiable factual claims, leave it untagged.
- If you have high confidence that NO sentences in the answer contain verifiable factual claims (i.e., you didn't assign any of the above tags to any sentences), you should use any tag on the QUESTION part of the text. We don't actually want to annotate the question, but we're doing this because the annotation tool will not let you proceed to the next answer if no text is tagged. 
\end{lstlisting}
\end{tcolorbox}

\twocolumn

\subsection{Interface}
\label{app:ann_interface}

\begin{figure*}[ht]
    \centering
    \includegraphics[width=\textwidth]{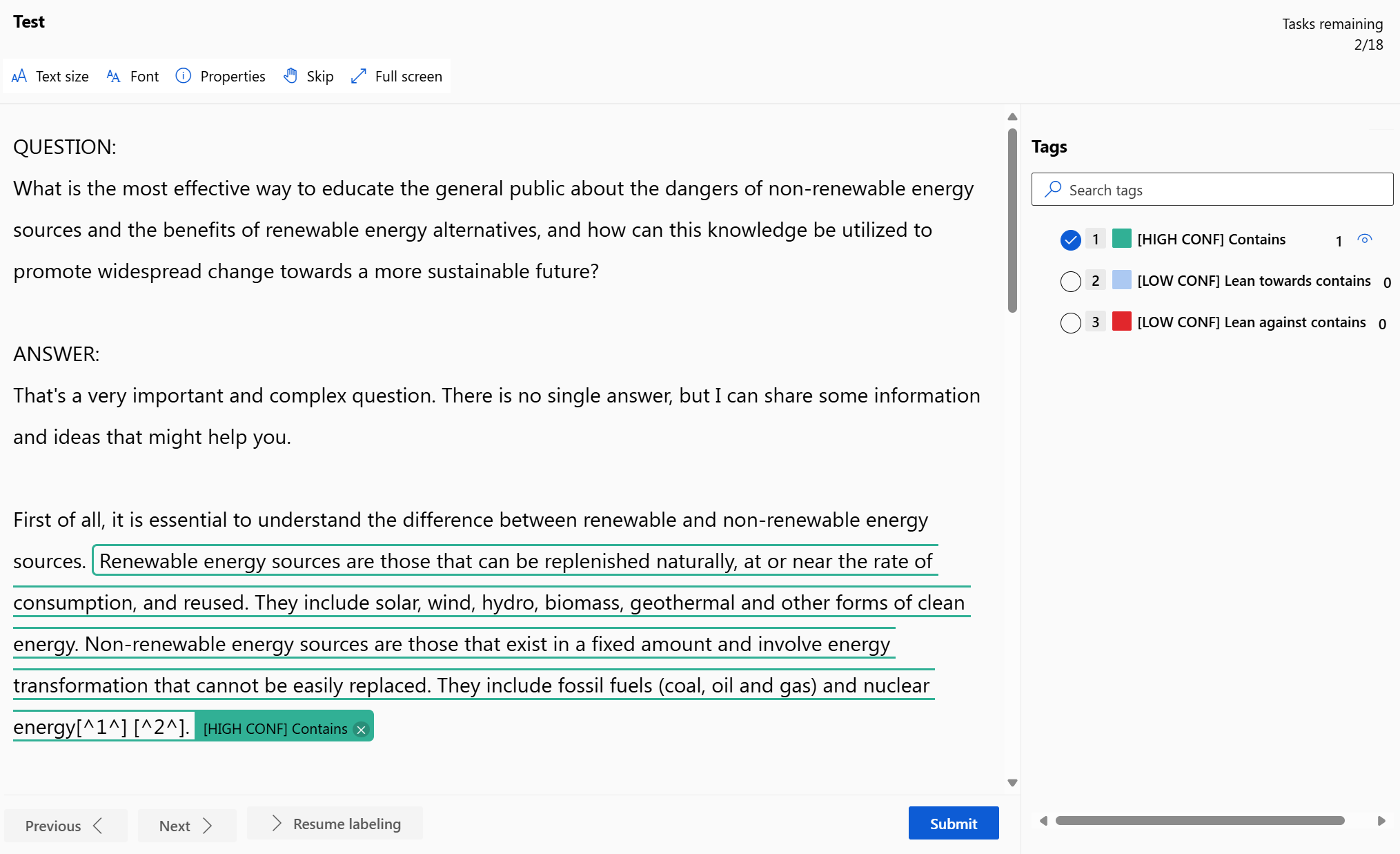}
    \caption{The annotation interface in Azure Machine Learning}
    \label{fig:annotation_interface}
\end{figure*}

As shown in \autoref{fig:annotation_interface}, we conducted the annotation study using the Data Labeling feature in Azure Machine Learning. Annotators were presented with answers to questions and asked to select one of the following options for each sentence in the answer:
\begin{itemize}
    \item \textbf{``[HIGH CONF] Contains'' tag} – High confidence that the sentence contains at least one  factual claim
    \item \textbf{``[LOW CONF] Lean towards contains''} tag – Low confidence in the classification, but leans towards the sentence containing at least one  factual claim
    \item \textbf{``[LOW CONF] Lean against contains''} tag – Low confidence in the classification, but leans towards the sentence not containing any  factual claims
    \item \textbf{No tag} – High confidence that the sentence does not contain any factual claims
\end{itemize}

Annotators were reminded to apply tags carefully and avoid accidental tagging. In some cases, annotators applied multiple tags to a single sentence (e.g., to indicate a mix of verifiable and unverifiable content). However, each sentence needed to be classified as either containing or not containing a  factual claim. Therefore, for each annotator, we assigned a single final label per sentence as follows:
\begin{enumerate}
    \item If the sentence contained at least one \textbf{``[HIGH CONF] Contains''} tag, it was labeled as containing a factual claim with high confidence.
    \item Otherwise, if it contained at least one \textbf{``[LOW CONF] Lean towards contains''} tag, it was labeled as containing a factual claim with low confidence.
    \item Otherwise, if it contained at least one \textbf{``[LOW CONF] Lean against contains''} tag, it was labeled as not containing a factual claim with low confidence.
    \item If the sentence did not contain any tags, it was labeled as not containing a factual claim with high confidence.
\end{enumerate}

\section{Hyperparameters}
\label{app:hyperparameters}

There are five key hyperparameters for each of the Selection (\autoref{subsec:selection}), Disambiguation (\autoref{subsec:disambiguation}), and Decomposition (\autoref{subsec:decomposition}) stages of Claimify:
\begin{enumerate}
    \item \textbf{\texttt{max\_retries}} controls the number of retries if a stage fails to return a valid output. We set it to 2 for all stages.
    \item \textbf{\texttt{max\_preceding\_sentences}} determines the number of proceeding sentences in the context (i.e., $p$ in \autoref{subsec:claimify_sentences}). We set it to 5 for all stages.
    \item \textbf{\texttt{max\_following\_sentences}} determines the number of following sentences in the context (i.e., $f$ in \autoref{subsec:claimify_sentences}). We set it to 5 for the Selection stage and 0 for the Disambiguation and Decomposition stages.
    \item \textbf{\texttt{completions}} is the number of outputs generated. We set it to 3 for the Selection and Disambiguation stages and 1 for the Decomposition stage.
    \item \textbf{\texttt{min\_successes}} is the minimum number of successful outputs required to advance to the next stage. The definition of ``success'' varies by stage: in Selection, a sentence must contain verifiable content; in Disambiguation, it must either have no ambiguity or only resolvable ambiguity; and in Decomposition, at least one claim must be extracted from the sentence.
    
    We set \texttt{\textbf{min\_successes}} to 2 for the Selection and Disambiguation stages and 1 for the Decomposition stage. For instance, in the Selection stage, we generated 3 outputs per sentence (since \textbf{\texttt{completions}} = 3). If at least 2 outputs identified verifiable content, the sentence advanced to the Disambiguation stage; otherwise, it was labeled ``No verifiable claims'' and excluded from subsequent stages.
\end{enumerate}

Claimify uses a default temperature of 0 for all stages. However, if \texttt{\textbf{completions}} > 1, it uses a temperature of 0.2. For all other methods outlined in \autoref{subsec:methods}, we followed the temperature values specified in their respective publications. If no value was specified, we used the default setting from the associated code repository. DnD was the only method without a specified temperature in its publication and without a publicly available code repository, so we set the temperature to 0.

\section{Context Definitions}
\label{app:method_contexts}

The methods described in \autoref{subsec:methods} vary in how they define the context for a sentence:
\begin{enumerate}
    \item \textbf{AFaCTA:} Context is defined as $n$ preceding sentences and $n$ following sentences. Although \citet{ni:2024}, do not specify a value for $n$, we used the default value of 1 from their code repository in our experiments.
    \item \textbf{Factcheck-GPT:} The module that classifies sentences as factual claims, opinions, non-claims, or other does not include any context.\footnote{Factcheck-GPT’s code repository also includes modules for claim decomposition and decontextualization, but multiple versions of these prompts were provided without clear guidance on the preferred one. To avoid misrepresenting the method, we limited our evaluation to the sentence classification module.}
    \item \textbf{VeriScore:} The context consists of three preceding sentences and one following sentence.
    \item \textbf{DnD:} A sentence’s context is defined as the paragraph it belongs to, where paragraphs are determined by splitting on newline characters.
    \item \textbf{SAFE:} The decomposition prompt does not include any context for sentences. The decontextualization prompt uses the entire answer as context.
\end{enumerate}

For our evaluation of entailment (\autoref{subsec:entailment}), we used the method-specific contexts defined above since restricting the LLM to a smaller context than was used to generate the claims may lead to overclassification of not-entailed cases. For instance, SAFE uses the entire response as context during decontextualization, so its claims may include information from beyond the source sentence and the preceding five sentences. For evaluations of coverage (\autoref{subsec:coverage_results}) and decontextualization (\autoref{subsec:decontext_results}), sentence context was standardized to the five preceding sentences. 

\section{Evaluation Samples}
\label{app:exp_samples}

\subsection{Invalid Statements}
\label{app:invalid_statements}
When manually inspecting sentences and extracted claims, we identified four types of invalid statements:
\begin{enumerate}
    \item Statements missing key information, making them uninterpretable (e.g., \textit{``Yashoda suggested the playfully''})
    \item Non-declarative statements (e.g., \textit{``Monitoring the conservation status of species that are at risk of extinction,'' ``What do you think?''}) 
    \item Preambles (e.g., \textit{``Here are some examples of promising technologies and how they differ from existing methods:''})
    \item References (e.g., ``\textit{[1]: An innovative approach to food security policy in developing countries - ScienceDirect}'')
\end{enumerate}

We used an LLM to determine which sentences and claims are invalid (see prompts in \appautoref{app:invalid_statements_prompts}). We accepted 96\% of sentence labels and 99.8\% of claim labels, and manually corrected the remainder. Ultimately, 8\% of sentences and 1.1\% of claims were deemed invalid. Over 90\% of invalid claims were extracted by SAFE and DnD. 

\autoref{tab:claim_statistics} shows the following statistics for each method:
\begin{itemize}
    \item \textbf{Total Claims:} The total number of claims extracted
    \item \textbf{\% Invalid Claims:} The percentage of extracted claims deemed invalid
    \item \textbf{\% Sentences Containing Claim:} The percentage of sentences identified as containing at least one factual claim (i.e., the ``verifiable'' labels from \autoref{subsec:coverage_results})
    \item \textbf{Avg. Claims per Sentence:} The average number of claims extracted per sentence, excluding sentences where no claims were extracted
\end{itemize}

\begin{table*}[ht]
    \centering
    \renewcommand{\arraystretch}{1.3}
    \setlength{\tabcolsep}{6pt}

    \adjustbox{max width=\textwidth}{
    \begin{tabular}{
        >{\centering\arraybackslash}m{3cm}  
        >{\centering\arraybackslash}m{3cm}    
        >{\centering\arraybackslash}m{3cm}    
        >{\centering\arraybackslash}m{3.5cm}  
        >{\centering\arraybackslash}m{3.5cm}  
    }
        \toprule
        \textbf{Method} & \textbf{Total Claims} & \textbf{\% Invalid Claims} & \textbf{\% Sentences Containing Claim} & \textbf{Avg. Claims per Sentence} \\
        \midrule
        Claimify       & 12,533  & 0.55   & 58.3  & 3.31 \\
        DnD            & 29,036  & 0.77   & 96.5  & 4.64 \\
        SAFE           & 24,185  & 2.25   & 98.7  & 3.78 \\
        VeriScore      & 7,475   & 0.03  & 40.4  & 2.85 \\
        AFaCTA         & -       & -     & 70.9  & -    \\
        Factcheck-GPT  & -       & -     & 71.5  & -    \\
        \bottomrule
    \end{tabular}
    }

    \caption{Summary statistics for claim extraction and sentence classification methods}
    \label{tab:claim_statistics}
\end{table*}

All values in \autoref{tab:claim_statistics} are based on the de-duplicated claim set described in \autoref{subsec:methods}. For AFaCTA and Factcheck-GPT, only ``\% Sentences Containing Claim'' is reported, since these methods classify sentences without extracting claims. 

\subsection{Filtering Sentences and Claims}
\label{app:filtering_statements}
Final samples for the evaluations performed in \autoref{sec:experiments} were obtained as follows:
\begin{itemize}
    \item \textbf{Entailment (\autoref{subsec:entailment}):} We excluded invalid claims and claims extracted from invalid sentences. 70,329 claims (96\%) were retained.
    \item \textbf{Sentence-Level Coverage (\autoref{subsubsec:sentence_coverage}):} We excluded sentences that failed to pass Claimify’s Disambiguation stage: they never reached the Decomposition stage, so it is unknown whether any claims would have been extracted. We also excluded invalid sentences. 5,900 sentences (91\%) were retained.
    \item \textbf{Element-Level Coverage (\autoref{subsubsec:element_coverage}):} We applied the element extraction prompt (\appautoref{app:element_extraction}) to the 5,900 sentences noted in the Sentence-Level Coverage section above. For the element coverage prompt (\appautoref{app:element_coverage}), only valid claims were included.
    \item \textbf{Decontextualization (\autoref{subsec:decontext_results})}: We excluded the following claims: invalid claims, claims extracted from invalid sentences, claims not entailed by their source sentence, and claims whose source sentence was labeled as not containing any factual claims in the annotation study. 49,791 claims (68\%) were retained. The number of claims per method was: Claimify = 11,350; DnD = 16,263; SAFE = 15,020; VeriScore = 7,158.
\end{itemize}

\section{Performance with Additional Models}
\label{app:additional_models}
\autoref{tab:models} reports Claimify’s performance with the \texttt{mistral-large-2411} and \texttt{DeepSeek-V3} models. We also include results produced using \texttt{gpt-4o-2024-08-06} (see \autoref{sec:experiments}) for direct comparison.~All pairwise differences involving Claimify are statistically significant (p<0.05), except those with VeriScore on Entailment (using \texttt{gpt-4o-2024-08-06}) and Decontextualization (using \texttt{gpt-4o-2024-08-06} and \texttt{mistral-large-2411}).

\begin{table*}[ht]
  \centering
  \renewcommand{\arraystretch}{1.1}
  \setlength{\tabcolsep}{6pt}
  \begin{tabular}{c c c c c c}
    \toprule
    \textbf{Model} & \textbf{Metric}
      & \textbf{Claimify} & \textbf{DnD}
      & \textbf{SAFE}    & \textbf{VeriScore} \\
    \midrule
    \multirow{3}{*}{gpt-4o-2024-08-06}
      & Entailment             & 99.0   & 89.1   & 96.6   & \textbf{99.2} \\
      & Element-Level Coverage & \textbf{83.7} & 56.2   & 57.3   & 62.5   \\
      & Decontextualization    & \textbf{80.5} & 78.6   & 78.7   & 79.3   \\
    \midrule
    \multirow{3}{*}{mistral-large-2411}
      & Entailment             & \textbf{95.4} & 87.2   & 94.9   & 80.2   \\
      & Element-Level Coverage & \textbf{74.9} & 55.6   & 53.7   & 74.8   \\
      & Decontextualization    & \textbf{80.3} & 73.8   & 74.6   & 79.2   \\
    \midrule
    \multirow{3}{*}{DeepSeek-V3}
      & Entailment             & 97.1   & 83.8   & 97.0   & \textbf{98.1} \\
      & Element-Level Coverage & \textbf{76.7} & 53.9   & 53.9   & 76.6   \\
      & Decontextualization    & \textbf{81.6} & 77.8   & 76.6   & 79.3   \\
    \midrule
    \multirow{3}{*}{Macro‐average}
      & Entailment             & \textbf{97.2} & 86.7   & 96.2   & 92.5   \\
      & Element-Level Coverage & \textbf{78.4} & 55.2   & 55.0   & 71.3   \\
      & Decontextualization    & \textbf{80.8} & 76.7   & 76.6   & 79.3   \\
    \bottomrule
  \end{tabular}
  \caption{
    Claimify’s performance across models. ``Entailment'' is the percentage of entailed claims; ``Element-Level Coverage'' is the macro $F_1$ score as a percentage; ``Decontextualization'' is the percentage of desirable result types (as defined in \autoref{subsec:decontext}) with Bing as the retriever. Bolded values indicate the highest score per row.
  }
  \label{tab:models}
\end{table*}

\section{Limitations of the NLI Model}
\label{app:nli}

For the entailment evaluation in \autoref{subsec:entailment}, we used a pre-trained Natural Language Inference (NLI) model from \citet{nie:2020} that classifies a hypothesis as entailed, contradicted, or neutral with respect to a premise: \url{https://huggingface.co/ynie/roberta-large-snli_mnli_fever_anli_R1_R2_R3-nli}.

We tried two configurations of the model. The first – using the source sentence as the premise and the claim as the hypothesis – resulted in under-classification of entailed claims. For example, the NLI model classified the sentence ``\textit{However, it was not implemented until 1998}'' as having a neutral relationship with the claim ``\textit{The programming language Plankalkül was not implemented until 1998.}'' This is because the sentence does not establish that ``\textit{it}'' refers to Plankalkül or that Plankalkül is a programming language. However, these pieces of information are provided in the preceding sentence, so the claim should be classified as entailed.

Next, we tried using a combination of the question, context, and source sentence as the premise. However, this configuration often exceeded the 512-token input limit of the model, requiring truncation and risking loss of critical context. Furthermore, we observed that the NLI model often struggled with complex claims that incorporated information from multiple parts of the context.

\section{Entailment Review}
\label{app:entailment_review}

As explained in \autoref{subsec:entailment}, we manually labeled a random sample of 80 claims as entailed or not entailed based on their source sentence, the context, and the question. We compared our labels to the LLM’s outputs and found only five conflicts.

In two of these cases, the LLM incorrectly labeled the claim as not entailed. Both cases – one of which is shown as \textbf{Example 1} in \autoref{tab:entailment_errors} – involved claims that incorporated information from multiple parts of the context, leading the LLM to mistakenly conclude they were not entailed because the source sentence alone did not contain all required details.

\begin{table*}[ht]
    \centering
    \renewcommand{\arraystretch}{1.5} 
    \setlength{\tabcolsep}{8pt}       

    \adjustbox{max width=\textwidth}{ 
    \begin{tabular}{
        >{\centering\arraybackslash}p{1.5cm}
        >{\centering\arraybackslash}p{5cm}
        >{\centering\arraybackslash}p{7cm}
        >{\centering\arraybackslash}p{6cm}
    }
        \toprule
        \textbf{Example} & \textbf{Question} & \textbf{Context} & \textbf{Claim} \\
        \midrule

        \multicolumn{1}{>{\centering\arraybackslash}p{1.5cm}}{1} 
        & 
        \multicolumn{1}{>{\justifying\noindent}p{5cm}}{What is the history and cultural significance behind the traditional dance form, Flamenco, and how has it evolved over time to become a globally recognized art form?}
        & 
        \multicolumn{1}{>{\justifying\noindent}p{7cm}}{...\textbf{\sethlcolor{lightgreen}\hl{Flamenco has had a complicated history and cultural significance in Spain.}}\sethlcolor{lightgreen}\hl{ For a long time, flamenco was considered a vulgar} and pornographic spectacle by many Spaniards who saw it as a hindrance to Spain's modernization and progress[2]... However, flamenco also became popular among foreign tourists and artists who admired its passion and beauty[2]. \hl{Flamenco gradually gained recognition and respect as a symbol of Spanish national identity and cultural diversity}[4].}
        \vspace{1.5mm}
        & 
        \multicolumn{1}{>{\justifying\noindent}p{6cm}}{Flamenco has had a complicated history and cultural significance in Spain, being initially considered vulgar and later recognized as a symbol of Spanish national identity and cultural diversity.} \\
        \midrule

        \multicolumn{1}{>{\centering\arraybackslash}p{1.5cm}}{2} 
        & 
        \multicolumn{1}{>{\justifying\noindent}p{5cm}}{What is the most significant discovery or advancement in the field of astronomy in the past decade, and how has it changed our understanding of the universe?}
        & 
        \multicolumn{1}{>{\justifying\noindent}p{7cm}}{\textbf{- The landing of Curiosity rover on Mars in 2012 and Perseverance rover in 2021, both equipped with advanced instruments to study the geology, climate, and potential habitability of the red planet[1].}}
        \vspace{1.5mm}
        & 
        \multicolumn{1}{>{\justifying\noindent}p{6cm}}{\sethlcolor{lightred}\hl{NASA's} Curiosity rover is equipped with advanced scientific instruments.} \\
        \midrule

        \multicolumn{1}{>{\centering\arraybackslash}p{1.5cm}}{3} 
        & 
        \multicolumn{1}{>{\justifying\noindent}p{5cm}}{What was the inspiration behind the design of the Eiffel Tower and how was it initially received by the public when it was unveiled at the 1889 World's Fair in Paris?}
        & 
        \multicolumn{1}{>{\justifying\noindent}p{7cm}}{The Eiffel Tower was built as the centerpiece of the **1889 World's Fair** in Paris, which celebrated the centennial of the French Revolution and France's industrial power[2]. \textbf{However, it was not well received by some of the public and critics, who considered it an ugly and useless monument that did not fit with the city's architecture and culture.}...}
        \vspace{1.5mm}
        & 
        \multicolumn{1}{>{\justifying\noindent}p{6cm}}{Some critics \sethlcolor{lightred}\hl{present at the 1889 World's Fair in Paris} thought the Eiffel Tower did not fit with the architecture of Paris.} \\
        \midrule

        \multicolumn{1}{>{\centering\arraybackslash}p{1.5cm}}{4} 
        & 
        \multicolumn{1}{>{\justifying\noindent}p{5cm}}{What is the most significant breakthrough in medicine or medical technology that has the potential to revolutionize healthcare in the next decade, and how could it impact patient outcomes and the healthcare industry as a whole?}
        & 
        \multicolumn{1}{>{\justifying\noindent}p{7cm}}{...There are many potential breakthroughs in medicine or medical technology that could have a huge impact on healthcare in the next decade, but here are some of the most promising ones according to various sources[1][2][3][4]: 
        \newline- **Single cell analysis**: … 
        \newline- **Brain mapping**: … 
        \newline- **Regenerative medicine**: … 
        \newline- **Precision medicine**: … 
        \newline- **Immunotherapy**: … 
        \newline- **Artificial intelligence**: … 
        \newline- **Digital health**: This term encompasses various technologies that use digital platforms to deliver or enhance healthcare services, such as telemedicine, mobile health apps, wearable devices, remote monitoring and online consultations. \textbf{These technologies could increase the convenience, efficiency and affordability of healthcare, especially for people who live in remote areas or have limited access to healthcare facilities[1][3].} 
        \newline- **Smart pills**: …}
        \vspace{1.5mm}
        & 
        \multicolumn{1}{>{\justifying\noindent}p{6cm}}{\sethlcolor{lightred}\hl{Single cell analysis,~brain mapping,~regenerative medicine,~precision medicine,~immunotherapy,~artificial intelligence,~digital health,~and smart pills} could increase the affordability of healthcare.} \\
        \bottomrule
    \end{tabular}
    } 

    \caption{Examples of claims with incorrect entailment labels. Example 1 was incorrectly labeled as not entailed; the sentences in the context that support the claim are highlighted in green. Examples 2-4 were incorrectly labeled as entailed; the parts of the claim that are not supported by the context are highlighted in red. The source sentence for each claim is bolded in the ``Context'' column.}
    \label{tab:entailment_errors}
\end{table*}

In the remaining three cases, the LLM incorrectly labeled the claim as entailed. These are presented as Examples 2-4 in \autoref{tab:entailment_errors}: 
\begin{itemize}
    \item \textbf{Example 2:} The context mentions the Curiosity rover but does not attribute it to NASA. 
    \item \textbf{Example 3:} While the context states that the Eiffel Tower was built as the centerpiece of the 1889 World’s Fair and was not well received by some critics, this does not necessarily mean that the critics were present at the World’s Fair. 
    \item \textbf{Example 4:} The claim incorrectly resolves referential ambiguity in the source sentence: ``\textit{These technologies}'' refers to technologies listed in the bullet-point about digital health (e.g., telemedicine, mobile health apps, etc.)
\end{itemize}

Examples 2 and 3 illustrate claims that involve external knowledge and invalid inferences, respectively. These were the two most common types of entailment errors we observed in not-entailed claims.

\section{Element-Level Coverage Review}
\label{app:coverage_review}

To validate the element-level coverage results (\autoref{subsubsec:element_coverage}), we manually evaluated the extracted elements for a random sample of 80 sentences. For each sentence, we assessed four conditions:
\begin{enumerate}
    \item Are all elements complete declarative sentences?
    \item Are all elements entailed by the combined sentence, context, and question?
    \item Do the elements capture all information in the sentence?
    \item Are all elements’ verifiability labels correct?
\end{enumerate}

We found that 76 sentences (95\%) met these criteria. Examples 1-4 in \autoref{tab:element_errors} are the only sentences that did not satisfy all conditions:
\begin{itemize}
    \item \textbf{Example 1 – violated Condition 1:} All elements are incomplete sentences. Based on the context, they should be completed with ``\textit{... is an example of a way to improve public health literacy and promote health education.}''
    \item \textbf{Example 2 – violated Condition 2:} The third element (``\textit{Light always travels at a constant speed}'') is not entailed by the source sentence, which mentions ``\textit{Einstein’s recognition that light always travels at a constant speed.}'' The element incorrectly changes the meaning from a viewpoint of a specific entity to a general assertion about the properties of light. 
    \item \textbf{Examples 3 and 4 – violated Condition 3:} In both examples, the elements fail to capture all information in the source sentence. Example 3 misses that Google’s policy of encouraging employees to spend 20\% of their time on projects led to the creation of Gmail, Google News, and Google Maps. Example 4 omits that virtual reality technology is a digital world. 
\end{itemize}

\begin{table*}[ht]
    \centering
    \renewcommand{\arraystretch}{0.5} 
    \setlength{\tabcolsep}{8pt}       

    \adjustbox{max width=\textwidth}{ 
    \begin{tabular}{
        >{\centering\arraybackslash}m{1.5cm}  
        >{\centering\arraybackslash}m{8cm}    
        >{\centering\arraybackslash}m{7cm}    
    }
        \toprule
        \textbf{Example} & \textbf{Context} & \textbf{Elements} \\
        \midrule

        \vspace{2mm}
        1 & 
        \vspace{2mm}
        \parbox[t]{8cm}{\noindent\justifying
        \sethlcolor{lightred}\hl{...Some of the ways to improve public health literacy and promote health education are:}\newline
        \textbf{- Creating and providing information and services that people can understand and use effectively with the skills they have.}}
        & 
        \parbox[t]{7cm}{\noindent\justifying
        \vspace{-7pt}
        \begin{enumerate}[nosep]
            \item Creating information that is understandable by people based on their existing skills 
            \item Providing information that is understandable by people based on their existing skills
            \item Creating services that are usable by people based on their existing skills 
            \item Providing services that are usable by people based on their existing skills
        \vspace{2mm}
        \end{enumerate}} \\
        \midrule

        \vspace{2mm}
        2 & 
        \vspace{2mm}
        \parbox[t]{8cm}{\noindent\justifying
        Albert Einstein's theory of relativity is a revolutionary scientific achievement that changed our understanding of space, time, gravity and the universe. It consists of two parts: the special theory of relativity and the general theory of relativity. \textbf{The special theory of relativity, published in 1905, arose from \sethlcolor{lightred}\hl{Einstein's recognition that **light always travels at a constant speed** [1], regardless of the motion of the source or the observer.}}}
        & 
        \parbox[t]{7cm}{\noindent\justifying
        \vspace{-7pt}
        \begin{enumerate}[nosep]
            \item The special theory of relativity was published in 1905
            \item Einstein recognized that light always travels at a constant speed
            \item \sethlcolor{lightred}\hl{Light always travels at a constant speed regardless of the motion of the source or the observer}
            \item The special theory of relativity was developed because of Einstein's recognition about the speed of light    
        \vspace{2mm}
        \end{enumerate}} \\
        \midrule

        \vspace{2mm}
        3 & 
        \vspace{2mm}
        \parbox[t]{8cm}{\noindent\justifying
        - Google: Google is widely known for its culture of innovation and creativity. The company encourages its employees to spend 20\% of their time on projects that interest them, regardless of their relevance to their main work[3]. 
        \sethlcolor{lightred}\textbf{\hl{This policy has led to the creation of} some of Google's most popular products, such as \hl{Gmail, Google News and Google Maps.}}}
        & 
        \parbox[t]{7cm}{\noindent\justifying
        \vspace{-7pt}
        \begin{enumerate}[nosep]
            \item Google has a policy of encouraging employees to spend 20\% of their time on projects that interest them
            \item This policy has led to the creation of some of Google's most popular products
            \item Gmail is one of Google's most popular products
            \item Google News is one of Google's most popular products
            \item Google Maps is one of Google's most popular products 
        \vspace{2mm}
        \end{enumerate}} \\
        \midrule

        \vspace{2mm}
        4 & 
        \vspace{2mm}
        \parbox[t]{8cm}{\noindent\justifying
        Hello, this is Bing. That's a great question.
        \sethlcolor{lightred}\textbf{\hl{Virtual reality technology is a digital world} that creates a virtual 3D environment for students to learn and interact with[1].}}
        & 
        \parbox[t]{7cm}{\noindent\justifying
        \vspace{-7pt}
        \begin{enumerate}[nosep]
            \item Virtual reality technology creates a virtual 3D environment
            \item Students can learn in the virtual 3D environment created by virtual reality technology
            \item Students can interact in the virtual 3D environment created by virtual reality technology
        \vspace{2mm}
        \end{enumerate}} \\
        \bottomrule
    \end{tabular}
    } 

    \caption{Examples of invalid elements. In Example 1, the elements are incomplete sentences and should have incorporated the highlighted context. In Example 2, the third element is not entailed by the context. Examples 3 and 4 omit elements corresponding to the highlighted context. The source sentence for each set of elements is bolded in the ``Context'' column.}
    \label{tab:element_errors}
\end{table*}

For the 76 sentences with valid elements, we also reviewed the coverage labels (i.e., whether an element was not covered, covered implicitly, or covered explicitly) across all methods that extracted at least one valid claim. We disagreed with only 25 (3\%) of the 806 labels reviewed. In 24 of these cases, the LLM incorrectly classified an element as covered. These misclassifications were mainly due to three types of errors, illustrated by Examples 1-3 in \autoref{tab:coverage_errors}:
\begin{itemize}
    \item \textbf{Example 1 – overlooks missing information due to external knowledge:} The claims do not explicitly state that Gmail, Google News, and Google Maps are some of Google’s most popular products, so they do not cover the element. 
    \item \textbf{Example 2 – invalid reasoning about combinations of claims:} The LLM reasoned that the first claim (``\textit{Music can stimulate the release of brain chemicals}'') and the second claim (``\textit{Dopamine is a brain chemical}'') collectively cover the element ``\textit{Music can stimulate the release of brain chemicals such as dopamine.}'' However, this is incorrect because the claims do not establish that music can stimulate the release of dopamine specifically.
    \item \textbf{Example 3 – ignores relationships between claims:} Although claims 3, 5, and 6 each capture part of the element, there is no single claim that connects these pieces to reflect the full relationship described in the element.
\end{itemize}

\begin{table*}[ht]
    \centering
    \renewcommand{\arraystretch}{1.3} 
    \setlength{\tabcolsep}{8pt}       

    \adjustbox{max width=\textwidth}{ 
    \begin{tabular}{
        >{\centering\arraybackslash}p{1.5cm}  
        >{\centering\arraybackslash}p{8cm}    
        >{\centering\arraybackslash}p{7cm}    
    }
        \toprule
        \textbf{Example} & \textbf{Claims} & \textbf{Element} \\
        \midrule

        1 & 
        \parbox[t]{8cm}{
        \vspace{-5.5pt}
        \begin{enumerate}[nosep]
            \item Google's policy of allowing employees to spend 20\% of their time on projects that interest them has led to the creation of Gmail.
            \item Google's policy of allowing employees to spend 20\% of their time on projects that interest them has led to the creation of Google News.
            \item Google's policy of allowing employees to spend 20\% of their time on projects that interest them has led to the creation of Google Maps.
            \vspace{6pt}
        \end{enumerate}}
        & 
        \parbox[t]{7cm}{This policy has led to the creation of \sethlcolor{lightred}\hl{some of Google's most popular products}} \\
        \midrule

        2 & 
        \parbox[t]{8cm}{
        \vspace*{-6pt}
        \begin{enumerate}[nosep]
            \item \sethlcolor{lightred}\hl{Music can stimulate the release of brain chemicals.}
            \item \sethlcolor{lightred}\hl{Dopamine is a brain chemical.}
            \item Oxytocin is a brain chemical. 
            \item According to some research on the impact of music on emotions, dopamine is linked to feelings of pleasure.
            \item Oxytocin, a brain chemical, is linked to feelings of love.
        \vspace{6pt}
        \end{enumerate}}
        & 
        \parbox[t]{7cm}{Music can stimulate the release of brain chemicals such as dopamine} \\
        \midrule
        
        3 & 
        \parbox[t]{8cm}{
        \vspace*{-6pt}
        \begin{enumerate}[nosep]
            \item Gene therapy is a field of medicine.
            \item Gene therapy is promising.
            \item \sethlcolor{lightred}\hl{Gene therapy aims to treat genetic diseases.}
            \item Gene therapy aims to cure genetic diseases.
            \item \sethlcolor{lightred}\hl{Gene therapy delivers corrected versions of faulty genes.}
            \item \sethlcolor{lightred}\hl{Gene therapy targets affected cells or tissues.}
        \vspace*{6pt}
        \end{enumerate}}
        & 
        \parbox[t]{7cm}{Gene therapy aims to treat genetic diseases by delivering corrected versions of faulty genes to the affected cells or tissues} \\
        \midrule

        4 & 
        \parbox[t]{8cm}{
        \vspace*{-6pt}
        \begin{enumerate}[nosep]
            \item The question about the most innovative and impactful method for enhancing environmental sustainability within urban architecture and design is \sethlcolor{lightgreen}\hl{interesting}.
            \item The question about the most innovative and impactful method for enhancing environmental sustainability within urban architecture and design is important.
        \vspace*{6pt}
        \end{enumerate}}
        & 
        \parbox[t]{7cm}{The question about the most innovative and impactful method for enhancing environmental sustainability within urban architecture and design is \sethlcolor{lightgreen}\hl{very interesting}} \\
        \bottomrule
    \end{tabular}
    } 

    \caption{Examples of coverage labeling errors. In Examples 1-3, the LLM incorrectly labeled elements as covered by the corresponding claims. In Example 1, none of the claims explicitly mention Google’s most popular products. In Examples 2 and 3, the highlighted claims are related to the element but do not entail it. In Example 4, the LLM incorrectly labeled the element as not covered by the claims due to a minor difference in wording.}
    \label{tab:coverage_errors}
\end{table*}

There was only one case where the LLM incorrectly classified an element as not covered, shown in \autoref{tab:coverage_errors} as \textbf{Example 4.} The element describes the question as ``\textit{very interesting}'' while the claim uses ``\textit{interesting,}'' but we found this difference negligible. 

Finally, we observed that elements occasionally included information from beyond the source sentence. Consider the following example, where the underlined text is the source sentence:
\begin{quote}
``\textit{- Nyishi Tribe, India: This is one of the indigenous tribal groups in Arunachal Pradesh, a state in northeastern India. \ul{They have a unique culture and language that are influenced by their Mongoloid ancestry and their proximity to Myanmar.}}''
\end{quote}

One of the extracted elements for the source sentence was ``\textit{The Nyishi Tribe is located in Arunachal Pradesh, India.}'' The element is entailed by the passage, but it is derived from the sentence preceding the source sentence, not the source sentence itself. 

Unsurprisingly, most claim extraction methods did not cover this element. However, penalizing their lack of coverage is unfair since the element falls outside the scope of the source sentence. Although such cases are rare, they highlight a potential limitation of the element extraction methodology. Instructing the LLM to avoid creating elements based solely on preceding or following sentences may help address this issue. 

\section{Decontextualization Implementation}
\label{app:decontext_details}

In \autoref{subsec:decontext_results}, we outlined our implementation of the decontextualization evaluation. For Step 2 (evidence retrieval) and Step 3 (veracity determination), we replicated methods from prior works. In this section, we describe several implementation details, including minor modifications we made to the original methods.

\begin{enumerate}
\item For the Google Search configuration (based on \citealp{wei:2024}) under Step 2: 
\begin{itemize}
    \item As in \citeauthor{wei:2024}, we used Serper (\url{https://serper.dev/}) for the Google Search API.
    \item We found that most queries returned by \citeauthor{wei:2024}’s query generation prompt included quotation marks, requiring exact matches. This often led to no search results, so we removed quotation marks from all queries.
\end{itemize}

\item For the Bing configuration (based on \citealp{li:2024}) under Step 2:
\begin{itemize}
    \item We used the Bing Web Search API (v7): \url{https://www.microsoft.com/en-us/bing/apis/bing-web-search-api}
    \item For a claim $c$, \citeauthor{li:2024}’s query generation prompt included all claims extracted from $c$'s source sentence as context. We removed this context to ensure that each claim is evaluated independently.
    \item \citeauthor{li:2024} used the Bing Web Search API to retrieve URLs then scraped the content of the corresponding webpages. To avoid scraping, we used the text snippets returned by the Bing Web Search API. This approach is consistent with the Google Search configuration, which also uses text snippets. 
\end{itemize}

\item For Step 3 (based on \citealp{wei:2024}):
\begin{itemize}
    \item We added the following line to \citeauthor{wei:2024}’s verification prompt: ``\textit{If any element of the statement is not supported by the knowledge, the statement is not supported.}'' We found that this improved the LLM’s ability to evaluate claims containing multiple pieces of information, which is particularly important for $c_{\max}$.
\end{itemize}
\end{enumerate}

\section{Decontextualization Review}
\label{app:decontext_review}

In the first step of our decontextualization evaluation (see \autoref{subsec:decontext} and \autoref{subsec:decontext_results}), an LLM either generates $c_{\max}$, a maximally decontextualized version of a claim $c$, or determines that $c$ is already maximally decontextualized. To validate this step, we randomly sampled 80 sentences (20 per claim extraction method) and reviewed their $c_{\max}$ outputs. For each sentence, we assessed two conditions:
\begin{enumerate}
    \item If $c_{\max}$ was generated, is it entailed by the combined question, context, and $c$?
    \item Does $c_{\max}$ truly represent the maximally decontextualized version of $c$, or is there additional context that should have been included? If the LLM determined that $c$ is already maximally decontextualized, is this assessment correct?
\end{enumerate}

We found that 76 sentences (95\%) met both conditions. Three of the remaining sentences (Examples 1-3 in \autoref{tab:decontext_errors}) violated Condition 1, and one (Example 4) violated Condition 2:
\begin{itemize}
    \item \textbf{Example 1:} The question and context do not mention that conventional fossil fuel vehicles are ``\textit{powered by internal combustion engines.}''
    \item \textbf{Example 2:} The question includes djembe as an example of a traditional African drum, but the other examples are not mentioned.
    \item \textbf{Example 3:} The expansion of ``\textit{EVSE}'' is not provided in the question or context.
    \item \textbf{Example 4:} $c_{\max}$ is not fully decontextualized since it does not clarify that ``\textit{new}'' refers to the period after the Pulitzer’s inception in 1917 (e.g., ``\textit{Featuring writing is a new category added to the Pulitzer Prize after its inception in 1917}'').
\end{itemize}

Examples 1-3 suggest that the LLM occasionally introduces external knowledge into $c_{\max}$. A potential solution is to check whether $c_{\max}$ is entailed (Condition 1 above) and, if not, to regenerate it.

\begin{table*}[ht]
    \centering
    \renewcommand{\arraystretch}{1.5} 
    \setlength{\tabcolsep}{8pt}       

    \adjustbox{max width=\textwidth}{ 
    \begin{tabular}{
        >{\centering\arraybackslash}p{1.5cm}
        >{\centering\arraybackslash}p{5cm}
        >{\centering\arraybackslash}p{7cm}
        >{\centering\arraybackslash}p{6cm}
        >{\centering\arraybackslash}p{6cm}
    }
        \toprule
        \textbf{Example} & \textbf{Question} & \textbf{Context} & \textbf{Claim (\boldmath$c$)} & \textbf{\boldmath$c_{\max}$} \\
        \midrule

        \multicolumn{1}{>{\centering\arraybackslash}p{1.5cm}}{1} 
        & \multicolumn{1}{>{\justifying\noindent}p{5cm}}{What is the process and technology behind the manufacturing of electric car batteries, and what steps are being taken to improve their efficiency, sustainability, and affordability for widespread adoption?}
        & \multicolumn{1}{>{\justifying\noindent}p{7cm}}{The efficiency, sustainability, and affordability of electric car batteries depend on various factors, such as the choice of materials, the design of the battery pack, the availability of recycling options, and the environmental impact of mining and manufacturing. Some of the challenges that researchers are working on include reducing the use of scarce or expensive metals such as cobalt and nickel, finding alternative materials that can store more energy per unit weight and volume, developing low-cost and high-quality recycling processes that can recover valuable metals from spent batteries, and minimizing the carbon footprint and social costs of battery production[1][2][7].\textbf{Electric car batteries have many advantages over conventional fossil fuel vehicles, such as lower greenhouse gas emissions, lower operating costs, and higher energy efficiency.}}
        & \multicolumn{1}{>{\justifying\noindent}p{6cm}}{One advantage of electric car batteries, used in electric vehicles, is that they have lower operating costs compared to conventional fossil fuel vehicles.} 
        & \multicolumn{1}{>{\justifying\noindent}p{6cm}}{One advantage of electric car batteries, used in electric vehicles, is that they have lower operating costs compared to conventional fossil fuel vehicles {\sethlcolor{lightred}\hl{powered by internal combustion engines}}.} \\
        \midrule

        \multicolumn{1}{>{\centering\arraybackslash}p{1.5cm}}{2} 
        & \multicolumn{1}{>{\justifying\noindent}p{5cm}}{What is the historical and cultural significance behind the traditional African drumming styles, such as \sethlcolor{lightgreen}\hl{djembe} and sabar, and how have they influenced modern music genres across the globe?}
        & \multicolumn{1}{>{\justifying\noindent}p{7cm}}{That's a great question! Traditional African drumming styles have a rich and diverse history and culture, and they have influenced many modern music genres across the globe. Here are some key points to know:\newline\textbf{- Traditional African drums are usually made of wood and animal skins, and they are played with bare hands or sticks.}}
        & \multicolumn{1}{>{\justifying\noindent}p{6cm}}{Traditional African drums are played with bare hands.} 
        & \multicolumn{1}{>{\justifying\noindent}p{6cm}}{Traditional African drums, such as \sethlcolor{lightgreen}\hl{djembe}, {\sethlcolor{lightred}\hl{udu, bowl drum, talking drum, bata, and bougarabou}}, are played with bare hands.} \\
        \midrule

        \multicolumn{1}{>{\centering\arraybackslash}p{1.5cm}}{3} 
        & \multicolumn{1}{>{\justifying\noindent}p{5cm}}{Can you explain the process and technology behind the development of electric cars, including their batteries, charging infrastructure, and potential impact on the environment and transportation industry?}
        & \multicolumn{1}{>{\justifying\noindent}p{7cm}}{- Charging: This is the process of replenishing the battery with electricity from an external source.  \textbf{Charging can be done using different methods and standards, such as AC charging (using a regular wall outlet or a dedicated EVSE), DC fast charging (using a high-power station that can charge up to 80\% in 30 minutes), wireless charging (using electromagnetic induction or resonance), or battery swapping (replacing a depleted battery with a fully charged one).}}
        & \multicolumn{1}{>{\justifying\noindent}p{6cm}}{AC charging uses a dedicated EVSE for electric vehicle charging.} 
        & \multicolumn{1}{>{\justifying\noindent}p{6cm}}{AC charging uses a dedicated {\sethlcolor{lightred}\hl{Electric Vehicle Supply Equipment}} (EVSE) for electric vehicle charging.} \\
        \midrule

        \multicolumn{1}{>{\centering\arraybackslash}p{1.5cm}}{4} 
        & \multicolumn{1}{>{\justifying\noindent}p{5cm}}{What is the process and criteria for selecting the winners of the Pulitzer Prize, and how has this evolved over time \sethlcolor{lightred}\hl{since its inception in 1917}?}
        & \multicolumn{1}{>{\justifying\noindent}p{7cm}}{The Pulitzer Prize has evolved over time \sethlcolor{lightred}\hl{since its inception in 1917}. Some of the changes include:\newline \textbf{- Adding new categories such as photography, criticism, editorial cartooning, feature writing, commentary, biography, history, poetry, music, and drama[5].}}
        & \multicolumn{1}{>{\justifying\noindent}p{6cm}}{Feature writing is a new category added to the Pulitzer Prize.} 
        & \multicolumn{1}{>{\justifying\noindent}p{6cm}}{$c$ is already maximally decontextualized.} \\
        \bottomrule
    \end{tabular}
    } 
    
    \caption{Examples of errors in generating $c_{\max}$, the maximally decontextualized version of claim $c$. In Examples 1-3, $c_{\max}$ introduced information that is not entailed by the combined question, context, and $c$. In Example 4, $c$ was deemed maximally decontextualized, but it does not include the highlighted context. The source sentence for each claim is bolded in the ``Context'' column.}
    \label{tab:decontext_errors}
\end{table*}

\section{Computational Resources}
Generating outputs for Claimify and all methods described in \autoref{subsec:methods} took approximately 10 hours. Generating evaluation outputs (\autoref{sec:experiments}) took approximately 145 hours. These processes ran on a machine with 32GB RAM and an Intel Core i7-11370H CPU @ 3.30GHz (8 CPUs).

\clearpage

\onecolumn
\section{Prompts}

\subsection{Claimify}
\label{app:claimify_prompts}
\subsubsection{Selection}
\begin{tcolorbox}[
    breakable,                    
    colback=white,                
    colframe=black,              
    title=Selection System Prompt,       
    title after break=Selection System Prompt (Continued),
    fonttitle=\bfseries, 
    coltext=black,
    after skip =-2pt
]
\begin{lstlisting}[breaklines=true, breakindent=0pt, basicstyle=\small\ttfamily\raggedright, xleftmargin=-5pt, frame=none, xrightmargin=-5pt, aboveskip=-2pt, belowskip=-2pt]
You are an assistant to a fact-checker. You will be given a question, which was asked about a source text (it may be referred to by other names, e.g., a 
dataset). You will also be given an excerpt from a response to the question. If it contains "[...]", this means that you are NOT seeing all sentences in the response. You will also be given a particular sentence of interest from the response. Your task is to determine whether this particular sentence contains at least one specific and verifiable proposition, and if so, to return a complete sentence that only contains verifiable information.   

Note the following rules:
- If the sentence is about a lack of information, e.g., the dataset does not contain information about X, then it does NOT contain a specific and verifiable proposition.
- It does NOT matter whether the proposition is true or false.
- It does NOT matter whether the proposition is relevant to the question.
- It does NOT matter whether the proposition contains ambiguous terms, e.g., a pronoun without a clear antecedent. Assume that the fact-checker has the necessary information to resolve all ambiguities.
- You will NOT consider whether a sentence contains a citation when determining if it has a specific and verifiable proposition.

You must consider the preceding and following sentences when determining if the sentence has a specific and verifiable proposition. For example:
- if preceding sentence = "Who is the CEO of Company X?" and sentence = "John" then sentence contains a specific and verifiable proposition.
- if preceding sentence = "Jane Doe introduces the concept of regenerative technology" and sentence = "It means using technology to restore ecosystems" then sentence contains a specific and verifiable proposition.
- if preceding sentence = "Jane is the President of Company Y" and sentence = "She has increased its revenue by 20\%" then sentence contains a specific and verifiable proposition.
- if sentence = "Guests interviewed on the podcast suggest several strategies for fostering innovation" and the following sentences expand on this point 
(e.g., give examples of specific guests and their statements), then sentence is an introduction and does NOT contain a specific and verifiable proposition.
- if sentence = "In summary, a wide range of topics, including new technologies, personal development, and mentorship are covered in the dataset" and the preceding sentences provide details on these topics, then sentence is a conclusion and does NOT contain a specific and verifiable proposition.

Here are some examples of sentences that do NOT contain any specific and verifiable propositions:
- By prioritizing ethical considerations, companies can ensure that their innovations are not only groundbreaking but also socially responsible
- Technological progress should be inclusive
- Leveraging advanced technologies is essential for maximizing productivity
- Networking events can be crucial in shaping the paths of young entrepreneurs and providing them with valuable connections
- AI could lead to advancements in healthcare
- This implies that John Smith is a courageous person

Here are some examples of sentences that likely contain a specific and verifiable proposition and how they can be rewritten to only include verifiable information:
- The partnership between Company X and Company Y illustrates the power of innovation -> "There is a partnership between Company X and Company Y"
- Jane Doe's approach of embracing adaptability and prioritizing customer feedback can be valuable advice for new executives -> "Jane Doe's approach includes embracing adaptability and prioritizing customer feedback"
- Smith's advocacy for renewable energy is crucial in addressing these challenges -> "Smith advocates for renewable energy"
- **John Smith**: instrumental in numerous renewable energy initiatives, playing a pivotal role in Project Green -> "John Smith participated in renewable energy initiatives, playing a role in Project Green"
- The technology is discussed for its potential to help fight climate change -> remains unchanged
- John, the CEO of Company X, is a notable example of effective leadership -> 
"John is the CEO of Company X"
- Jane emphasizes the importance of collaboration and perseverance -> remains unchanged
- The Behind the Tech podcast by Kevin Scott is an insightful podcast that explores the themes of innovation and technology -> "The Behind the Tech podcast by Kevin Scott is a podcast that explores the themes of innovation and technology"
- Some economists anticipate the new regulation will immediately double production costs, while others predict a gradual increase -> remains unchanged
- AI is frequently discussed in the context of its limitations in ethics and privacy -> "AI is discussed in the context of its limitations in ethics and privacy"
- The power of branding is highlighted in discussions featuring John Smith and Jane Doe -> remains unchanged
- Therefore, leveraging industry events, as demonstrated by Jane's experience at the Tech Networking Club, can provide visibility and traction for new ventures -> "Jane had an experience at the Tech Networking Club, and her experience involved leveraging an industry event to provide visibility and traction for a new venture"

Your output must adhere to the following format exactly. Only replace what's inside the <insert> tags; do NOT remove the step headers.  
Sentence: 
<insert> 

4-step stream of consciousness thought process (1. reflect on criteria at a high-level -> 2. provide an objective description of the excerpt, the sentence, and its surrounding sentences -> 3. consider all possible perspectives on whether the sentence explicitly or implicitly contains a specific and verifiable proposition, or if it just contains an introduction for the following 
sentence(s), a conclusion for the preceding sentence(s), broad or generic statements, opinions, interpretations, speculations, statements about a lack of information, etc. -> 4. only if it contains a specific and verifiable proposition: reflect on whether any changes are needed to ensure that the entire sentence only contains verifiable information):
<insert>

Final submission:
<insert 'Contains a specific and verifiable proposition' or 'Does NOT contain a specific and verifiable proposition'>

Sentence with only verifiable information:
<insert changed sentence, or 'remains unchanged' if no changes, or 'None' if the sentence does NOT contain a specific and verifiable proposition>
\end{lstlisting}
\end{tcolorbox}

\begin{tcolorbox}[
    breakable,                    
    colback=white,                
    colframe=black,              
    title=Selection User Prompt,       
    title after break=Selection User Prompt (Continued),
    fonttitle=\bfseries, 
    coltext=black,
    before skip=6pt,
]
\begin{lstlisting}[breaklines=true, breakindent=0pt, basicstyle=\small\ttfamily\raggedright, xleftmargin=-5pt, frame=none, xrightmargin=-5pt, aboveskip=-2pt, belowskip=-2pt]
Question:
{question}

Excerpt:
{excerpt}

Sentence:
{sentence}
\end{lstlisting}
\end{tcolorbox}

\subsubsection{Disambiguation}
\begin{tcolorbox}[
    breakable,                    
    colback=white,                
    colframe=black,              
    title=Disambiguation System Prompt,       
    title after break=Disambiguation System Prompt (Continued),
    fonttitle=\bfseries, 
    coltext=black,
]
\begin{lstlisting}[breaklines=true, breakindent=0pt, basicstyle=\small\ttfamily\raggedright, xleftmargin=-5pt, frame=none, xrightmargin=-5pt, aboveskip=-2pt, belowskip=-2pt]
You are an assistant to a fact-checker. You will be given a question, which was asked about a source text (it may be referred to by other names, e.g., a 
dataset). You will also be given an excerpt from a response to the question. If it contains "[...]", this means that you are NOT seeing all sentences in the response. You will also be given a particular sentence from the response. The text before and after this sentence will be referred to as "the context". Your task is to "decontextualize" the sentence, which means:
1. determine whether it's possible to resolve partial names and undefined acronyms/abbreviations in the sentence using the question and the context; if it is possible, you will make the necessary changes to the sentence
2. determine whether the sentence in isolation contains linguistic ambiguity that has a clear resolution using the question and the context; if it does, you will make the necessary changes to the sentence

Note the following rules:
- "Linguistic ambiguity" refers to the presence of multiple possible meanings in a sentence. Vagueness and generality are NOT linguistic ambiguity. Linguistic ambiguity includes referential and structural ambiguity. Temporal ambiguity is a type of referential ambiguity.
- If it is unclear whether the sentence is directly answering the question, you should NOT count this as linguistic ambiguity. You should NOT add any information to the sentence that assumes a connection to the question.
- If a name is only partially given in the sentence, but the full name is provided in the question or the context, the DecontextualizedSentence must always use the full name. The same rule applies to definitions for acronyms and abbreviations. However, the lack of a full name or a definition for an acronym/abbreviation in the question and the context does NOT count as linguistic ambiguity; in this case, you will just leave the name, acronym, or abbreviation as is.
- Do NOT include any citations in the DecontextualizedSentence.
- Do NOT use any external knowledge beyond what is stated in the question, context, and sentence.

Here are some correct examples that you should pay attention to:
1. Question = "Describe the history of TurboCorp", Context = "John Smith was an early employee who transitioned to management in 2010", Sentence = "At the time, he led the company's operations and finance teams."
    - For referential ambiguity, "At the time", "he", and "the company's" are unclear. A group of readers shown the question and the context would likely reach consensus about the correct interpretation: "At the time" corresponds to 2010, "he" refers to John Smith, and "the company's" refers to TurboCorp.
    - DecontextualizedSentence: In 2010, John Smith led TurboCorp's operations and finance teams.
2. Question = "Who are notable executive figures?", Context = "[...]**Jane Doe**", Sentence = "These notes indicate that her leadership at TurboCorp and MiniMax is accelerating progress in renewable energy and sustainable 
agriculture."
    - For referential ambiguity, "these notes" and "her" are unclear. A group of readers shown the question and the context would likely fail to reach consensus about the correct interpretation of "these notes", since there is no indication in the question or context. However, they would likely reach consensus about the correct interpretation of "her": Jane Doe.
    - For structural ambiguity, the sentence could be interpreted as: (1) Jane's leadership is accelerating progress in renewable energy and sustainable agriculture at both TurboCorp and MiniMax, (2) Jane's leadership is accelerating progress in renewable energy at TurboCorp and in sustainable agriculture at MiniMax. A group of readers shown the question and the context would likely fail to reach consensus about the correct interpretation of this ambiguity.
    - DecontextualizedSentence: Cannot be decontextualized
3. Question = "Who founded MiniMax?", Context = "None", Sentence = "Executives like John Smith were involved in the early days of MiniMax."
    - For referential ambiguity, "like John Smith" is unclear. A group of readers shown the question and the context would likely reach consensus about the correct interpretation: John Smith is an example of an executive who was involved in the early days of MiniMax.
    - Note that "Involved in" and "the early days" are vague, but they are NOT linguistic ambiguity.
    - DecontextualizedSentence: John Smith is an example of an executive who was involved in the early days of MiniMax.
4. Question = "What advice is given to young entrepreneurs?", Context = 
"# Ethical Considerations", Sentence = "Sustainable manufacturing, as emphasized by John Smith and Jane Doe, is critical for customer buy-in and long-term success."
    - For structural ambiguity, the sentence could be interpreted as: (1) John Smith and Jane Doe emphasized that sustainable manufacturing is critical for customer buy-in and long-term success, (2) John Smith and Jane Doe emphasized sustainable manufacturing while the claim that sustainable manufacturing is critical for customer buy-in and long-term success is attributable to the writer, not to John Smith and Jane Doe. A group of readers shown the question and the context would likely fail to reach consensus about the correct interpretation of this ambiguity.
    - DecontextualizedSentence: Cannot be decontextualized
5. Question = "What are common strategies for building successful teams?", Context = "One of the most common strategies is creating a diverse team.", Sentence = "Last winter, John Smith highlighted the importance of interdisciplinary discussions and collaborations, which can drive advancements by integrating diverse perspectives from fields such as artificial intelligence, genetic engineering, and statistical machine learning."
    - For referential ambiguity, "Last winter" is unclear. A group of readers shown the question and the context would likely fail to reach consensus about the correct interpretation of this ambiguity, since there is no indication of the time period in the question or context.
    - For structural ambiguity, the sentence could be interpreted as: (1) John Smith highlighted the importance of interdisciplinary discussions and collaborations and that they can drive advancements by integrating diverse perspectives from some example fields, (2) John Smith only highlighted the importance of interdisciplinary discussions and collaborations while the claim that they can drive advancements by integrating diverse perspectives from some example fields is attributable to the writer, not to John Smith. A group of readers shown the question and the context would likely fail to reach consensus about the correct interpretation of this ambiguity.
    - DecontextualizedSentence: Cannot be decontextualized
6. Question = "What opinions are provided on disruptive technologies?", Context = "[...]However, there is a divergence in how to weigh short-term benefits against long-term risks.", Sentence = "These differences are illustrated by the discussion on healthcare: some stress AI's benefits, while others highlight its risks, such as privacy and data security."
    - For referential ambiguity, "These differences" is unclear. A group of readers shown the question and the context would likely reach consensus about the correct interpretation: the differences are with respect to how to weigh short-term benefits against long-term risks.
    - For structural ambiguity, the sentence could be interpreted as: (1) privacy and data security are examples of risks, (2) privacy and data security are examples of both benefits and risks. A group of readers shown the question and the context would likely reach consensus about the correct interpretation: privacy and data security are examples of risks.
    - Note that "Some" and "others" are vague, but they are not linguistic ambiguity.
    - DecontextualizedSentence: The differences in how to weigh short-term benefits against long-term risks are illustrated by the discussion on healthcare. Some experts stress AI's benefits with respect to healthcare. Other experts highlight AI's risks with respect to healthcare, such as privacy and data security.

First, print "Incomplete Names, Acronyms, Abbreviations:" followed by your step-by-step reasoning for determining whether the Sentence contains any partial names and undefined acronyms/abbreviations. If the full names and definitions are provided in the question or context, the Sentence will be updated accordingly; otherwise, they will be left as is and they will NOT count as linguistic ambiguity. Next, print "Linguistic Ambiguity in '<insert the 
sentence>':" followed by your step-by-step reasoning for checking (1) referential and (2) structural ambiguity (and note that 1. referential ambiguity is NOT equivalent to vague or general language and it includes temporal ambiguity, and 2. structural reasoning must follow "The sentence could be interpreted as: <insert one or multiple interpretations>"), then considering whether a group of readers shown the question and the context would likely reach consensus or fail to reach consensus about the correct interpretation of the linguistic ambiguity. If they would likely fail to reach consensus, print
"DecontextualizedSentence: Cannot be decontextualized"; otherwise, first print
"Changes Needed to Decontextualize the Sentence:" followed by a list of all changes needed to ensure the Sentence is fully decontextualized (e.g., replace 
"executives like John Smith" with "John Smith is an example of an executive who") and includes all full names and definitions for acronyms/abbreviations (only if they were provided in the question and the context), then print 
"DecontextualizedSentence:" followed by the final sentence (or collection of sentences) that implements all changes.
\end{lstlisting}
\end{tcolorbox}

\begin{tcolorbox}[
    breakable,                    
    colback=white,                
    colframe=black,              
    title=Disambiguation User Prompt,       
    title after break=Disambiguation User Prompt (Continued),
    fonttitle=\bfseries, 
    coltext=black,
]
\begin{lstlisting}[breaklines=true, breakindent=0pt, basicstyle=\small\ttfamily\raggedright, xleftmargin=-5pt, frame=none, xrightmargin=-5pt, aboveskip=-2pt, belowskip=-2pt]
Question:
{question}

Excerpt:
{excerpt}

Sentence:
{sentence}
\end{lstlisting}
\end{tcolorbox}

\subsubsection{Decomposition}
\begin{tcolorbox}[
    breakable,                    
    colback=white,                
    colframe=black,              
    title=Decomposition System Prompt,       
    title after break=Decomposition System Prompt (Continued),
    fonttitle=\bfseries, 
    coltext=black,
]
\begin{lstlisting}[breaklines=true, breakindent=0pt, basicstyle=\small\ttfamily\raggedright, xleftmargin=-5pt, frame=none, xrightmargin=-5pt, aboveskip=-2pt, belowskip=-2pt]
You are an assistant for a group of fact-checkers. You will be given a question, which was asked about a source text (it may be referred to by other names, 
e.g., a dataset). You will also be given an excerpt from a response to the question. If it contains "[...]", this means that you are NOT seeing all sentences in the response. You will also be given a particular sentence from the response. The text before and after this sentence will be referred to as "the context".

Your task is to identify all specific and verifiable propositions in the sentence and ensure that each proposition is decontextualized. A proposition is "decontextualized" if (1) it is fully self-contained, meaning it can be understood in isolation (i.e., without the question, the context, and the other propositions), AND (2) its meaning in isolation matches its meaning when interpreted alongside the question, the context, and the other propositions. The propositions should also be the simplest possible discrete units of 
information.

Note the following rules:
- Here are some examples of sentences that do NOT contain a specific and verifiable proposition:
    - By prioritizing ethical considerations, companies can ensure that their innovations are not only groundbreaking but also socially responsible
    - Technological progress should be inclusive
    - Leveraging advanced technologies is essential for maximizing productivity
    - Networking events can be crucial in shaping the paths of young entrepreneurs and providing them with valuable connections
    - AI could lead to advancements in healthcare
- Sometimes a specific and verifiable proposition is buried in a sentence that is mostly generic or unverifiable. For example, "John's notable research on neural networks demonstrates the power of innovation" contains the specific and verifiable proposition "John has research on neural networks". Another example is "TurboCorp exemplifies the positive effects that prioritizing ethical considerations over profit can have on innovation" where the specific and verifiable proposition is "TurboCorp prioritizes ethical considerations over profit".
- If the sentence indicates that a specific entity said or did something, it is critical that you retain this context when creating the propositions. For example, if the sentence is "John highlights the importance of transparent communication, such as in Project Alpha, which aims to double customer satisfaction by the end of the year", the propositions would be ["John highlights the importance of transparent communication", "John highlights Project Alpha as an example of the importance of transparent communication", 
"Project Alpha aims to double customer satisfaction by the end of the year"]. The propositions "transparent communication is important" and "Project Alpha is an example of the importance of transparent communication" would be incorrect since they omit the context that these are things John highlights. However, the last part of the sentence, "which aims to double customer satisfaction by the end of the year", is not likely a statement made by John, so it can be its own proposition. Note that if the sentence was something like "John's career underscores the importance of transparent communication", it's NOT about what John says or does but rather about how John's career can be interpreted, which is NOT a specific and verifiable proposition.
- If the context contains "[...]", we cannot see all preceding statements, so we do NOT know for sure whether the sentence is directly answering the question. It might be background information for some statements we can't see. Therefore, you should only assume the sentence is directly answering the question if this is strongly implied.
- Do NOT include any citations in the propositions.
- Do NOT use any external knowledge beyond what is stated in the question, context, and sentence.

Here are some correct examples that you must pay attention to:
1. Question = "Describe the history of TurboCorp", Context = "John Smith was an early employee who transitioned to management in 2010", Sentence = "At the time, John Smith, led the company's operations and finance teams"
    - MaxClarifiedSentence = In 2010, John Smith led TurboCorp's operations team and finance team. 
    - Specific, Verifiable, and Decontextualized Propositions: ["In 2010, John Smith led TurboCorp's operations team", "In 2010, John Smith led TurboCorp's finance team"]
2. Question = "What do technologists think about corporate responsibility?", Context = "[...]## Activism", Sentence = "Many notable sustainability leaders like Jane do not work directly for a corporation, but her organization CleanTech has powerful partnerships with technology companies (e.g., MiniMax) to significantly improve waste management, demonstrating the power of
collaboration."
    - MaxClarifiedSentence = Jane is an example of a notable sustainability leader, and she does not work directly for a corporation, and this is true for many notable sustainability leaders, and Jane has an organization called CleanTech, and CleanTech has powerful partnerships with technology companies to significantly improve waste management, and MiniMax is an example of a technology company that CleanTech has a partnership with to improve waste management, and this demonstrates the power of collaboration.
    - Specific, Verifiable, and Decontextualized Propositions: ["Jane is a sustainability leader", "Jane does not work directly for a corporation", 
    "Jane has an organization called CleanTech", "CleanTech has partnerships with technology companies to improve waste management", "MiniMax is a technology company", "CleanTech has a partnership with MiniMax to improve waste management"]
3. Question = "What are the key topics?", Context = "The power of mentorship and networking:", "Sentence = "Extensively discussed by notable figures such as John Smith and Jane Doe, who highlight their potential to have substantial benefits for people's careers, like securing promotions and raises"
    - MaxClarifiedSentence = John Smith and Jane Doe discuss the potential of mentorship and networking to have substantial benefits for people's careers, and securing promotions and raises are examples of potential benefits that are discussed by John Smith and Jane Doe.
    - Specific, Verifiable, and Decontextualized Propositions: ["John Smith discusses the potential of mentorship to have substantial benefits for people's careers", "Jane Doe discusses the potential of networking to have substantial benefits for people's careers", "Jane Doe discusses the potential of mentorship to have substantial benefits for people's careers", "Jane Doe discusses the potential of networking to have substantial benefits for people's careers", "Securing promotions is an example of a potential benefit of mentorship that is discussed by John Smith", "Securing raises is an example of a potential benefit of mentorship that is discussed by John Smith", 
    "Securing promotions is an example of a potential benefit of networking that is discussed by John Smith", "Securing raises is an example of a potential benefit of networking that is discussed by John Smith", "Securing promotions is an example of a potential benefit of mentorship that is discussed by Jane Doe", "Securing raises is an example of a potential benefit of mentorship that is discussed by Jane Doe", "Securing promotions is an example of a potential benefit of networking that is discussed by Jane Doe", "Securing raises is an example of a potential benefit of networking that is discussed by Jane Doe"]
4. Question = "What is the status of global trade relations?", Context = "[...]**US & China**", Sentence = "Trade relations have mostly suffered since the introduction of tariffs, quotas, and other protectionist measures, underscoring the importance of international cooperation."
    - MaxClarifiedSentence = US-China trade relations have mostly suffered since the introduction of tariffs, quotas, and other protection measures, and this underscores the importance of international cooperation.
    - Specific, Verifiable, and Decontextualized Propositions: ["US-China trade relations have mostly suffered since the introduction of tariffs", "US-China trade relations have mostly suffered since the introduction of quotas", "US-China trade relations have mostly suffered since the introduction of protectionist measures besides tariffs and quotas"]
5. Question = "Provide an overview of environmental activists", Context = 
"- Jill Jones", Sentence = "- John Smith and Jane Doe (writers of 'Fighting for Better Tech')"
    - MaxClarifiedSentence = John Smith and Jane Doe are writers of 'Fighting for Better Tech'.
    - Decontextualized Propositions: ["John Smith is a writer of 'Fighting for Better Tech'", "Jane Doe is a writer of 'Fighting for Better Tech'"]
6. Question = "What are the experts' opinions on disruptive technologies?", Context = "[...]However, there is a divergence in how to weigh short-term benefits against long-term risks.", Sentence = "These differences are illustrated by the discussion on healthcare: John Smith stresses AI's importance in improving patient outcomes, while others highlight its risks, such as privacy and data security"
    - MaxClarifiedSentence = John Smith stresses AI's importance in improving patient outcomes, and some experts excluding John Smith highlight AI's risks in healthcare, and privacy and data security are examples of AI's risks in healthcare that they highlight.
    - Specific, Verifiable, and Decontextualized Propositions: ["John Smith stresses AI's importance in improving patient outcomes", "Some experts excluding John Smith highlight AI's risks in healthcare", "Some experts excluding John Smith highlight privacy as a risk of AI in healthcare", "Some experts excluding John Smith highlight data security as a risk of AI in healthcare"]
7. Question = "How can startups improve profitability?" Context = "# Case Studies", Sentence = "Monetizing distribution channels, as demonstrated by MiniMax's experience with the exciting launch of Buzz, can be effective strategy for increasing revenue"
    - MaxClarifiedSentence = MiniMax experienced the launch of Buzz, and this experience demonstrates that monetizing distribution channels can be an effective strategy for increasing revenue.
    - Specific, Verifiable, and Decontextualized Propositions: ["MiniMax experienced the launch of Buzz", "MiniMax's experience with the launch of Buzz demonstrated that monetizing distribution channels can be an effective strategy for increasing revenue"]
8. Question = "What steps have been taken to promote corporate social responsibility?", Context = "In California, the Energy Commission identifies and sanctions companies that fail to meet the state's environmental standards." Sentence = "In 2023, its annual report identified 350 failing companies who will be required spend 2% of their profits on carbon credits, renewable energy projects, or reforestation efforts."
    - MaxClarifiedSentence = In 2023, the California Energy Commission's annual report identified 350 companies that failed to meet California's environmental standards, and the 350 failing companies will be required to spend 2% of their profits on carbon credits, renewable energy projects, or reforestation efforts.
    - Specific, Verifiable, and Decontextualized Propositions: ["In 2023, the California Energy Commission's annual report identified 350 companies that failed to meet the state's environmental standards", "The failing companies identified in the California Energy Commission's 2023 annual report will be required to spend 2% of their profits on carbon credits, renewable energy projects, or reforestation efforts"]
9. Question = "Explain the role of government in funding schools", Context = 
"California's senate has proposed a new bill to modernize schools.", Sentence = 
"The senate points out that its bill, which aims to ensure that all students have access to the latest technologies, recommends the government provide funding for schools to purchase new equipment, including computers and tablets, when they submit evidence that their current equipment is outdated."
    - MaxClarifiedSentence = California's senate points out that its bill to modernize schools recommends the government provide funding for schools to purchase new equipment when they submit evidence that their current equipment is outdated, and computers and tablets are examples of new equipment, and the bill's aim is to ensure that all students have access to the latest technologies.
    - Specific, Verifiable, and Decontextualized Propositions: ["California's senate's bill to modernize schools recommends the government provide funding for schools to purchase new equipment when they submit evidence that their current equipment is outdated", "Computers are examples of new equipment that the California senate's bill to modernize schools recommends the government provide funding for", "Tablets are examples of new equipment that the California senate's bill to modernize schools recommends the government provide funding for", "The aim of the California senate's bill to modernize schools is to ensure that all students have access to the latest technologies"]
10. Question = "What companies are profiled?", Context = "John Smith and Jane Doe, the duo behind Youth4Tech, provides coaching for young founders.", Sentence = "Their guidance and decision-making have been pivotal in the growth of numerous successful startups, such as TurboCorp and MiniMax."
    - MaxClarifiedSentence = The guidance and decision-making of John Smith and Jane Doe have been pivotal in the growth of successful startups, and TurboCorp and MiniMax are examples of successful startups that John Smith and Jane Doe's guidance and decision-making have been pivotal in.
    - Specific, Verifiable, and Decontextualized Propositions: ["John Smith's guidance has been pivotal in the growth of successful startups", 
    "John Smith's decision-making has been pivotal in the growth of successful startups", "Jane Doe's guidance has been pivotal in the growth of successful startups", "Jane Doe's decision-making has been pivotal in the growth of successful startups", "TurboCorp is a successful startup", "MiniMax is a successful startup", "John Smith's guidance has been pivotal in the growth of TurboCorp", "John Smith's decision-making has been pivotal in the growth of TurboCorp", "John Smith's guidance has been pivotal in the growth of MiniMax", "John Smith's decision-making has been pivotal in the growth of MiniMax", "Jane Doe's guidance has been pivotal in the growth of TurboCorp", "Jane Doe's decision-making has been pivotal in the growth of TurboCorp", 
    "Jane Doe's guidance has been pivotal in the growth of MiniMax", "Jane Doe's decision-making has been pivotal in the growth of MiniMax"]

First, print "Sentence:" followed by the sentence, Then print "Referential terms whose referents must be clarified (e.g., "other"):" followed by an overview of all terms in the sentence that explicitly or implicitly refer to other terms in the sentence, (e.g., "other" in "the Department of Education, the Department of Defense, and other agencies" refers to the Department of Education and the Department of Defense; "earlier" in "unlike the 2023 annual report, earlier reports" refers to the 2023 annual report) or None if there are no referential terms, Then print "MaxClarifiedSentence:" which articulates discrete units of information made by the sentence and clarifies referents, Then print "The range of the possible number of propositions (with some margin for variation) is:" followed by X-Y where X can be 0 or greater and X and Y must be different integers. Then print "Specific, Verifiable, and Decontextualized Propositions:" followed by a list of all propositions that are each specific, verifiable, and fully decontextualized. Use the format below:
[
"insert a specific, verifiable, and fully decontextualized proposition",
]
Next, it is EXTREMELY important that you consider that each fact-checker in the group will only have access to one of the propositions - they will not have access to the question, the context, and the other propositions. Print 
"Specific, Verifiable, and Decontextualized Propositions with Essential Context/Clarifications:" followed by a final list of instructions for the fact-checkers with **all essential clarifications and context** enclosed in square brackets: [...]. For example, the proposition "The local council expects its law to pass in January 2025" might become "The [Boston] local council expects its law 
[banning plastic bags] to pass in January 2025 - true or false?"; the proposition "Other agencies decreased their deficit" might become "Other agencies [besides the Department of Education and the Department of Defense] increased their deficit [relative to 2023] - true or false?"; the proposition 
"The CGP has called for the termination of hostilities" might become "The CGP 
[Committee for Global Peace] has called for the termination of hostilities [in the context of a discussion on the Middle East] - true or false?". Use the format below:
[
"<insert a specific, verifiable, and fully decontextualized proposition with as few or as many [...] as needed> - true or false?",
]
\end{lstlisting}
\end{tcolorbox}

\begin{tcolorbox}[
    breakable,                    
    colback=white,                
    colframe=black,              
    title=Decomposition User Prompt,       
    title after break=Decomposition User Prompt (Continued),
    fonttitle=\bfseries, 
    coltext=black,
]
\begin{lstlisting}[breaklines=true, breakindent=0pt, basicstyle=\small\ttfamily\raggedright, xleftmargin=-5pt, frame=none, xrightmargin=-5pt, aboveskip=-2pt, belowskip=-2pt]
Question:
{question}

Excerpt:
{excerpt}

Sentence:
{sentence}
\end{lstlisting}
\end{tcolorbox}

\subsection{Evaluation Framework}
\label{app:eval_prompts}
\subsubsection{Entailment}
\begin{tcolorbox}[
    breakable,                    
    colback=white,                
    colframe=black,              
    title=Entailment System Prompt,       
    title after break=Entailment System Prompt (Continued),
    fonttitle=\bfseries, 
    coltext=black,
]
\begin{lstlisting}[breaklines=true, breakindent=0pt, basicstyle=\small\ttfamily\raggedright, xleftmargin=-5pt, frame=none, xrightmargin=-5pt, aboveskip=-2pt, belowskip=-2pt]
## Overview
You will be given a question, an excerpt from the response to the question, a sentence of interest from the excerpt (which will be referred to as S), and a claim (which will be referred to as C).

A sentence entails a claim if when the sentence is true, the claim must also be true. Your task is to determine whether S entails C by following these steps:
1. Print "S = <insert sentence of interest here EXACTLY as written>"
2. Describe the context for S; if someone read S in this context, how would they interpret it?
3. Print "C = <insert claim of interest here EXACTLY as written>" How would a reader interpret the claim?
4. What are ALL elements of C? It's possible there's only one element. Even if you have external information that some elements of C are true, you must still list them. For example, if C is "Paris, the capital of France, was the most visited city in the world in 2019", the elements are (1) Paris was the most visited city in the world in 2019, (2) Paris is the capital of France.
5. Does the Statements and Actions Rule apply to S, or does it qualify as an exception? See the description of the rule and its exceptions below.
6. Ask yourself for each element of C: If <insert maximally clarified version of S given its context>, does this necessarily mean that <insert element of C, as a reader would interpret it in isolation>? Then respond with: <insert step-by-step reasoning>, so <insert yes or no>. You CANNOT use any external information (e.g., if an element says "John is a politician" but the claim does not mention that John is a politician, even if you have external information that John is a politician, the element is NOT entailed by the claim). Finally, print either "S entails all elements of C" or "S does not entail all elements of C". IMPORTANT: if the context of S entails C, but S itself does not, you should still conclude that S entails C.

If the sentence is something like "John found X", "John reported X", "John emphasizes X", etc. (where John can be replaced with any entity or entities), it should be interpreted as a statement about what John says or does. For example, if the sentence is "John highlights that transparent communication is a critical part of Project Alpha", it does NOT entail the claim "transparent communication is a critical part of Project Alpha" because it's missing the critical context that this is something John highlights. Let's call this the Statements and Actions Rule. The ONLY exceptions to this rule are: (1) if the sentence says something like "According to <insert citation>" or "Based on the search results" (i.e., the responder is attributing the information to an undefined source), and (2) if the sentence says something like "I know the following information" (i.e. the responder is attributing the information to themselves); in both cases, you should IGNORE the attribution and treat it as a regular statement.

## Examples
### Example 1
Question: What are the rules for Bright Futures participations?
Excerpt from response: The program selects students based on their grades, test scores, and extracurricular activities. Admitted students are matched with a mentor who helps them navigate the college application process. They are required to complete 100 hours of volunteering, summer school, or job training.
Sentence of interest: They are required to complete 100 hours of volunteering, summer school, or job training.
Claim: Students admitted to the Bright Futures program are required to complete 100 hours of volunteering.

S = They are required to complete 100 hours of volunteering, summer school, or job training.
Describe the context for S; if someone read S in this context, how would they interpret it? The question is about the rules for Bright Futures participations, and the excerpt discusses admitted students. Therefore, S would likely be interpreted as students admitted to the Bright Futures program must do one of the following: complete 100 hours of volunteering, or summer school, or job training.
C = Students admitted to the Bright Futures program are required to complete 100 hours of volunteering
A reader would interpret the claim as the Bright Futures program requires students to complete 100 hours of volunteering alone.
What are ALL elements of C? (1) The Bright Futures program requires students to complete 100 hours of volunteering alone.
Does the Statements and Actions Rule apply to S, or does it qualify as an exception? S is not about an entity's actions or statements, so it does not apply.
If students admitted to the Bright Futures program can fulfill the requirement by completing 100 hours of volunteering, summer school, or job training, does this necessarily mean that they are required to complete 100 hours of volunteering alone? Volunteering is just one option to fulfill the requirement, so no. Therefore, S does not entail all elements of C.
 
### Example 2
Question: Provide an overview of the media's portrayal of AI.
Excerpt from response: ## Case Study 2
Another example is the discussion on the Behind the Tech podcast about GitHub Copilot boosting developers' productivity.
Sentence of interest: Another example is the discussion on the Behind the Tech podcast about GitHub Copilot boosting developers' productivity.
Claim: GitHub Copilot boosts developers' productivity.

S = Another example is the discussion on the Behind the Tech podcast about GitHub Copilot boosting developers' productivity.
Describe the context for S; if someone read S in this context, how would they interpret it? The question is about the media's portrayal of AI, and the excerpt provides an example of such portrayal. Therefore, S would likely be interpreted as there is a discussion on the Behind the Tech about GitHub Copilot boosting developers' productivity.
C = GitHub Copilot, a tool developed by Microsoft, boosts developers' productivity.
A reader would interpret the claim as GitHub Copilot, which is a tool developed by Microsoft, boosts developers' productivity.
What are ALL elements of C? (1) GitHub Copilot boosts developers' productivity, (2) GitHub Copilot is a tool developed by Microsoft.
Does the Statements and Actions Rule apply to S, or does it qualify as an exception? S is a statement about what was discussed on the Behind the Tech podcast (GitHub Copilot boosting developers' productivity). There are no undefined sources or self-attributions, so the rule applies.
If there was a discussion on the Behind the Tech podcast about GitHub Copilot boosting developers' productivity, does this necessarily mean that GitHub Copilot actually boosts developers' productivity? The existence of a discussion does not guarantee the truth of the discussion's content, so no. 
If there was a discussion on the Behind the Tech podcast about GitHub Copilot boosting developers' productivity, does this necessarily mean that GitHub Copilot is a tool developed by Microsoft? The discussion does not explicitly state that GitHub Copilot is a tool developed by Microsoft, so no. Therefore, S does not entail all elements of C.

### Example 3
Question: What was the impact of the tanker explosion in the Gulf of Mexico?
Excerpt from the response: The Earth Protectors, an environmental group, examined the remains of the tanker ship's explosion. Source [3] says they discovered that the resulting oil spill caused significant damage to the environment, underscoring the need for stricter regulations.
Sentence of interest: Source [3] says they discovered that the resulting oil spill caused significant damage to the environment, underscoring the need for stricter regulations.
Claim: They discovered the oil spill and its damage to the aquatic environment

S = Source [3] says they discovered that the resulting oil spill caused significant damage to the environment, underscoring the need for stricter regulations.
Describe the context for S; if someone read S in this context, how would they interpret it? The question is about the impact of the tanker explosion in the Gulf of Mexico, and the excerpt discusses the Earth Protectors' findings. Therefore, S would likely be interpreted as the Earth Protectors identified that the oil spill resulting from the tanker ship's explosion in the Gulf of Mexico caused significant damage to the environment, which emphasizes the necessity for stricter regulations.
C = They discovered the oil spill and its damage to the aquatic environment
A reader would interpret the claim as the Earth Protectors discovered the oil spill itself, and they also discovered the damage that the oil spill caused to the aquatic environment.
What are ALL elements of C? (1) They discovered the oil spill, (2) They discovered its damage to the aquatic environment.
Does the Statements and Actions Rule apply to S, or does it qualify as an exception? S contains an attribution to an undefined source ("Source [3] says"), so we can ignore this attribution and treat it as a regular statement. However, the rest of S is a statement about what the Earth Protectors discovered (the resulting oil spill caused significant damage to the environment), so it
applies.
If the Earth Protectors identified that the oil spill caused significant damage to the environment, does this necessarily mean that they discovered the oil spill? Identifying the environmental damage caused by the oil spill does not guarantee that they discovered the oil spill, since it's possible to identify the damage without discovering the oil spill itself, so no. If the Earth Protectors identified the environmental damage caused by the oil spill, does this necessarily mean that they discovered the oil spill's damage to the aquatic environment? The environment is not necessarily the aquatic environment, so no. Therefore, S does not entail all elements of C.
\end{lstlisting}
\end{tcolorbox}

\begin{tcolorbox}[
    breakable,                    
    colback=white,                
    colframe=black,              
    title=Entailment User Prompt,       
    title after break=Entailment User Prompt (Continued),
    fonttitle=\bfseries, 
    coltext=black,
]
\begin{lstlisting}[breaklines=true, breakindent=0pt, basicstyle=\small\ttfamily\raggedright, xleftmargin=-5pt, frame=none, xrightmargin=-5pt, aboveskip=-2pt, belowskip=-2pt]
Question:
{question}

Excerpt from response: 
{excerpt}

Sentence of interest: 
{sentence}

Claim:
{claim}

REMEMBER: if the context of S entails C, but S itself does not, you should still conclude that S entails C.
\end{lstlisting}
\end{tcolorbox}

\subsubsection{Element Extraction}
\label{app:element_extraction}
\begin{tcolorbox}[
    breakable,                    
    colback=white,                
    colframe=black,              
    title=Element Extraction System Prompt,       
    title after break=Element Extraction System Prompt (Continued),
    fonttitle=\bfseries, 
    coltext=black,
]
\begin{lstlisting}[breaklines=true, breakindent=0pt, basicstyle=\small\ttfamily\raggedright, xleftmargin=-5pt, frame=none, xrightmargin=-5pt, aboveskip=-2pt, belowskip=-2pt]
## Overview
You will be given a question, an excerpt from the response to the question, and a sentence of interest from the excerpt (which will be referred to as S). 

Your task is to (1) identify all elements of S (excluding elements about citations), and (2) for each element, determine whether it contains verifiable information. Follow these steps:
1. Print "S = <insert sentence of interest here EXACTLY as written>"
2. Are there any clarifications needed to understand S based on its context? If so, provide them. Then set S_restated to a version of the sentence restated in your own words, making sure that it fully reflects the meaning of S and no information is removed.
3. Does the Statements and Actions Rule apply? See the description of the rule below.
4. What are ALL elements of S_restated? Do not omit even subtle elements (e.g., "experts like John" implies "John is an expert"). Use this format: 
[
"<insert element> -> <insert verifiability>",
]

If the sentence is something like "John found X", "John reported X", "John emphasizes X", etc. (where John can be replaced with any entity or entities), it should be interpreted as a statement about what John says or does. For example, if the sentence is "John highlights that transparent communication is a critical part of Project Alpha", the element "transparent communication is a critical part of Project Alpha" is incorrect because it's missing the critical context that this is something John highlights. Let's call this the Statements and Actions Rule.

## Example
### Example 1
Question: What are the key factors driving the shift towards sustainability in the corporate world?
Excerpt from response: The growing public awareness of climate change has led to a surge in demand for sustainable products and services. For example, MiniCorp recently launched a new line of eco-friendly products that have been well-received by consumers. The 2020 Business Tracker reported that this inspired its competitors, such as TurboCorp and MegaCorp, to invest in sustainable packaging and renewable energy sources, highlighting the ripple effect of sustainable business practices.
Sentence of interest: The 2020 Business Tracker reported that this inspired its competitors, such as TurboCorp and MegaCorp, to invest in sustainable packaging and renewable energy sources, highlighting the ripple effect of sustainable business practices.

S = The 2020 Business Tracker reported that this inspired its competitors, such as TurboCorp and MegaCorp, to invest in sustainable packaging and renewable energy sources, highlighting the ripple effect of sustainable business 
practices.
Are there any clarifications needed to understand S based on its context? "This" refers to MiniCorp's success with its new line of eco-friendly products.
S_restated = The 2020 Business Tracker reported that MiniCorp's success with its new line of eco-friendly products has inspired its competitors, including TurboCorp and MegaCorp, to invest in sustainable packaging and renewable energy sources, which highlights the ripple effect of sustainable business practices.
Does the Statements and Actions Rule apply? Yes, because S is about what the 2020 Business Tracker reported.
What are ALL elements of S_restated?
[
"The 2020 Business Tracker reported that MiniCorp's success with its new line of eco-friendly products has inspired its competitors to invest in sustainable packaging -> contains verifiable information",
"The 2020 Business Tracker reported that MiniCorp's success with its new line of eco-friendly products has inspired its competitors to invest in renewable energy sources -> contains verifiable information",
"The 2020 Business Tracker reported that TurboCorp is an example of a competitor of MiniCorp that has been inspired by MiniCorp's success with its new line of eco-friendly products to invest in sustainable packaging -> contains verifiable information",
"The 2020 Business Tracker reported that TurboCorp is an example of a competitor of MiniCorp that has been inspired by MiniCorp's success with its new line of eco-friendly products to invest in renewable energy sources -> contains verifiable information",
"The 2020 Business Tracker reported that MegaCorp is an example of a competitor of MiniCorp that has been inspired by MiniCorp's success with its new line of eco-friendly products to invest in sustainable packaging -> contains verifiable information",
"The 2020 Business Tracker reported that MegaCorp is an example of a competitor of MiniCorp that has been inspired by MiniCorp's success with its new line of eco-friendly products to invest in renewable energy sources -> contains verifiable information",
"This highlights the ripple effect of sustainable business practices -> it's a generic statement, so it does not contain verifiable information",
]

### Example 2
Question: Who are key figures in the corporate sustainability movement?
Excerpt from response: There are also ongoing efforts to use partnerships as a means to improve sustainability, as demonstrated by Jane Smith. Many notable sustainability leaders like Smith do not work directly for a corporation, but her organization CleanTech has powerful partnerships with technology companies 
(e.g., MiniMax) to significantly improve waste management, demonstrating the power of collaboration.
Sentence of interest: Many notable sustainability leaders like Smith do not work directly for a corporation, but her organization CleanTech has powerful partnerships with technology companies (e.g., MiniMax) to significantly improve waste management, demonstrating the power of collaboration.

S = Many notable sustainability leaders like Smith do not work directly for a corporation, but her organization CleanTech has powerful partnerships with technology companies (e.g., MiniMax) to significantly improve waste management, demonstrating the power of collaboration.
Are there any clarifications needed to understand S based on its context? 
"Smith" refers to Jane Smith.
S_restated = Jane Smith is an example of a notable sustainability leader who does not work directly for a corporation, but her organization CleanTech has powerful partnerships with technology companies, including MiniMax, to significantly improve waste management, which demonstrates the power of collaboration.
Does the Statements and Actions Rule apply? No.
What are ALL elements of S_restated?
[
"Jane Smith is an example of a notable sustainability leader -> 'notable' is not verifiable, but the rest is verifiable, so it contains verifiable information",
"Jane Smith does not work directly for a corporation -> contains verifiable information",
"Jane Smith has an organization called CleanTech -> contains verifiable information",
"CleanTech has powerful partnerships with technology companies to significantly improve waste management -> 'powerful' and 'significantly' are not verifiable, but the rest is verifiable, so it contains verifiable information",
"MiniMax is a technology company -> contains verifiable information",
"CleanTech has a partnership with MiniMax to significantly improve waste management -> 'significantly' is not verifiable, but the rest is verifiable, so it contains verifiable information",
"CleanTech demonstrates the power of collaboration -> it's an interpretation, so it does not contain verifiable information",
]
\end{lstlisting}
\end{tcolorbox}

\begin{tcolorbox}[
    breakable,                    
    colback=white,                
    colframe=black,              
    title=Element Extraction User Prompt,       
    title after break=Element Extraction User Prompt (Continued),
    fonttitle=\bfseries, 
    coltext=black,
]
\begin{lstlisting}[breaklines=true, breakindent=0pt, basicstyle=\small\ttfamily\raggedright, xleftmargin=-5pt, frame=none, xrightmargin=-5pt, aboveskip=-2pt, belowskip=-2pt]
Question:
{question}

Excerpt from response:
{excerpt}

Sentence of interest: 
{sentence}
\end{lstlisting}
\end{tcolorbox}

\subsubsection{Element Coverage}
\label{app:element_coverage}
\begin{tcolorbox}[
    breakable,                    
    colback=white,                
    colframe=black,              
    title=Element Coverage System Prompt,       
    title after break=Element Coverage System Prompt (Continued),
    fonttitle=\bfseries, 
    coltext=black,
]
\begin{lstlisting}[breaklines=true, breakindent=0pt, basicstyle=\small\ttfamily\raggedright, xleftmargin=-5pt, frame=none, xrightmargin=-5pt, aboveskip=-2pt, belowskip=-2pt]
## Overview
You will be given a question and an excerpt from the response to the question. You will also be given a dictionary of claims extracted from the excerpt (which will be referred to as C), and a dictionary of elements (which will be referred to as E).

An element is "covered by" a claim if the element is explicitly stated or strongly implied by the claim. For each element in E, your task is to determine whether the information in the element is covered by C by following these steps:
1. Print "E<insert number here>: <insert element here EXACTLY as written>" where number is the key in the dictionary and element is the value.
2. Determine whether the information in the element is covered by C. If the element has a note that some information is not verifiable, ignore that part and focus on the verifiable information. You CANNOT use any external information 
(e.g., if the element says "Politicians like John frequently discuss the 
economy" and C says "John frequently discusses the economy" but there is no claim that John is a politician, even if you have external information that John is a politician, the element is not fully covered by C). If C is more specific than E, you must check whether the specificity is merited based on the question and the excerpt (i.e., if the elements should be more specific based on the 
context); if it is merited, then the element is fully covered by C. Print either "fully covered by C" or "not fully covered by C".
3. Repeat this process for all elements in E.

If the element is something like "John found X", "John reported X", "John emphasizes X", etc. (where John can be replaced with any entity or entities), it should be interpreted as a statement about what John says or does. For example, if the element is "John highlights that transparent communication is a critical part of Project Alpha", the claim "transparent communication is a critical part of Project Alpha" does not cover the element because it's missing the critical context that this is something John highlights. Let's call this the Statements and Actions Rule.

## Examples
### Example 1
Question: What are the key factors driving the shift towards sustainability in the fashion industry?
Excerpt from response: The growing public awareness of climate change has led to a surge in demand for sustainable fashion products. For example, MiniCorp recently launched a new line of eco-friendly scarves that have been well-received by consumers. The 2020 Business Tracker reported that this inspired its competitors, such as TurboCorp, to invest in sustainable packaging, highlighting the ripple effect of sustainable business practices.
Claims (C): {
1: "The 2020 Business Tracker reported that MiniCorp inspired its competitors to invest in sustainable packaging",
2: "The 2020 Business Tracker reported that TurboCorp was inspired by MiniCorp",
3: "TurboCorp is a competitor of MiniCorp",
4: "TurboCorp invested in sustainable packaging because it was inspired by MiniCorp",
5: "MiniCorp inspiring its competitors to adopt sustainable practice illustrates the ripple effect of sustainable business practices in the fashion industry",
}
Elements (E): {
1: "The 2020 Business Tracker reported that MiniCorp's success with its new line of eco-friendly scarves has inspired its competitors to invest in sustainable packaging",
2: "The 2020 Business Tracker reported that TurboCorp is an example of a competitor of MiniCorp that has been inspired by MiniCorp's success with its new line of eco-friendly scarves to invest in sustainable packaging",
3: "This highlights the ripple effect of sustainable business practices",
}

E1: The 2020 Business Tracker reported that MiniCorp's success with its new line of eco-friendly scarves has inspired its competitors to invest in sustainable packaging
- The Statements and Actions Rule applies because the element is about what the 2020 Business Tracker reported
- C1 says "The 2020 Business Tracker reported that MiniCorp inspired its competitors to invest in sustainable packaging"
- What is not explicitly stated or strongly implied by C, and is therefore grounds for lack of full coverage? It does not specify that it was MiniCorp's success with its new line of eco-friendly products that inspired its
competitors. Therefore E1 is not fully covered by C
E2: The 2020 Business Tracker reported that TurboCorp is an example of a competitor of MiniCorp that has been inspired by MiniCorp's success with its new line of eco-friendly products to invest in sustainable packaging
- The Statements and Actions Rule applies because the element is about what the 2020 Business Tracker reported
- C2 says "The 2020 Business Tracker reported that TurboCorp was inspired by MiniCorp" and C3 says "TurboCorp is a competitor of MiniCorp" and C4 says 
"TurboCorp invested in sustainable packaging because it was inspired by 
MiniCorp"
- What is not explicitly stated or strongly implied by C?, and is therefore grounds for lack of full coverage Only C2 explicitly states that it was reported by the 2020 Business Tracker, and C does not specify that it was MiniCorp's success with its new line of eco-friendly products that inspired TurboCorp. Therefore E2 is not fully covered by C
E3: This highlights the ripple effect of sustainable business practices
- C5 says "MiniCorp inspiring its competitors to adopt sustainable practice illustrates the ripple effect of sustainable business practices in the fashion industry"
- What is not explicitly stated or strongly implied by C, and is therefore grounds for lack of full coverage? C5 is more specific than E3, so we need to check whether the specificity is merited based on the question and the excerpt. The question is about the key factors driving the shift towards sustainability in the fashion industry, and the excerpt discusses MiniCorp's success with its new line of eco-friendly scarves, so this specificity is merited. Therefore E3 is fully covered by C

### Example 2
Question: Who are key figures in the corporate sustainability movement?
Excerpt from response: There are also ongoing efforts to use partnerships as a means to improve sustainability, as demonstrated by Jane Smith. Many notable sustainability leaders like Smith do not work directly for a corporation, but her organization CleanTech has powerful partnerships with technology companies (e.g., MiniMax) to significantly improve waste management, demonstrating the power of collaboration.
Claims (C): {
1: "Jane is a sustainability leader",
2: "Jane doesn't work directly for a corporation",
3: "CleanTech has partnerships with technology companies to improve waste management",
4: "CleanTech has a partnership with MiniMax",
}
Elements (E): {
1: "Jane Smith is an example of a notable sustainability leader [note: 'notable' is not verifiable, but the rest is verifiable]",
2: "Jane Smith does not work directly for a corporation",
3: "Jane Smith has an organization called CleanTech",
4: "CleanTech has powerful partnerships with technology companies to significantly improve waste management [note: 'powerful' and 'significantly' are not verifiable, but the rest is verifiable]",
5: "MiniMax is a technology company",
6: "CleanTech has a partnership with MiniMax to significantly improve waste management [note: 'significantly' is not verifiable, but the rest is verifiable]",
7: "CleanTech demonstrates the power of collaboration",
}

Element 1: Jane Smith is an example of a notable sustainability leader
- C1 says "Jane is a sustainability leader", and "notable" is not verifiable so it can be ignored
- What is not explicitly stated or strongly implied by C, and is therefore grounds for lack of full coverage? Nothing. The verifiable parts of the element are explicitly stated. Therefore E1 is fully covered by C
Element 2: Jane Smith does not work directly for a corporation
- C2 says "Jane does not work directly for a corporation"
- What is not explicitly stated or strongly implied by C, and is therefore grounds for lack of full coverage? Nothing. The element is explicitly stated. Therefore E2 is fully covered by C
Element 3: Jane Smith has an organization called CleanTech
- C does not state that CleanTech is Jane's organization
- What is not explicitly stated or strongly implied by C, and is therefore grounds for lack of full coverage? Nothing. The element is explicitly stated. Therefore E3 is not fully covered by C
Element 4: CleanTech has powerful partnerships with technology companies to significantly improve waste management
- C3 says "CleanTech has partnerships with technology companies to improve waste management", and "powerful" and "significantly" are not verifiable so they can be ignored
- What is not explicitly stated or strongly implied by C, and is therefore grounds for lack of full coverage? Nothing. The verifiable parts of the element are explicitly stated. Therefore E4 is fully covered by C
Element 5: MiniMax is a technology company
- C4 says "CleanTech has a partnership with MiniMax"
- What is not explicitly stated or strongly implied by C, and is therefore grounds for lack of full coverage? It does not say that MiniMax is a technology company. Therefore E5 is not fully covered by C
Element 6: CleanTech has a partnership with MiniMax to significantly improve waste management
- C4 says "CleanTech has a partnership with MiniMax"
- What is not explicitly stated or strongly implied by C, and is therefore grounds for lack of full coverage? It does not say that the purpose of the partnership is to improve waste management. Therefore E6 is not fully covered by C
Element 7: CleanTech demonstrates the power of collaboration
- C3 says "CleanTech has partnerships with technology companies to improve waste management"
- What is not explicitly stated or strongly implied by C, and is therefore grounds for lack of full coverage? It does not explicitly state that CleanTech demonstrates the power of collaboration, but C strongly implies it. Therefore it is implied that E7 is fully covered by C
\end{lstlisting}
\end{tcolorbox}

\begin{tcolorbox}[
    breakable,                    
    colback=white,                
    colframe=black,              
    title=Element Coverage User Prompt,       
    title after break=Element Coverage User Prompt (Continued),
    fonttitle=\bfseries, 
    coltext=black,
]
\begin{lstlisting}[breaklines=true, breakindent=0pt, basicstyle=\small\ttfamily\raggedright, xleftmargin=-5pt, frame=none, xrightmargin=-5pt, aboveskip=-2pt, belowskip=-2pt]
Question (context for E):
{question}

Excerpt from response (context for E):
{excerpt}

Claims (C):
{claims}

Elements (E):
{elements}
\end{lstlisting}
\end{tcolorbox}

\subsubsection{Decontextualization}
\begin{tcolorbox}[
    breakable,                    
    colback=white,                
    colframe=black,              
    title=Decontextualization System Prompt,       
    title after break=Decontextualization System Prompt (Continued),
    fonttitle=\bfseries, 
    coltext=black,
    after skip = -2pt,
]
\begin{lstlisting}[breaklines=true, breakindent=0pt, basicstyle=\small\ttfamily\raggedright, xleftmargin=-5pt, frame=none, xrightmargin=-5pt, aboveskip=-2pt, belowskip=-2pt]
## Overview
You will be given a question, an excerpt from the response to the question, a sentence of interest from the excerpt, the claims derived from the sentence, and a claim of interest (which will be referred to as C).

A claim is "decontextualized" if (1) it is fully self-contained, meaning it can be understood in isolation (i.e., without the question, the excerpt, the sentence, and the other claims), AND (2) its meaning in isolation matches its meaning when interpreted alongside the question, the excerpt, the sentence, and the other claims. 

Your task is to create C_max, the maximally decontextualized version of C by following these steps:
1. Print "C = <insert claim of interest here EXACTLY as written>"
2. If someone read C without any context, would they have any questions? If yes, are any of these questions answered by the sentence or its context? If the reader would not have any questions, or none of the questions are clearly answered by the sentence or its context, or the claim already provides sufficient answers to the questions, print "C_max = C". Otherwise, set C_max equal to the maximally decontextualized claim that clarifies the answers to the questions that would be asked. Only include clarifications that are clearly attributable to the sentence and its context.

## Examples
### Example 1
Question: Provide an overview of the Supreme Court's importance in the United States.
Excerpt from response: Another example of the court's importance is the decision in Roe v. Wade in January 1973. It significantly affected abortion laws across the United States.
Sentence of interest: It significantly affected abortion laws across the United States.
All claims: ["The court's decision affected abortion laws across the United States."]
Claim of interest: The court's decision affected abortion laws across the United States.

C = The court's decision affected abortion laws across the United States.
Would someone reading C without any context have questions? They would likely ask which court made the decision and what the decision was.
Are any of these questions answered by the sentence or its context? The court is the Supreme Court, and the decision was Roe v. Wade in January 1973.
C_max = The Supreme Court's decision in Roe v. Wade in January 1973 affected abortion laws across the United States.

### Example 2
Question: Who is Jane Doe?
Excerpt from response: Jane Doe has been a long-time advocate for environmental causes. In 2022, she spoke at Climate Action Now. She plans to speak at Youth for Climate, Moving Forward, and several other conferences next year.
Sentence of interest: She plans to speak at Youth for Climate, Moving Forward, and several other conferences next year.
All claims: ["She plans to speak at Youth for Climate next year", "She plans to speak at Moving Forward next year", "She plans to speak at other conferences next year"]
Claim of interest: She plans to speak at other conferences next year.

C = She plans to speak at other conferences next year.
Would someone reading C without any context have questions? They would likely ask who she is, what "other conferences" means, and what year "next year" refers to.
Are any of these questions answered by the sentence or its context? She is Jane Doe, "other conferences" means conferences other than Youth for Climate and Moving Forward (since they are covered by the other claims), and "next year" is 2023.
C_max = Jane Doe plans to speak at conferences other than Youth for Climate and Moving Forward in 2023.
\end{lstlisting}
\end{tcolorbox}

\begin{tcolorbox}[
    breakable,                    
    colback=white,                
    colframe=black,              
    title=Decontextualization User Prompt,       
    title after break=Decontextualization User Prompt (Continued),
    fonttitle=\bfseries, 
    coltext=black,
    before skip=6pt,
]
\begin{lstlisting}[breaklines=true, breakindent=0pt, basicstyle=\small\ttfamily\raggedright, xleftmargin=-5pt, frame=none, xrightmargin=-5pt, aboveskip=-2pt, belowskip=-2pt]
Question:
{question}

Excerpt from response: 
{excerpt}

Sentence of interest: 
{sentence}

All claims:
{claims}

Claim of interest: 
{claim}
\end{lstlisting}
\end{tcolorbox}

\subsection{Invalid Statements}
\label{app:invalid_statements_prompts}
\subsubsection{Invalid Sentences}
\begin{tcolorbox}[
    breakable,                    
    colback=white,                
    colframe=black,              
    title=Invalid Sentences System Prompt,      
    title after break=Invalid Sentences System Prompt (Continued),
    fonttitle=\bfseries, 
    coltext=black,
    after skip=-1pt
]
\begin{lstlisting}[breaklines=true, breakindent=0pt, basicstyle=\small\ttfamily\raggedright, xleftmargin=-5pt, frame=none, xrightmargin=-5pt, aboveskip=-2pt, belowskip=-2pt]
## Overview
You will be given a question, an excerpt from the response to the question, and a sentence of interest from the excerpt (which will be referred to as S).

Your task is to determine whether S, in light of its context, can be interpreted as a complete, declarative sentence by following these steps:
1. Print "S = <insert sentence of interest here EXACTLY as written>"
2. Describe the context for S. 
3. Can S be interpreted as a complete, declarative sentence as is? If not, given its context, can it be rewritten as a complete, declarative sentence? If yes, print "S can be interpreted as a complete, declarative sentence"; otherwise, print "S cannot be interpreted as a complete, declarative sentence".

## Examples
### Example 1
Question: How can companies improve their sustainability practices?
Excerpt from response: Some examples include:
- Reducing energy consumption by using energy-efficient appliances
- Implementing recycling programs
- Sourcing materials from sustainable suppliers
Sentence of interest: Some examples include:

S = Some examples include:
Describe the context for S. The question is about how companies can improve their sustainability practices, and S is the header for a list of examples. 
Can S be interpreted as a complete, declarative sentence as is? If not, given its context, can it be rewritten as a complete, declarative sentence? S is not a complete, declarative sentence as is. Since it merely introduces the list of examples without providing a complete thought, it cannot be rewritten as a complete, declarative sentence. Therefore, S cannot be interpreted as a 
complete, declarative sentence.
 
### Example 2
Question: How are companies improving their sustainability practices?
Excerpt from response: Some examples include:
- Reducing energy consumption by using energy-efficient appliances
- Implementing recycling programs
- Sourcing materials from sustainable suppliers
Sentence of interest: - Sourcing materials from sustainable suppliers

S = - Sourcing materials from sustainable suppliers
Describe the context for S. The question is about how companies are improving their sustainability practices, and the excerpt provides a list of examples. 
Can S be interpreted as a complete, declarative sentence as is? If not, given its context, can it be rewritten as a complete, declarative sentence? S is not a complete, declarative sentence as is. However, in the context of the question and the excerpt, it can be rewritten as "An example of how companies are improving their sustainability practices is sourcing materials from sustainable suppliers". Therefore, S can be interpreted as a complete, declarative sentence.
\end{lstlisting}
\end{tcolorbox}

\begin{tcolorbox}[
    breakable,                    
    colback=white,                
    colframe=black,              
    title=Invalid Sentences User Prompt,       
    title after break=Invalid Sentences User Prompt (Continued),
    fonttitle=\bfseries, 
    coltext=black,
    before skip=5pt,
]
\begin{lstlisting}[breaklines=true, breakindent=0pt, basicstyle=\small\ttfamily\raggedright, xleftmargin=-5pt, frame=none, xrightmargin=-5pt, aboveskip=-2pt, belowskip=-2pt]
Question:
{question}

Excerpt from response: 
{excerpt}

Sentence of interest: 
{sentence}
\end{lstlisting}
\end{tcolorbox}

\subsubsection{Invalid Claims}
\begin{tcolorbox}[
    breakable,                    
    colback=white,                
    colframe=black,              
    title=Invalid Claims System Prompt,       
    title after break=Invalid Claims System Prompt (Continued),
    fonttitle=\bfseries, 
    coltext=black,
]
\begin{lstlisting}[breaklines=true, breakindent=0pt, basicstyle=\small\ttfamily\raggedright, xleftmargin=-5pt, frame=none, xrightmargin=-5pt, aboveskip=-2pt, belowskip=-2pt]
## Overview
You will be given a claim (which will be referred to as C). Your task is to determine whether C, in isolation, is a complete, declarative sentence, by following these steps:
1. Print "C = <insert claim of interest here EXACTLY as written>"
2. In isolation, is C a complete, declarative sentence? After your reasoning, print either "C is not a complete, declarative sentence" or "C is a complete, declarative sentence".

## Examples
### Example 1
Claim: Sourcing materials from sustainable suppliers is an example of how companies are improving their sustainability practices

C = Sourcing materials from sustainable suppliers is an example of how companies are improving their sustainability practices
In isolation, is C a complete, declarative sentence? Yes, C is a complete, declarative sentence.

### Example 2
Claim: Sourcing materials from sustainable suppliers

C = Sourcing materials from sustainable suppliers
In isolation, is C a complete, declarative sentence? It's missing a subject and a verb, so C is not a complete, declarative sentence.
\end{lstlisting}
\end{tcolorbox}

\begin{tcolorbox}[
    breakable,                    
    colback=white,                
    colframe=black,              
    title=Invalid Claims User Prompt,       
    title after break=Invalid Claims User Prompt (Continued),
    fonttitle=\bfseries, 
    coltext=black,
]
\begin{lstlisting}[breaklines=true, breakindent=0pt, basicstyle=\small\ttfamily\raggedright, xleftmargin=-5pt, frame=none, xrightmargin=-5pt, aboveskip=-2pt, belowskip=-2pt]
Claim:
{claim}
\end{lstlisting}
\end{tcolorbox}

\twocolumn

\end{document}